%% file: paper.tex
\documentclass{article}

\usepackage{microtype}
\usepackage{graphicx}
\usepackage{booktabs} 

\usepackage{hyperref}

\usepackage[accepted]{icml2024}

\usepackage{amsmath}
\usepackage{amssymb}
\usepackage{mathtools}
\usepackage{amsthm}

\usepackage[capitalize,noabbrev]{cleveref}

\theoremstyle{plain}
\newtheorem{theorem}{Theorem}[section]
\newtheorem{proposition}[theorem]{Proposition}

\theoremstyle{definition}

\theoremstyle{remark}
\newtheorem{remark}[theorem]{Remark}

\usepackage[textsize=tiny]{todonotes}

\input{math_commands.tex}

\usepackage{url}
\usepackage{caption}
\usepackage{makecell}
\usepackage{blindtext}
\usepackage{bm}
\usepackage{bbm} 
\usepackage{comment}
\usepackage{algorithm}
\usepackage{algorithmic}
\usepackage{multirow}
\usepackage{array}
\usepackage{enumitem}
\usepackage{setspace}
\usepackage[safe]{tipa} 
\usepackage{color, colortbl}
\usepackage[misc,geometry]{ifsym} 
\usepackage{pifont,makecell,subcaption}
\usepackage{wrapfig}

\providecommand{\eg}{\textit{e.g.,}\@\xspace}
\providecommand{\ie}{\textit{i.e.,}\@\xspace}

\usepackage[capitalize]{cleveref}
\Crefname{section}{Section}{Sections}
\crefname{section}{Sec.}{Sec.}
\Crefname{table}{Table}{Tables}
\crefname{table}{Tab.}{Tab.}
\Crefname{figure}{Figure}{Figures}
\crefname{figure}{Fig.}{Fig.}
\Crefname{equation}{Equation}{Equations}
\crefname{equation}{Eq.}{Eq.}

\def\algo{{LIFT}}
\def\gray{\color{gray}}
\def\red{\color{red}}

\def\oP{{\operatorname{P}}}

\definecolor{mplgreen}{rgb}{0.17254901960784313, 0.6274509803921569, 0.17254901960784313}
\definecolor{myblue}{rgb}{0.2, 0.2, 0.9}
\definecolor{myred}{rgb}{1.0, 0.1, 0.1}

\urldef\imageneturl\url{https://storage.googleapis.com/bit_models/imagenet21k_wordnet_lemmas.txt}
\urldef\cifarurl\url{https://www.cs.toronto.edu/%7Ekriz/cifar.html}
\urldef\adaptformerurl\url{https://github.com/ShoufaChen/AdaptFormer/blob/main/models/adapter.py}

\newcolumntype{L}[1]{>{\raggedright\arraybackslash}p{#1}}
\newcolumntype{R}[1]{>{\raggedleft\arraybackslash}p{#1}}
\newcolumntype{C}[1]{>{\centering\arraybackslash}p{#1}}

\captionsetup[figure]{aboveskip=0.1in, belowskip=0.0in}
\captionsetup[table]{aboveskip=0.1in, belowskip=0.02in}
\captionsetup[subfigure]{aboveskip=0.05in,belowskip=0.0in}
\captionsetup[subtable]{aboveskip=0.05in,belowskip=0.0in}

\allowdisplaybreaks

\icmltitlerunning{Long-Tail Learning with Foundation Model: Heavy Fine-Tuning Hurts}

\begin{document}

\twocolumn[
\icmltitle{Long-Tail Learning with Foundation Model: Heavy Fine-Tuning Hurts}



\icmlsetsymbol{equal}{*}

\begin{icmlauthorlist}
\icmlauthor{Jiang-Xin Shi}{equal,skl,njuai}
\icmlauthor{Tong Wei}{equal,seucse,edulab}
\icmlauthor{Zhi Zhou}{skl}
\icmlauthor{Jie-Jing Shao}{skl}
\icmlauthor{Xin-Yan Han}{skl}
\icmlauthor{Yu-Feng Li}{skl,njuai}
\end{icmlauthorlist}

\icmlaffiliation{skl}{National Key Laboratory for Novel Software Technology, Nanjing University, China}
\icmlaffiliation{njuai}{School of Artificial Intelligence, Nanjing University, China}
\icmlaffiliation{seucse}{School of Computer Science and Engineering, Southeast University, China}
\icmlaffiliation{edulab}{Key Laboratory of Computer Network and Information Integration, Southeast University, Ministry of Education, China}

\icmlcorrespondingauthor{Yu-Feng Li}{liyf@nju.edu.cn}

\icmlkeywords{Machine Learning, ICML}

\vskip 0.3in
]



\printAffiliationsAndNotice{\icmlEqualContribution} 

\begin{abstract}
The fine-tuning paradigm in addressing long-tail learning tasks has sparked significant interest since the emergence of foundation models. Nonetheless, how fine-tuning impacts performance in long-tail learning was not explicitly quantified. In this paper, we disclose that heavy fine-tuning may even lead to non-negligible performance deterioration on tail classes, and lightweight fine-tuning is more effective. The reason is attributed to inconsistent class conditions caused by heavy fine-tuning. With the observation above, we develop a low-complexity and accurate long-tail learning algorithms \algo\ with the goal of facilitating fast prediction and compact models by adaptive lightweight fine-tuning. Experiments clearly verify that both the training time and the learned parameters are significantly reduced with more accurate predictive performance compared with state-of-the-art approaches. The implementation code is available at \url{https://github.com/shijxcs/LIFT}.
\end{abstract}

\section{Introduction}
\begin{figure}[!h]
    \centering
    \includegraphics[width=0.49\linewidth]{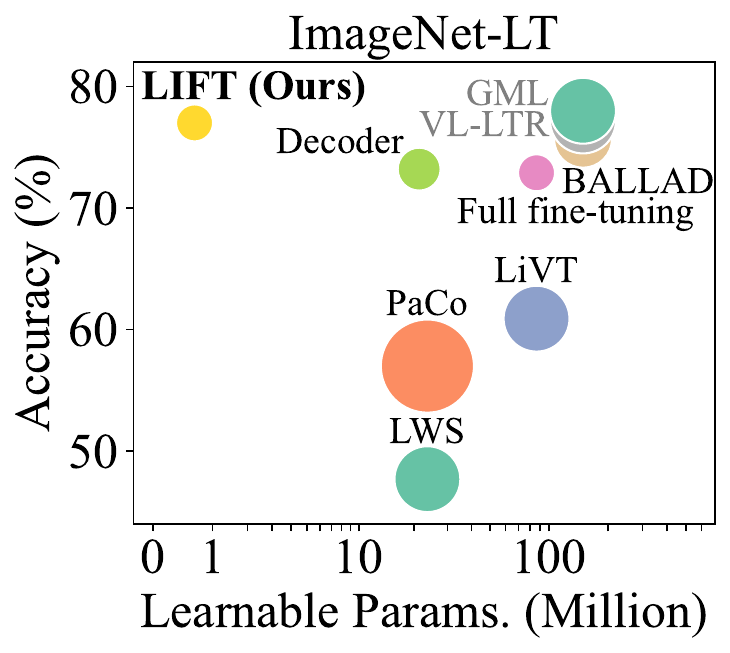}
    \includegraphics[width=0.49\linewidth]{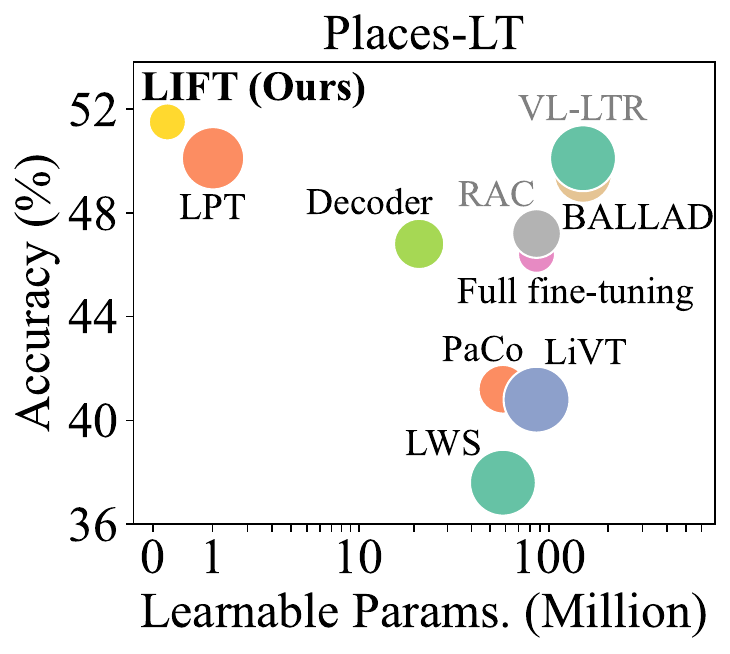}
    \caption{Comparison with state-of-the-art methods. The x-axis represents the number of learnable parameters and the y-axis shows the test accuracy. \textbf{The size of each point corresponds to the number of training epochs}, with larger points indicating longer training time. Gray labels denote methods that incorporate external data. \algo\ consistently achieves higher performance with lower costs and is even comparable with methods that leverage external data.}
    \label{fig:sota}
\end{figure}

\begin{figure*}[!t]
    \centering
    \begin{subfigure}{0.38\linewidth}
        \includegraphics[width=\linewidth, trim=0 9 0 0]{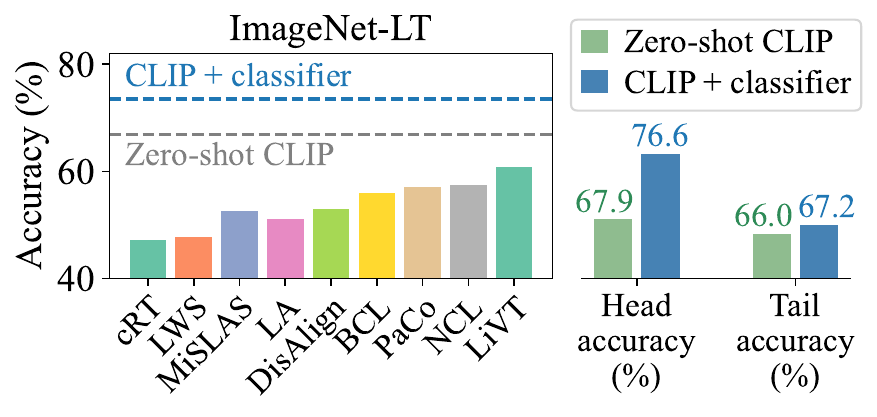}
        \abovecaptionskip=0.0in
        \caption{}
        \label{fig:zsclip-a}
    \end{subfigure}
    \hfill
    \begin{subfigure}{0.36\linewidth}
        \includegraphics[width=\linewidth, trim=0 9 0 0]{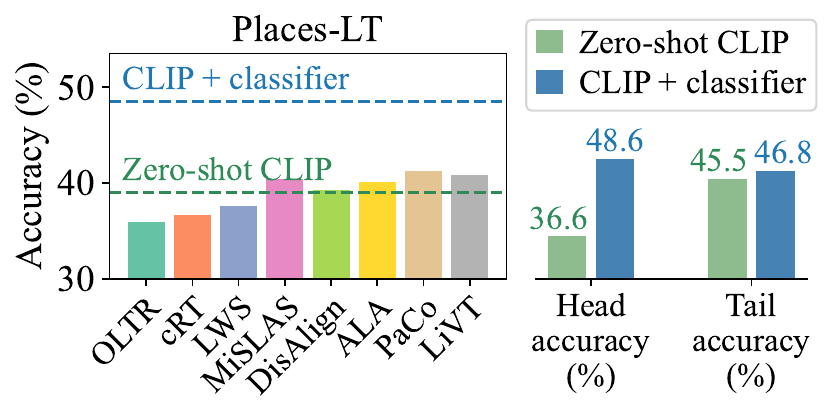}
        \abovecaptionskip=0.0in
        \caption{}\label{fig:zsclip-b}
    \end{subfigure}
    \hfill
    \begin{subfigure}{0.216\linewidth}
        \includegraphics[width=\linewidth, trim=0 0 0 0]{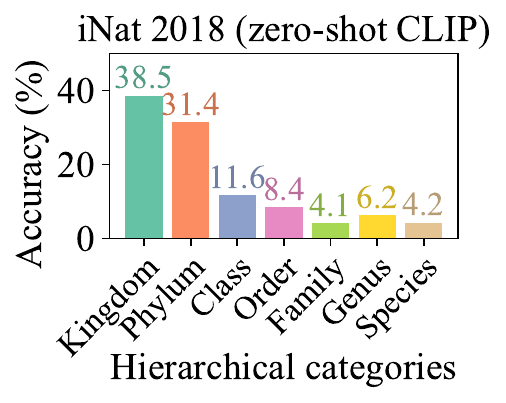}
        \abovecaptionskip=0.0in
        \caption{}\label{fig:zsclip-c}
    \end{subfigure}
    \caption{(a-b) On ImageNet-LT and Places-LT, zero-shot CLIP has surpassed many prior methods. By simply introducing an additional classifier, the accuracy further increases. However, the improvements mainly come from the head classes, while the tail classes only achieve marginal enhancements. (c) On iNaturalist 2018, zero-shot CLIP encounters challenges in achieving high accuracy for fine-grained long-tail categories.}
    \label{fig:zsclip}
\end{figure*}

Long-tail learning addresses the challenge of learning from highly imbalanced data, where a small set of classes (head classes) is well-represented in the training data, while other classes (tail classes) have only a limited number of training samples available. Given its widespread attention, numerous long-tail learning approaches have emerged to enhance generalization, particularly for tail classes. These approaches typically fall into three categories: 1) data manipulation \citep{zhou2020bbn,kang2020decoupling}, 2) representation learning \citep{zhong2021improving,cui2021parametric}, and 3) model output adjustment \citep{ren2020balanced,menon2021longtail}. While these methods have made substantial strides, a significant gap still persists compared to models trained on class-balanced datasets.

Instead of training deep neural networks from scratch, recent results from BALLAD \citep{ma2021simple}, RAC \citep{long2022retrieval}, VL-LTR \citep{tian2022vl}, and LPT \citep{dong2023lpt} show that proper fine-tuning of foundation models such as CLIP \citep{radford2021clip} can surprisingly improve long-tail learning performance.
For example, BALLAD first fully fine-tunes the foundation model, then freezes the backbone and optimizes a linear adapter on the re-sampled data.
VL-LTR incorporates additional image-text web data during the fine-tuning process. RAC jointly fine-tunes an encoder and trains a retrieval module to augment the input image with external datasets such as ImageNet-21K. LPT fine-tunes the foundation model utilizing prompt tuning \citep{jia2022visual} via two-phrase training.

While these works introduce a new paradigm for long-tail learning, how fine-tuning affects performance in long-tail learning remains unexplored. In this paper, we reveal that heavy fine-tuning may lead to non-negligible performance deterioration on tail classes. To uncover the reasons behind it,
we identify that full fine-tuning distorts the intra-class distance distributions. In theory, we demonstrate that such distortions break the consistency assumption of class-conditions, consequently resulting in biased predictions. Based on the above analysis and observation, we realize that optimizing a small proportion of the pre-trained weights may not only enhance discriminative capacity, but also leave the intra-class distributions to be unaffected. To this end, we propose a low-complexity and accurate long-tail learning method named \textit{LIghtweight Fine-Tuning} (\algo) with the goal
of facilitating fast prediction and compact models
by adaptive lightweight fine-tuning.

To achieve performance improvements and balanced prediction, previous work usually comes at the cost with 1) long training epochs ($\approx 100$); 2) a two-staged procedure; and 3) an external training dataset ($size \approx 10^6$). To overcome it, \algo\ is a single-staged framework, which achieves convergence in fewer than 20 training epochs without requiring external training data. Furthermore, 
we introduce a semantic-aware classifier initialization to equip the model with a robust starting point.
Moreover, we employ a test-time ensembling to compensate for the intrinsic defects of foundation models and enhance the generalization. \algo\ brings small computational overhead, while consistently outperforming the state-of-the-art methods on a series of long-tail benchmark datasets. \Cref{fig:sota} gives a performance comparison on two typical datasets ImageNet-LT and Places-LT.

The contributions of this paper are summarized as follows: \textbf{1)} We identify a limitation in heavy fine-tuning distorts the tail-class performance; \textbf{2)} We discover that optimizing a small proportion of parameters helps;
\textbf{3)} We introduce a low-complexity and accurate long-tail learning method \algo\ with lightweight fine-tuning;
\textbf{4)} Comprehensive experiments demonstrate that \algo\ consistently outperforms the state-of-the-art methods with a lower computational cost.

\section{Long-Tail Learning with Foundation Model}

\paragraph{Preliminary.}
Different from conventional neural networks, in the foundation models, the Transformer architecture \citep{vaswani2017attention,dosovitskiy2021an} is more simply designed while exhibiting remarkable generalization capabilities.
It has been proven adaptable to a wide range of computer vision and natural language processing tasks \citep{kenton2019bert,dosovitskiy2021an}. Notably, the recent vision-language pre-training model, CLIP, further underscores its efficacy by demonstrating impressive zero-shot performance \citep{radford2021clip}.
When processing an image $\vx$ and considering $K$ candidate classes, CLIP first generates the textual prompts for the classes. These prompts are descriptive phrases, such as ``a photo of a cat'' or ``a photo of a dog''. CLIP extracts corresponding textual features $\vf_{T_1},\cdots,\vf_{T_K}$ and the image feature $\vf_{I}$. To predict the label for the given image, CLIP compares the cosine similarity between the image and each of the class prompts:
\begin{equation}
    y_{\text{pred}}=\operatorname*{arg\  max}\limits_{k\in[K]} \frac{\vf_I^{\top}\vf_{T_k}}{\Vert \vf_I\Vert_2 \Vert \vf_{T_k}\Vert_2}
\end{equation}

\paragraph{Long-Tail Learning with CLIP.}
In \Cref{fig:zsclip-a,fig:zsclip-b,fig:zsclip-c}, we evaluate the performance of CLIP on typical long-tail datasets. We discover that zero-shot CLIP outperforms most conventional methods. Furthermore, by freezing the backbone and learning an additional classifier, the performance can be further improved, thereby highlighting the effectiveness of the representations learned by CLIP. Nonetheless, the majority of performance gains are dominated by the head classes, while the tail classes struggle to achieve comparable improvements.

Moreover, while zero-shot CLIP shows impressive performance,
its good performance cannot be stably generalized to common long-tail datasets.
For example, in iNaturalist 2018 dataset, it poses a fine-grained long-tail challenge, featuring a hierarchical categorization system spanning from 7 kingdoms to 8142 species. Although zero-shot CLIP achieves high accuracy for predicting coarse-grained categories like ``kingdom'' and ``phylum'', it performs poorly in predicting fine-grained species, \ie long-tail classes.

\begin{figure*}[!t]
\centering
\setlength{\tabcolsep}{0.2ex} 
\begin{subfigure}{0.38\linewidth}
    \includegraphics[width=0.5\linewidth]{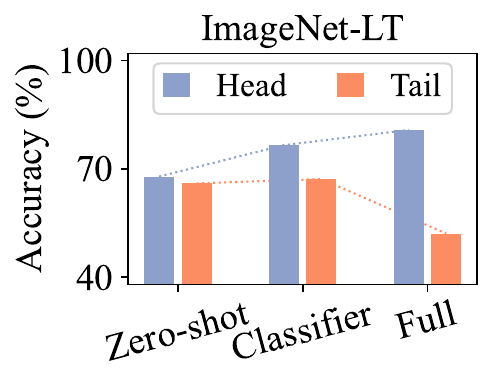}
    \includegraphics[width=0.48\linewidth]{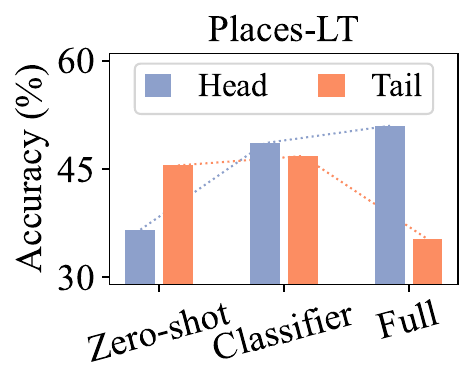}
    \caption{Performance of fine-tuning.}
    \label{fig:deterioration}
\end{subfigure}
\hfill
\begin{subfigure}{0.28\linewidth}
    \begin{tabular}{cc}
    \begin{minipage}{0.51\linewidth}
        \centering
        \includegraphics[width=\linewidth]{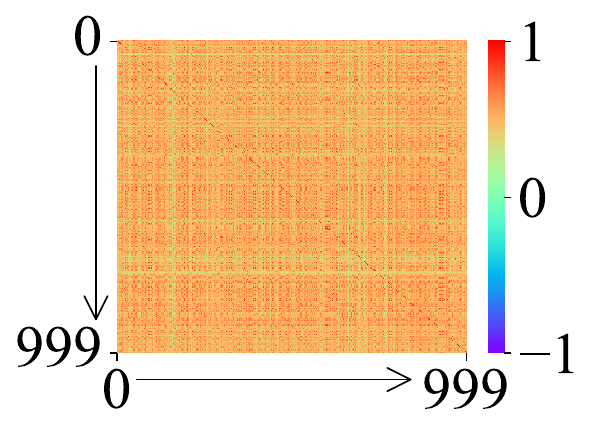}
    \end{minipage} &
    \begin{minipage}{0.47\linewidth}
        \centering
        \includegraphics[width=\linewidth]{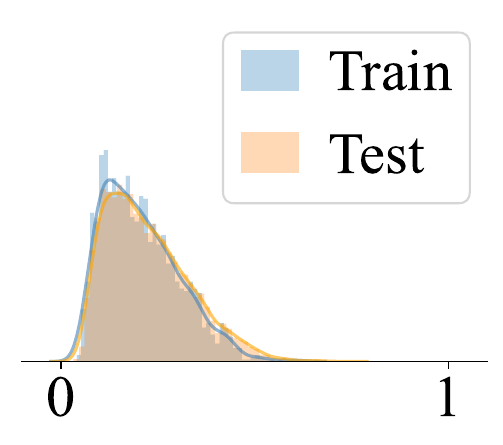}
    \end{minipage} \\
    \color{red}{\fontsize{8.5pt}{8.5pt}\selectfont Head acc.: 76.6\%} &
    \color{mplgreen}{\fontsize{8.5pt}{8.5pt}\selectfont Tail acc.: 67.2\%} \\
    \end{tabular}
    \caption{Classifier fine-tuning.}
    \label{fig:feature-classifier}
\end{subfigure}
\hfill
\begin{subfigure}{0.28\linewidth}
    \begin{tabular}{cc}
    \begin{minipage}{0.51\linewidth}
        \centering
        \includegraphics[width=\linewidth]{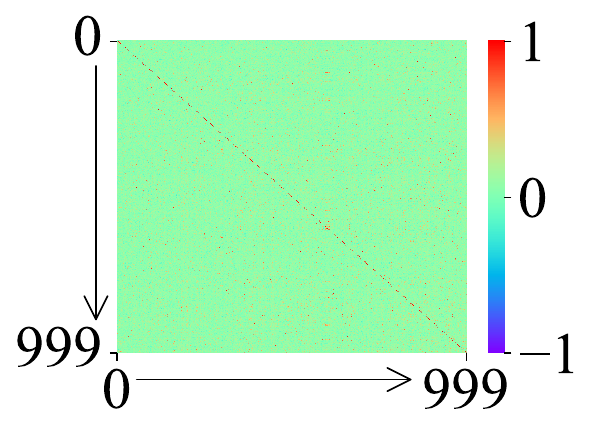}
    \end{minipage} &
    \begin{minipage}{0.47\linewidth}
        \centering
        \includegraphics[width=\linewidth]{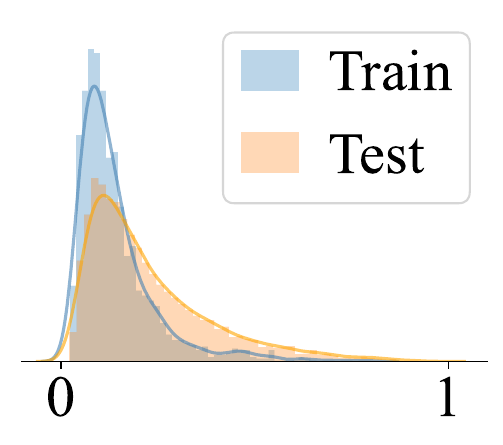}
    \end{minipage} \\
    \color{mplgreen}{\fontsize{8.5pt}{8.5pt}\selectfont Head acc.: 80.8\%} &
    \color{red}{\fontsize{8.5pt}{8.5pt}\selectfont Tail acc.: 52.1\%} \\
    \end{tabular}
    \caption{Full fine-tuning.}
    \label{fig:feature-full}
\end{subfigure}
\caption{(a) Full fine-tuning improves head-class accuracy while decreasing tail-class accuracy, even if we optimize the balanced LA loss. (b-c) Inter-class feature similarities (heatmaps) and intra-class distributions from tail classes (histograms) on ImageNet-LT. Classifier fine-tuning limits head-class performance due to unoptimized inter-class similarities. Full fine-tuning optimizes inter-class similarities but leads to inconsistent distribution between train and test data on tail classes.}
\label{fig:ft}
\end{figure*}

\section{Heavy Fine-Tuning Hurts}

Although the foundation model has shown commendable performance in downstream long-tail learning tasks, it has indeed some limitations. As discussed earlier, CLIP has achieved high accuracy for head classes, but it has not yet achieved strong accuracy for tail classes. This naturally raises a question, is the fine-tuning not sufficient?

However, the results in \Cref{fig:deterioration} show that full fine-tuning yields improvements in head-class accuracy, while at the expense of a reduction in tail-class accuracy. Note that we have already optimized the balanced logit-adjusted (LA) loss \citep{menon2021longtail} and applied a balanced classifier initialization (will be introduced in \Cref{sec:sai}), but the prediction results are still biased. In fact, performance deterioration on tail classes makes the overall accuracy of full fine-tuning even worse than that of classifier fine-tuning (72.9\% vs. 73.5\% on ImageNet-LT, and 46.4\% vs. 48.5\% on Places-LT).

\begin{figure*}[!t]
\centering
\setlength{\tabcolsep}{0.2ex} 
\begin{subfigure}{0.42\linewidth}
    \includegraphics[width=0.42\linewidth, trim=0 0 6 0]{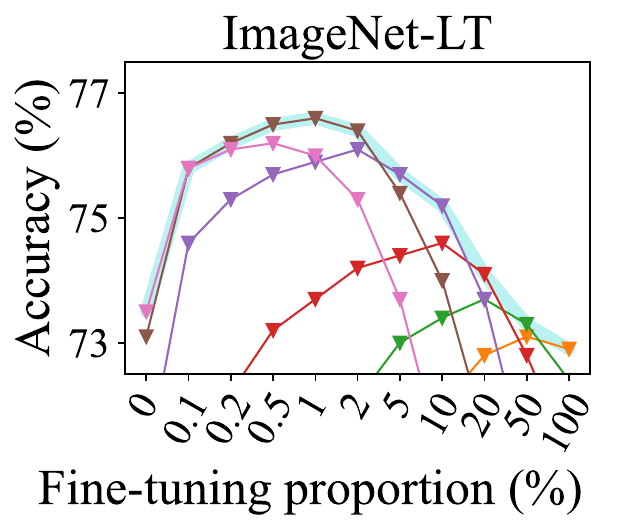}
    \includegraphics[width=0.42\linewidth, trim=6 0 0 0]{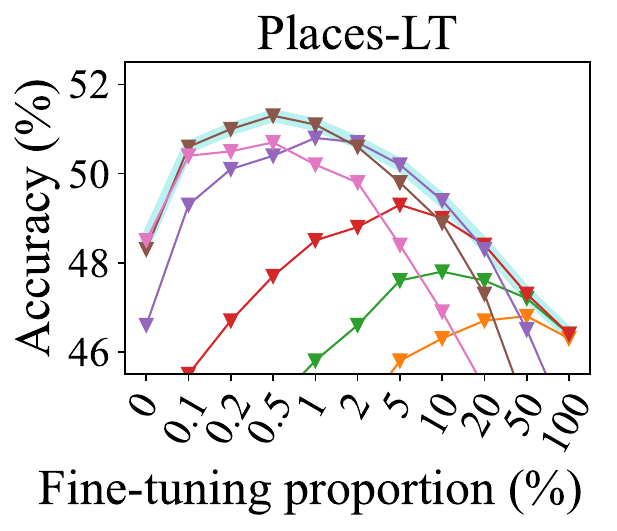}
    \includegraphics[width=0.14\linewidth, trim=12 -10
 12 0]{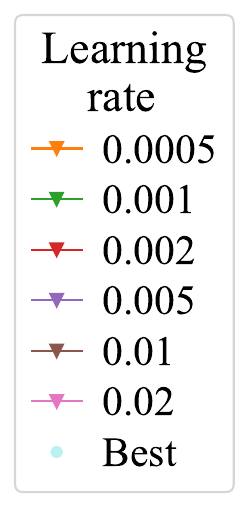}
    \caption{Fine-tuning with varying proportions.}
    \label{fig:ft-proportion}
\end{subfigure}
\hfill
\begin{subfigure}{0.28\linewidth}
    \begin{tabular}{cc}
    \begin{minipage}{0.51\linewidth}
        \centering
        \includegraphics[width=\linewidth]{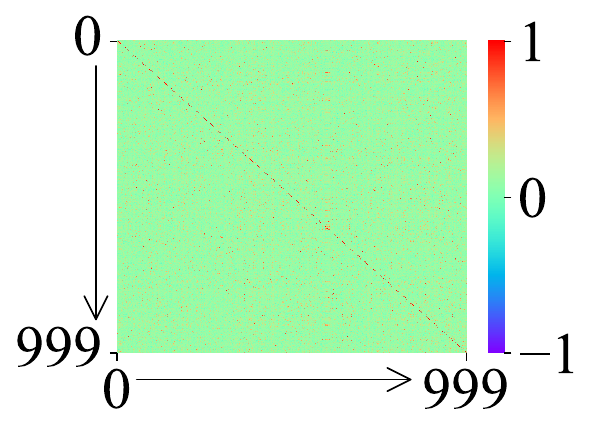}
    \end{minipage} &
    \begin{minipage}{0.47\linewidth}
        \centering
        \includegraphics[width=\linewidth]{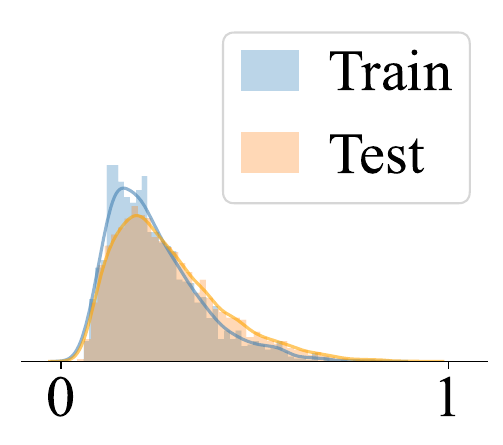}
    \end{minipage} \\
    \color{mplgreen}{\fontsize{8.5pt}{8.5pt}\selectfont Head acc.: 79.9\%} &
    \color{mplgreen}{\fontsize{8.5pt}{8.5pt}\selectfont Tail acc.: 70.2\%} \\
    \end{tabular}
    \caption{Arbitrary lightweight fine-tuning.}
    \label{fig:feature-alf}
\end{subfigure}
\begin{subfigure}{0.29\linewidth}
    \begin{tabular}{cc}
    \begin{minipage}{0.51\linewidth}
        \centering
        \includegraphics[width=\linewidth]{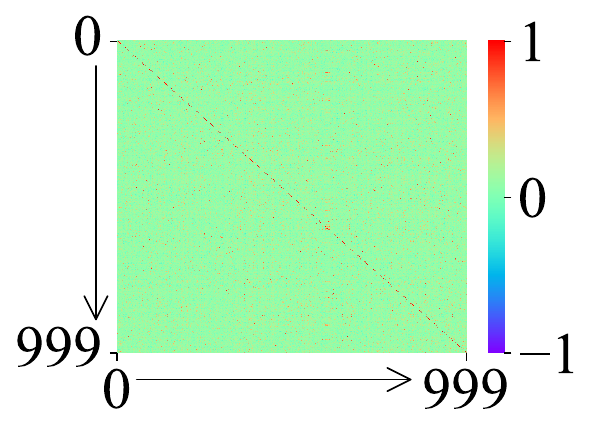}
    \end{minipage} &
    \begin{minipage}{0.47\linewidth}
        \centering
        \includegraphics[width=\linewidth]{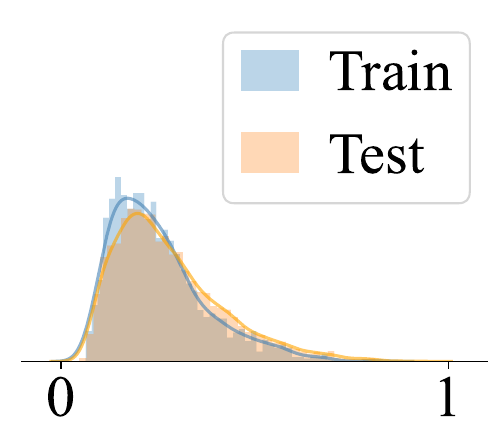}
    \end{minipage} \\
    \color{mplgreen}{\fontsize{8.5pt}{8.5pt}\selectfont Head acc.: 80.2\%} &
    \color{mplgreen}{\fontsize{8.5pt}{8.5pt}\selectfont Tail acc.: 71.5\%} \\
    \end{tabular}
    \caption{Structured lightweight fine-tuning.}
    \label{fig:feature-slf}
\end{subfigure}
\belowcaptionskip=-0.05in
\caption{(a) Fine-tuning a small proportion of all parameters (\eg 0.1\%-2\%) yields superior performance. As the proportion increases, performance deteriorates even when we search for the best learning rate. (b-c) Inter-class feature similarities (heatmaps) and intra-class distributions from tail classes (histograms) on ImageNet-LT. Both arbitrary and structured lightweight fine-tuning perform well in optimizing inter-class similarities and preserving intra-class distributions.}
\label{fig:lightweight}
\end{figure*}

To uncover the reasons behind it, we identify that full fine-tuning may distort the intra-class distance distributions. Specifically,
we calculate the inter-class feature similarities and the intra-class distance distributions on ImageNet-LT and report the results in \Cref{fig:ft}. Notably, \Cref{fig:feature-full} shows that full fine-tuning yields more discriminative representations as it reduces the inter-class similarities to approximately zero, which means the features of different classes are almost orthogonal. However, it also distorts the intra-class distribution, leading to a distribution shift between train and test data on tail classes. In this case, using the fine-tuned model to estimate the unknown tail-class data will inevitably lead to an underestimated class-conditional probability. We have conducted the following analysis on this matter.
\begin{proposition}
\label{prop:ccd}
Underestimated class-conditional probability $\oP(\vx\mid y=j)$ leads to an underestimated loss on class $j$ and a biased prediction towards other classes.
\end{proposition}
\begin{proof}
Denote $\oP_s$ and $\oP_t$ as the probability distribution in the source (training) and target (test) domain, respectively. For long-tail learning, $\oP_s(y)$ appears a long-tail distribution and $\oP_t(y)$ is a uniform distribution, \ie $\oP_t(y=k)\equiv 1/K$. We then have,
\begin{align}
&\oP_s(y=j\mid\vx)
= \cfrac{\oP_t(y=j\mid\vx)\cdot\cfrac{\oP_s(y=j\mid\vx)}{\oP_t(y=j\mid\vx)}}{\sum_{k\in[K]}\oP_t(y=k\mid\vx)\cdot\cfrac{\oP_s(y=k\mid\vx)}{\oP_t(y=k\mid\vx)}} \nonumber \\
&= \cfrac{\oP_t(y=j\mid\vx)\cdot\cfrac{\oP_s(\vx\mid y=j)}{\oP_t(\vx\mid y=j)}\cdot\cfrac{\oP_s(y=j)}{\oP_t(y=j)}}{\sum_{k\in[K]}\oP_t(y=k\mid\vx)\cdot\cfrac{\oP_s(\vx\mid y=k)}{\oP_t(\vx\mid y=k)}\cdot\cfrac{\oP_s(y=k)}{\oP_t(y=k)}} \nonumber \\
&= \cfrac{\oP_t(y=j\mid\vx)\cdot\oP_s(y=j)\cdot\zeta_{s-t}(j)}{\sum_{k\in[K]}\oP_t(y=k\mid\vx)\cdot\oP_s(y=k)\cdot\zeta_{s-t}(k)}
\end{align}
\begin{align}
&\gL(\vx,y=j)=-\log\oP_s(y=j\mid\vx) \nonumber \\
&=-\log\cfrac{\exp\left(z_j+\log\oP_s(y=j)+\log\zeta_{s-t}(j)\right)}{\sum_{k\in[K]}\exp\left(z_k+\log\oP_s(y=k)+\log\zeta_{s-t}(k)\right)}
\end{align}
where $\zeta_{s-t}(j)=\cfrac{\oP_s(\vx\mid y=j)}{\oP_t(\vx\mid y=j)}$ and $z_j$ is the predicted logit on class $j$.
For samples $\vx$ from class $j$, the underestimated $\oP_t(\vx\mid y=j)$ results in an underestimation of $\gL(\vx,y=j)$, and consequently leads to a biased optimization towards other classes. A more detailed proof is presented in \Cref{sec:la_loss}.
\end{proof}

\begin{remark}
\Cref{prop:ccd} indicates that the performance deterioration of full fine-tuning is attributed to the inconsistent class-conditional distributions among the tail classes. Previous works such as LA \citep{menon2021longtail} assume that the class-conditional distribution is consistent between the source and target domains, \ie $\zeta_{s-t}(j)=1$. However, our empirical findings in \Cref{fig:feature-full} reveal that full fine-tuning breaks this assumption. An ideal solution is to estimate the underlying class-conditional distribution; however, this is unfeasible due to the scarcity of tail-class data. Another approach is to prevent such distribution distortions. To achieve this goal, we introduce lightweight fine-tuning in \Cref{sec:method}.
\end{remark}

In addition to the aforementioned issue, full fine-tuning also tends to encounter severe overfitting on long-tail datasets, particularly on the tail classes.
We quantitatively assess the overfitting issue of full fine-tuning, and present the results in \Cref{sec:overfitting} due to the space constraint.
To mitigate the issue of performance deterioration on tail classes, recent approaches have explored two-stage training procedures \citep{ma2021simple} or the inclusion of additional training data \citep{tian2022vl,long2022retrieval}. However, these strategies often introduce significant training overhead or require external data, thereby limiting their practicality. In response to this, we introduce LIFT, an efficient and accurate lightweight fine-tuning framework tailored for long-tail learning.

\section{Efficient and Accurate Long-Tail Learning}
\label{sec:method}

\subsection{Lightweight Fine-Tuning Helps}
\label{sec:sparse}

To alleviate the distortion of intra-class distributions, a direct method is to constrain the number of learnable parameters. Formally, for each weight matrix $\mW\in\mathbb{R}^{d_1\times d_2}$ in the foundation model, we optimizes a specified proportion $\alpha$ of parameters within $\mW$, while keeping the rest frozen. Thus, only $\alpha d_1 d_2$ parameters are optimized. In practice, a sparse 0-1 mask $\mM\in \{0,1\}^{d_1\times d_2}$ is used to control the optimized parameters:
\begin{equation}
    \mX\mW \rightarrow \mX(\mW\circ \mM)+\underbrace{\mX(\mW\circ (1-\mM))}_{\textrm{gradient detached}}
\end{equation}
where $\Vert \mM\Vert_0=\alpha d_1 d_2$. In \Cref{fig:ft-proportion}, we investigate the impact of varying proportions of fine-tuned parameters. The result demonstrates the benefits of lightweight fine-tuning since only a small proportion (\eg 0.1\%) yields a substantial improvement. As the proportion increases, performance faces a risk of degradation, highlighting the drawbacks of heavy fine-tuning. Besides, heavy fine-tuning is sensitive to hyperparameters, as it requires searching for an optimal learning rate.

It is noteworthy that the optimized parameters are selected arbitrarily, but the performance improvement is remarkable. This indicates that lightweight fine-tuning is crucial for enhancing long-tail learning, even without specific fine-tuning strategies. 
To get a deeper understanding, we visualize the inter-class feature similarities and the intra-class distributions in \Cref{fig:feature-alf}. The results demonstrate that arbitrary lightweight fine-tuning yields comparable feature separability with full fine-tuning, while its class conditions are preserved to be consistent between training and test data. Furthermore, prediction accuracy verifies the adaptation and generalization ability of arbitrary lightweight fine-tuning, as it achieves the same high performance as full-fine-tuning on head classes, and its tail-class performance is even superior.

\subsection{The Proposed Fine-Tuning (\algo) Method}
We have demonstrated that arbitrary lightweight fine-tuning, even without any guidance for parameter selection, performs better than full fine-tuning. This indicates that a small learnable parameter quantity is more important for fine-tuning.
To this end, we first investigate \textit{structured lightweight fine-tuning} which learns a small set of task-specific parameters in a structured manner. Some related ideas, although already used in foundation models, are not directly suitable to long-tail learning. For more detailed introductions, please refer to \Cref{sec:peft_analysis}.

\Cref{fig:feature-slf} shows the potential of a very simple idea for structured lightweight fine-tuning.
It optimizes the inter-class similarities to be nearly orthogonal and preserves the intra-class distributions undistorted. As can be seen, its performance already surpasses arbitrary lightweight fine-tuning on both head and tail classes. The findings indicate that more well-designed structured lightweight fine-tuning will further enhance the feature separability as well as ensure the consistency of class-conditional distributions, and will be more accurate.

Our analysis above justifies that lightweight fine-tuning can significantly mitigate performance deterioration and improve generalization. To this end, we propose a low-complexity and accurate long-tail learning method termed \textit{LIghtweight Fine-Tuning} (\algo).
\algo\ is versatile and inclusive, allowing for the incorporation of a range of structured lightweight fine-tuning methods.

In order to have better adaptability among tasks, apart from the lightweight fine-tuning, a classifier is introduced to discern and select the features for different tasks. Without loss of generality, the linear classifier is employed due to its simplicity and versatility. Specifically, given a feature vector $\vf$, the predicted logit for class $j$ is computed as $z_j=\vw_j^{\top}\vf+b$. Note that when training with long-tail data, the norms of classifier weight $\vw_j$ tend to exhibit an imbalanced distribution, which can lead to biased predictions \citep{kang2020decoupling,wei2021towards}. To overcome it and draw on the strengths of CLIP which optimizes cosine distances within its feature space, inspired by \citet{wei2021towards},
we propose to use a enhanced cosine classifier $z_j=\sigma\cdot\frac{\vw_j^{\top}\vf}{\Vert \vw_j\Vert_2 \Vert \vf\Vert_2}$.
Here, $\sigma$ is a scaling factor, and the biased prediction is alleviated by dividing the classifier norm. To further improve the effectiveness of \algo, we opt for the LA loss for optimization:
\begin{equation}
    \gL_{\text{LA}}(\vx,y=j)=-\log\frac{\exp({z_j}+\log \oP(y=j))}{\sum_{k\in[K]}\exp({z_k}+\log \oP(y=k))}
\end{equation}
Here, $y=j$ represents the ground-truth label of $\vx$ and $z_j$ is the predicted logit. $\oP(y=j)$ signifies the class prior probability, which can be estimated based on the training data. More theoretical understanding of the LA loss can be found in \Cref{sec:la_loss}.

\subsection{Semantic-Aware Initialization}
\label{sec:sai}
Note that an uninitialized classifier is discovered to have a negative impact on fine-tuning the model \citep{kumar2022finetuning}. Therefore, it is crucial to set an appropriate initial state for the classifier. A straightforward method is to apply linear probing using re-weighted or LA loss. Another approach is to compute the class mean feature as initialization. However, these two approaches not only require extracting features of training data but also are not available with scarce tail-class data.
To overcome it, we tend to leverage the semantic knowledge from the text modality of CLIP.
Specifically, we generate hand-crafted textual prompts (\eg ``\texttt{a photo of a [CLASS].}'') and compute their features $\vf_{T_1},\cdots,\vf_{T_K}$, which are then employed to initialize the classfier weights $\vw_1,\cdots,\vw_K$. 
We call this way as \textit{semantic-aware initialization} (SAI).
Unlike prior methods that fine-tune both the image encoder and the text encoder in optimization processes \citep{ma2021simple,tian2022vl}, SAI relies on a single forward pass of the text encoder for each class description. After that, the text encoder is discarded. This simple approach allows us to achieve a better initial state of the classifier with small computational overhead.

\subsection{Test-Time Ensembling}
It is well-established that applying random perturbations to each input can lead to improved generalization \citep{sun2020test,wang2021tent,zhou2023ods}. This principle may also be useful for the Transformer-based foundation model, where an image is divided into multiple patches, potentially resulting in the segmentation of continuous patterns into different patches. To further enhance the generalization robustness, we propose to aggregate the predictions from a set of perturbed versions of the input. Formally, given a test data point $\boldsymbol{x}$, its predicted logits $\vz$ are obtained by averaging the predictions from $M$ perturbed versions:
\begin{equation}
\vz=\log\oP(\vy\mid \vx)=\frac{1}{M} \sum_{i=1}^M \log\oP(\vy \mid \alpha_{i}(\vx))
\end{equation}
Here, $\alpha_i(\boldsymbol{x})$ represents different perturbed versions of $\boldsymbol{x}$.
This approach helps mitigate bias introduced by image cropping.
Due to the page limit, we present the algorithm procedure in \Cref{sec:tte}.
We term this technique \textit{test-time ensembling} (TTE), and 
it can be integrated with small computational overhead to enhance performance.

\begin{table*}[!t]
\caption{Comparison with state-of-the-art methods on ImageNet-LT.}
\label{table:comp_imagenetlt}
\setlength{\tabcolsep}{1.2ex} 
\centering
\begin{small}
\begin{tabular}{l|c|c|c|cccc}
\toprule
\bf Methods &\bf Backbone &\makecell{\bf Learnable \\ \bf Params.} &\bf \#Epochs &\bf Overall &\bf Head &\bf Medium &\bf Tail \\
\midrule
\multicolumn{5}{l}{\bf Training from scratch} \\
\midrule
cRT \citep{kang2020decoupling} & ResNet-50 & 23.51M & 90+10 & 47.3 & 58.8 & 44.0 & 26.1 \\
LWS \citep{kang2020decoupling} & ResNet-50 & 23.51M & 90+10 & 47.7 & 57.1 & 45.2 & 29.3 \\
MiSLAS \citep{zhong2021improving} & ResNet-50 & 23.51M & 180+10 & 52.7 & 62.9 & 50.7 & 34.3 \\
LA \citep{menon2021longtail} & ResNet-50 & 23.51M & 90 & 51.1 & - & - & - \\
DisAlign \citep{zhang2021distribution} & ResNet-50 & 23.51M & 90 & 52.9 & 61.3 & 52.2 & 31.4 \\
BCL \citep{zhu2022balanced} & ResNet-50 & 23.51M & 100 & 56.0 & - & - & - \\
PaCo \citep{cui2021parametric} & ResNet-50 & 23.51M & 400 & 57.0 & - & - & - \\
NCL \citep{li2022nested} & ResNet-50 & 23.51M & 400 & 57.4 & - & - & -  \\
LiVT \citep{xu2023learning} & ViT-B/16 & 85.80M & 100 & 60.9 & 73.6 & 56.4 & 41.0 \\
\midrule
\multicolumn{8}{l}{\bf Fine-tuning foundation model} \\
\midrule
BALLAD \citep{ma2021simple} & ViT-B/16 & 149.62M & 50+10 & 75.7 & 79.1 & 74.5 & 69.8 \\
Decoder \citep{wang2023exploring} & ViT-B/16 & 21.26M & $\sim$18 & 73.2 & - & - & - \\
\algo\ (Ours) & ViT-B/16 &\bf 0.62M & \bf 10 & 77.0 & 80.2 & 76.1 & 71.5 \\
\algo\ w/ TTE (Ours) & ViT-B/16 &\bf 0.62M & \bf 10 &\bf 78.3 &\bf 81.3 &\bf 77.4 &\bf 73.4 \\
\midrule
\multicolumn{8}{l}{\gray\bf Fine-tuning with extra data} \\
\midrule
\gray VL-LTR \citep{tian2022vl} &\gray ViT-B/16 &\gray 149.62M &\gray 100 &\gray 77.2 &\gray 84.5 &\gray 74.6 &\gray 59.3 \\
\gray GML \citep{suh2023long} &\gray ViT-B/16 &\gray 149.62M &\gray 100 &\gray 78.0 &\gray - &\gray - &\gray - \\
\bottomrule
\end{tabular}
\end{small}
\end{table*}

\begin{table*}[!t]
\caption{Comparison with state-of-the-art methods on Places-LT.}
\label{table:comp_placeslt}
\setlength{\tabcolsep}{1.2ex} 
\centering
\begin{small}
\begin{tabular}{l|c|c|c|cccc}
\toprule
\bf Methods &\bf Backbone &\makecell{\bf Learnable \\ \bf Params.} &\bf \#Epochs &\bf Overall &\bf Head &\bf Medium &\bf Tail\\ 
\midrule
\multicolumn{8}{l}{\bf Training from scratch (with an ImageNet-1K pre-trained backbone)} \\
\midrule
OLTR \citep{liu2019large} & ResNet-152 & 58.14M & 30 & 35.9 & 44.7 & 37.0 & 25.3 \\
cRT \citep{kang2020decoupling} & ResNet-152 & 58.14M & 90+10 & 36.7 & 42.0 & 37.6 & 24.9 \\
LWS \citep{kang2020decoupling} & ResNet-152 & 58.14M & 90+10 & 37.6 & 40.6 & 39.1 & 28.6 \\
MiSLAS \citep{zhong2021improving} & ResNet-152 & 58.14M & 90+10 & 40.4 & 39.6 & 43.3 & 36.1 \\
DisAlign \citep{zhang2021distribution} & ResNet-152 & 58.14M & 30 & 39.3 & 40.4 & 42.4 & 30.1 \\
ALA \citep{zhao2022adaptive} & ResNet-152 & 58.14M & 30 & 40.1 & 43.9 & 40.1 & 32.9 \\
PaCo \citep{cui2021parametric} & ResNet-152 & 58.14M & 30 & 41.2 & 36.1 & 47.9 & 35.3  \\
LiVT \citep{xu2023learning} & ViT-B/16 & 85.80M & 100 & 40.8 & 48.1 & 40.6 & 27.5 \\
\midrule
\multicolumn{8}{l}{\bf Fine-tuning foundation model} \\
\midrule
BALLAD \citep{ma2021simple} & ViT-B/16 & 149.62M & 50+10 & 49.5 & 49.3 & 50.2 & 48.4 \\
Decoder \citep{wang2023exploring} & ViT-B/16 & 21.26M & $\sim$34 & 46.8 & - & - & - \\
LPT \citep{dong2023lpt} & ViT-B/16 & 1.01M & 40+40 & 50.1 & 49.3 & 52.3 & 46.9 \\
\algo\ (Ours) &  ViT-B/16 &\bf 0.18M &\bf 10 & 51.5 & 51.3 & 52.2 & 50.5 \\
\algo\ w/ TTE (Ours) &  ViT-B/16 &\bf 0.18M &\bf 10 &\bf 52.2 &\bf 51.7 &\bf 53.1 &\bf 50.9 \\
\midrule
\multicolumn{8}{l}{\gray\bf Fine-tuning with extra data} \\
\midrule
\gray VL-LTR \citep{tian2022vl} &\gray ViT-B/16 &\gray 149.62M &\gray 100 &\gray 50.1 &\gray 54.2 &\gray 48.5 &\gray 42.0 \\
\gray RAC \citep{long2022retrieval} &\gray ViT-B/16 &\gray 85.80M &\gray 30 &\gray 47.2 &\gray 48.7 &\gray 48.3 &\gray 41.8 \\
\bottomrule
\end{tabular}
\end{small}
\end{table*}

\begin{table*}[!t]
\caption{Comparison with state-of-the-art methods on iNaturalist 2018.}
\label{table:comp_inat18}
\setlength{\tabcolsep}{1.2ex} 
\centering
\begin{small}
\begin{tabular}{l|c|c|c|cccc}
\toprule
\bf Methods &\bf Backbone &\makecell{\bf Learnable \\ \bf Params.} &\bf \#Epochs &\bf Overall &\bf Head &\bf Medium &\bf Tail \\
\midrule
\multicolumn{8}{l}{\bf Training from scratch} \\
\midrule
cRT \citep{kang2020decoupling} & ResNet-50 & 23.51M & 90+10 & 65.2 & 69.0 & 66.0 & 63.2 \\
LWS \citep{kang2020decoupling} & ResNet-50 & 23.51M & 90+10 & 65.9 & 65.0 & 66.3 & 65.5 \\
MiSLAS \citep{zhong2021improving} & ResNet-50 & 23.51M & 200+30 & 71.6 & 73.2 & 72.4 & 70.4 \\
DiVE \citep{he2021distilling} & ResNet-50 & 23.51M & 90 & 69.1 & 70.6 & 70.0 & 67.6 \\
DisAlign \citep{zhang2021distribution} & ResNet-50 & 23.51M & 90 & 69.5 & 61.6 & 70.8 & 69.9 \\
ALA \citep{zhao2022adaptive} & ResNet-50 & 23.51M & 90 & 70.7 & 71.3 & 70.8 & 70.4 \\
RIDE \citep{wang2021longtailed} & ResNet-50 & 23.51M & 100 & 72.6 & 70.9 & 72.4 & 73.1 \\
RIDE+CR \citep{ma2023curvature} & ResNet-50 & 23.51M & 200 & 73.5 & 71.0 & 73.8 & 74.3 \\
RIDE+OTmix \citep{gao2023enhancing} & ResNet-50 & 23.51M & 210 & 73.0 & 71.3 & 72.8 & 73.8 \\
BCL \citep{zhu2022balanced} & ResNet-50 & 23.51M & 100 & 71.8 & - & - & - \\
PaCo \citep{cui2021parametric} & ResNet-50 & 23.51M & 400 & 73.2 & 70.4 & 72.8 & 73.6 \\
NCL \citep{li2022nested} & ResNet-50 & 23.51M & 400 & 74.2 & 72.0 & 74.9 & 73.8 \\
GML \citep{suh2023long} & ResNet-50 & 23.51M & 400 & 74.5 & - & - & - \\
LiVT \citep{xu2023learning} & ViT-B/16 & 85.80M & 100 & 76.1 &\bf 78.9 & 76.5 & 74.8 \\
\midrule
\multicolumn{8}{l}{\bf Fine-tuning foundation model} \\
\midrule
Decoder \citep{wang2023exploring} & ViT-B/16 & 21.26M &\bf $\sim$5 & 59.2 & - & - & - \\
LPT \citep{dong2023lpt} & ViT-B/16 &\bf 1.01M & 80+80 & 76.1 & - & - & 79.3 \\
\algo\ (Ours) & ViT-B/16 & 4.75M & 20 & 79.1 & 72.4 & 79.0 & 81.1 \\
\algo\ w/ TTE (Ours) & ViT-B/16 & 4.75M & 20 &\bf 80.4 & 74.0 &\bf 80.3 &\bf 82.2 \\
\midrule
\multicolumn{8}{l}{\gray\bf Fine-tuning with extra data} \\
\midrule
\gray VL-LTR \citep{tian2022vl} &\gray ViT-B/16 &\gray 149.62M &\gray 100 &\gray 76.8 &\gray - &\gray - &\gray - \\
\gray RAC \citep{long2022retrieval} &\gray ViT-B/16 &\gray 85.80M &\gray 20 &\gray 80.2 &\gray 75.9 &\gray 80.5 &\gray 81.1 \\
\bottomrule
\end{tabular}
\end{small}
\end{table*}

\section{Empirical Study}

\subsection{Experimental Settings}
\label{sec:exp_settings}
We conduct experiments on four long-tail datasets, including ImageNet-LT \citep{liu2019large}, Places-LT \citep{liu2019large}, iNaturalist 2018 \citep{van2018inaturalist} and CIFAR-100-LT \citep{cao2019learning}.
ImageNet-LT has 115.8K images from 1000 classes, with a maximum of 1280 and a minimum of 5 images per class.
Places-LT contains 62.5K images from 365 classes, from a maximal 4980 to a minimum of 5 images per class.
iNaturalist 2018 consists of 437.5K images distributed across 8142 species, with the number of images per species varying from as few as 2 to as many as 1000.
CIFAR-100-LT is constructed with various imbalance ratios, including 100, 50, and 10.
In addition to measuring overall accuracy, we adhere to the evaluation protocol introduced by \citet{liu2019large} to report accuracy across three splits of classes: head classes ($>$100 images), medium classes (20$\sim$100 images), and tail classes ($<$20 images).

\subsection{Comparison with State-of-the-art Methods}
\paragraph{Results on ImageNet-LT.} We report the test accuracy in \Cref{table:comp_imagenetlt}. While existing approaches such as VL-LTR \citep{tian2022vl} and GML \citep{suh2023long} rely on extensive auxiliary data to facilitate fine-tuning, our method \algo\ achieves superior performance by leveraging the test-time ensembling (TTE) technique alone. The use of external data not only incurs significant computational overhead but also reduces practicality due to the unavailability of such data in many real-world applications. The advantage of \algo\ is more significant compared with methods that do not use auxiliary data, \ie\ \algo\ surpasses the previous best method by 1.3\% in accuracy. Importantly, \algo\ only needs 10 epochs of training and fine-tunes far fewer model parameters (\ie from 21.26M to 0.62M). It is worth noting that we do not include LPT \citep{dong2023lpt} for comparison since it is pre-trained on ImageNet-21K, its results on ImageNet-LT were not reported in the original paper.

\paragraph{Results on Places-LT.} \Cref{table:comp_placeslt} shows that \algo\ outperforms existing methods by a larger margin on Places-LT. Even without TTE, \algo\ surpasses VL-LTR and RAC which use external training data by 1.4\% in accuracy. By integrating TTE, the number increases to 2.1\%. Similar to ImageNet-LT, we only need 10 epochs of training in contrast to 80 epochs (40 in each stage) for LPT. The amount of tunable parameters is also much fewer, \ie 0.18M. Nevertheless, \algo\ outperforms LPT by 1.4\% in accuracy. When taking a closer look, we can see that \algo\ significantly improves the tail class performance, \ie from 46.9 to 50.5.

\begin{table*}[!h]
\caption{Comparison with state-of-the-art methods on CIFAR-100-LT with various imbalance ratios. $^\dagger$Pre-trained model from ImageNet-21K\protect\footnotemark[1] has several classes related to CIFAR-100\protect\footnotemark[2], which potentially leads to data leakage.}
\label{table:comp_cifar100lt}
\setlength{\tabcolsep}{1.2ex} 
\centering
\begin{small}
\begin{tabular}{l|c|c|c| C{8ex} C{8ex} C{8ex} }
\toprule
\multirow{2}{*}{\bf Methods} & \multirow{2}{*}{\bf Backbone} &\multirow{2}{*}{\makecell{\bf Learnable \\ \bf Params.}} & \multirow{2}{*}{\bf \#Epochs} & \multicolumn{3}{c}{\bf Imbalance Ratio} \\ 
& & & & \bf 100 &\bf 50 &\bf 10 \\
\midrule
\multicolumn{5}{l}{\bf Training from scratch} \\
\midrule
LDAM \citep{cao2019learning} & ResNet-32 & 0.46M & 200 & 42.0 & 46.6 & 58.7 \\
BBN \citep{zhou2020bbn} & ResNet-32 & 0.46M & 200 & 42.6 & 47.0 & 59.1 \\
DiVE \citep{he2021distilling} & ResNet-32 & 0.46M & 200 & 45.4 & 51.1 & 62.0 \\
MiSLAS \citep{zhong2021improving} & ResNet-32 & 0.46M & 200+10 & 47.0 & 52.3 & 63.2 \\
BS \citep{ren2020balanced} & ResNet-32 & 0.46M & 400 & 50.8 & 54.2 & 63.0 \\
PaCo \citep{cui2021parametric} & ResNet-32 & 0.46M & 400 & 52.0 & 56.0 & 64.2 \\
BCL \citep{zhu2022balanced} & ResNet-32 & 0.46M & 200 & 51.9 & 56.6 & 64.9 \\
\midrule
\multicolumn{5}{l}{\bf Fine-tuning pre-trained model} \\
\midrule
LiVT \citep{xu2023learning} & ViT-B/16 & 85.80M & 100 & 58.2 & - & 69.2 \\
BALLAD \citep{ma2021simple} & ViT-B/16 & 149.62M & 50+10 & 77.8 & - & - \\
\algo\ (Ours) & ViT-B/16 &\bf 0.10M &\bf 10 & 80.3 & 82.0 & 83.8 \\
\algo\ w/ TTE (Ours)  & ViT-B/16 &\bf 0.10M &\bf 10 &\bf 81.7 &\bf 83.1 &\bf 84.9 \\
\midrule
\multicolumn{7}{l}{\gray\bf Fine-tuning pre-trained model from ImageNet-21K$^\dagger$} \\
\midrule
\gray LPT \citep{dong2023lpt} &\gray ViT-B/16 &\gray 1.01M &\gray 40+40 &\textcolor{gray}{\bf 89.1} &\textcolor{gray}{90.0} &\textcolor{gray}{91.0} \\
\gray\algo\ (Ours) &\gray ViT-B/16 &\gray\bf 0.10M &\gray\bf 10 &\textcolor{gray}{\bf 89.1} &\textcolor{gray}{\bf 90.2} &\textcolor{gray}{\bf 91.3} \\
\bottomrule
\end{tabular}
\end{small}
\end{table*}

\begin{figure*}[!t]
    \centering
    \begin{subfigure}{0.49\linewidth}
        \includegraphics[width=0.49\linewidth]{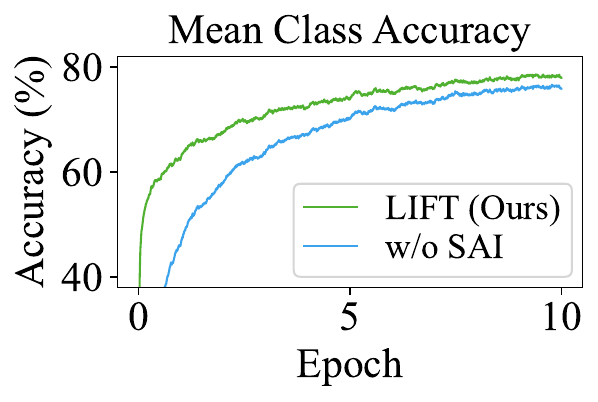}
        \includegraphics[width=0.49\linewidth]{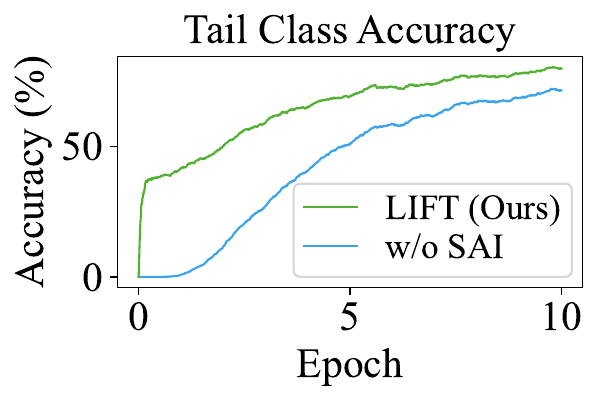}
        \caption{ImageNet-LT.}
    \end{subfigure}
    \hfill
    \begin{subfigure}{0.49\linewidth}
        \includegraphics[width=0.49\linewidth]{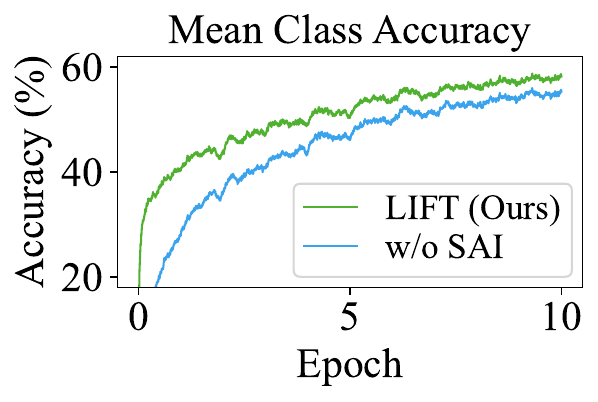}
        \includegraphics[width=0.49\linewidth]{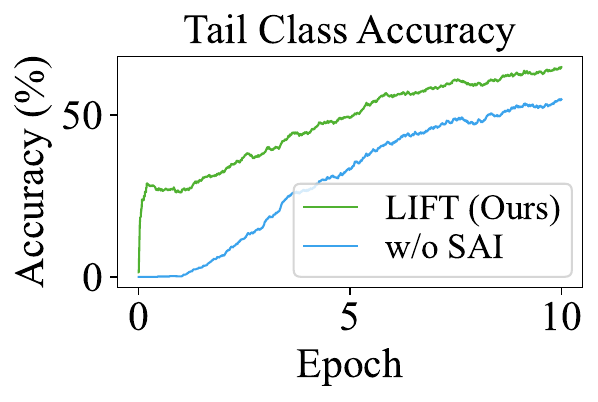}
        \caption{Places-LT.}
    \end{subfigure}
    \belowcaptionskip=-0.05in
    \caption{Convergence curve of mean class and tail class training accuracy.}
    \label{fig:convergence}
\end{figure*}

\begin{table*}[!t]
\caption{Ablation study of each key component in \algo. The baseline involves learning a cosine classifier using CE loss.}
\label{table:ablation}
\setlength{\tabcolsep}{1.2ex} 
\centering
\begin{small}
\begin{tabular}{C{4.3ex} C{4.3ex} C{4.3ex} C{4.3ex} | C{6ex} C{6ex} C{6ex} C{6ex} | C{6ex} C{6ex} C{6ex} C{6ex}}
\toprule
\multirow{2}{*}{\bf SLF} & \multirow{2}{*}{\bf LA} & \multirow{2}{*}{\bf SAI} & \multirow{2}{*}{\bf TTE} & \multicolumn{4}{c|}{\bf ImageNet-LT} & \multicolumn{4}{c}{\bf Places-LT} \\ 
& & & &\bf Overall &\bf Head &\bf Medium &\bf Tail &\bf Overall &\bf Head &\bf Medium &\bf Tail \\
\midrule
& & & & 60.9 & 82.6 & 56.7 & 13.8 & 34.3 & 53.6 & 30.5 & 7.5 \\
$\checkmark$ & & & & 68.9 &\bf 84.7 & 66.3 & 33.7 & 38.9 &\bf 55.6 & 35.3 & 16.4 \\
$\checkmark$ & $\checkmark$ & & & 74.9 & 79.7 & 74.7 & 61.7 & 48.7 & 50.6 & 50.8 & 40.5 \\
$\checkmark$ & $\checkmark$ & $\checkmark$ & & 77.0 & 80.2 & 76.1 & 71.5 & 51.5 & 51.3 & 52.2 & 50.5 \\
$\checkmark$ & $\checkmark$ & $\checkmark$ & $\checkmark$ &\bf 78.3 & 81.3 &\bf 77.4 &\bf 73.4 &\bf 52.2 & 51.7 &\bf 53.1 &\bf 50.9 \\
\bottomrule
\end{tabular}
\end{small}
\end{table*}

\paragraph{Results on iNaturalist 2018.} We report the results in \Cref{table:comp_inat18}. Overall, our method achieves the best performance on this challenging dataset, surpassing LPT, VL-LTR, and RAC. We acknowledge that Decoder \citep{wang2023exploring} uses fewer training epochs, however, its performance trails far behind \algo. Particularly, \algo\ (without using TTE) improves LPT by 3\% in accuracy and \algo\ only needs 20 epochs of training compared with 160 epochs (80 per stage) for LPT. Although LPT uses fewer learnable parameters, we can reduce the parameters of \algo\ to reach a comparable quantity (\ie reduce the bottleneck dimension $r$ to 64, more details are given in \Cref{fig:bottle_dim}). In this case, \algo\ achieves an accuracy of 77.7\% (without TTE) / 79.0\% (with TTE), which still outperforms LPT. In fact, due to the large number of classes of iNaturalist 2018, the classifier already contains 6.25M parameters. Therefore, the parameter quantity of \algo\ does not lead to too much overhead.

\paragraph{Results on CIFAR-100-LT.}

The results in \Cref{table:comp_cifar100lt} demonstrate that \algo\ outperforms other methods, including LiVT, BALLAD, and various training-from-scratch approaches. These results hold regardless of whether TTE is applied or not. Additionally, we extend our experiments by replacing the CLIP with the pre-trained model from ImageNet-21K. In this case, we employ the class mean features to initialize the classifier due to the lack of a corresponding text encoder. Despite the inherent class overlaps between ImageNet-21K and CIFAR-100, which naturally lead to higher performance, our method surpasses LPT with fewer training epochs and learnable parameters.

\subsection{More Advantages and Ablation Studies}

\paragraph{\algo\ Improves Convergence.}
\Cref{fig:convergence} presents the mean class and tail class training accuracy as a function of epochs. Overall, we can observe that \algo\ converges rapidly with 10 training epochs on both datasets. As expected, semantic-aware initialization (SAI) attributes to fast convergence, especially in the case of the tail classes.

\paragraph{Ablation Studies.}
To assess the impact of each component, we conduct a systematical ablation study on different components in \algo\, including 1) the structured lightweight fine-tuning (SLF) module, 2) the logit-adjusted (LA) loss, 3) semantic-aware initialization (SAI), and 4) test-time ensembling (TTE). The results presented in \Cref{table:ablation} demonstrate the effectiveness of each component. Specifically, 1) SLF enhances performance on both head and tail classes; 2) without the LA loss, predictions tend to be biased to head classes; 3) SAI consistently improves the generalization, particularly on tail classes; 4) TTE further boosts the performance across both head and tail classes.

\footnotetext[1]{\footnotesize\imageneturl}
\footnotetext[2]{\footnotesize\cifarurl}

\begin{table*}[!t]
\caption{Comparison of different fine-tuning methods. All methods use SAI (if have classifier) and TTE for fair comparison.}
\label{table:comp_peft_module}
\setlength{\tabcolsep}{1.2ex} 
\centering
\begin{small}
\begin{tabular}{L{10.5ex} L{12.5ex} | C{6ex} C{6ex} C{6ex} C{6ex} | C{6ex} C{6ex} C{6ex} C{6ex}}
\toprule
\multicolumn{2}{l|}{\multirow{2}{*}{\bf Methods}} & \multicolumn{4}{c|}{\bf ImageNet-LT} & \multicolumn{4}{c}{\bf Places-LT} \\ 
\multicolumn{2}{l|}{}&\bf Overall &\bf Head &\bf Medium &\bf Tail &\bf Overall &\bf Head &\bf Medium &\bf Tail \\
\midrule
\multicolumn{2}{l|}{Zero-shot CLIP} & 68.3 & 69.2 & 67.6 & 67.7 & 40.2 & 38.3 & 39.2 & 45.9 \\
\multicolumn{2}{l|}{Classifier fine-tuning} & 74.4 & 77.6 & 73.7 & 68.0 & 49.4 & 49.1 & 50.3 & 48.2 \\
\multicolumn{2}{l|}{Full fine-tuning} & 74.4 &\bf 82.2 & 73.9 & 53.8 & 47.2 & 51.6 & 48.5 & 36.2 \\
\multicolumn{2}{l|}{\algo\ (Arbitrary)} & 77.8 & 80.9 & 76.9 & 72.2 & 51.9 & 51.3 & 52.7 &\bf 51.1 \\
\arrayrulecolor{gray}\midrule\arrayrulecolor{black}
\multirow{7}{*}{\makecell[l]{\algo\\(Structured)}} & BitFit & 77.0 & 79.7 & 76.3 & 71.9 & 51.5 & 51.2 & 52.3 & 50.0 \\
& VPT-shallow & 75.2 & 78.8 & 74.8 & 66.8 & 49.9 & 50.5 & 51.4 & 45.3 \\
& VPT-deep & 77.2 & 79.5 & 76.5 & 72.8 & 51.5 & 51.4 & 52.3 & 49.8 \\
& Adapter & 78.1 & 81.3 & 77.1 & 72.8 & 52.0 &\bf 51.7 & 52.7 & 51.0 \\
& LoRA & 76.9 & 79.6 & 76.2 & 71.7 & 51.8 & 51.5 & 52.5 & 50.5 \\ 
& AdaptFormer &\bf 78.3 & 81.3 &\bf 77.4 &\bf 73.4 &\bf 52.2 &\bf 51.7 &\bf 53.1 & 50.9 \\
\bottomrule
\end{tabular}
\end{small}
\end{table*}

\begin{table*}[!t]
\caption{Comparison of \algo\ with different classifier initialization methods.}
\label{table:comp_clf_init}
\setlength{\tabcolsep}{1.2ex} 
\centering
\begin{small}
\begin{tabular}{l| C{6ex} C{6ex} C{6ex} C{6ex} | C{6ex} C{6ex} C{6ex} C{6ex}}
\toprule
\multirow{2}{*}{\bf Methods} & \multicolumn{4}{c|}{\bf ImageNet-LT} & \multicolumn{4}{c}{\bf Places-LT} \\ 
&\bf Overall &\bf Head &\bf Medium &\bf Tail &\bf Overall &\bf Head &\bf Medium &\bf Tail \\
\midrule
Random initialization & 76.1 & 80.8 & 75.9 & 63.2 & 49.6 & 51.2 & 52.1 & 41.0 \\
Linear probing (re-weighted) & 77.1 &\bf 81.8 & 76.4 & 66.6 & 49.9 & 51.2 & 51.0 & 45.0 \\
Linear probing (w/ LA loss) & 77.2 & 81.3 & 76.4 & 68.4 & 50.2 & 51.3 & 51.1 & 46.2 \\
Class mean features & 77.5 & 81.3 & 76.8 & 69.4 & 51.3 & 51.6 & 52.1 & 48.8 \\
Semantic-aware initialization &\bf 78.3 & 81.3 &\bf 77.4 &\bf 73.4 &\bf 52.2 &\bf 51.7 &\bf 53.1 &\bf 50.9 \\
\bottomrule
\end{tabular}
\end{small}
\end{table*}

\paragraph{Combined with Varying Fine-Tuning Methods.} \algo\ is a general framework in which many lightweight fine-tuning methods can be integrated. In addition to zero-shot CLIP, classifier fine-tuning, and full fine-tuning, we test \algo\ with arbitrary lightweight fine-tuning and other 6 structured lightweight methods, including
\textit{BitFit} \citep{zaken2022bitfit}, 
\textit{VPT-shallow} \citep{jia2022visual}, 
\textit{VPT-deep} \citep{jia2022visual}, 
\textit{Adapter} \citep{houlsby2019parameter}, 
\textit{LoRA} \citep{hu2022lora}, 
and \textit{AdaptFormer} \citep{chen2022adaptformer}.
The results in \Cref{table:comp_peft_module} demonstrate that arbitrary lightweight fine-tuning surpasses the baseline methods by a large margin across both overall and tail-class performance. This underscores the efficacy of lightweight fine-tuning even in the absence of specific strategies. Furthermore, the integration of all structured methods leads to improved performance. Specifically, AdaptFormer performs best on both datasets and Adapter achieves slightly low accuracy (\ie 0.2\%).

\paragraph{Effect of Semantic-Aware Initialization.} In \Cref{table:comp_clf_init}, we test three kinds of classifier initialization strategies in comparison with the random initialization baseline. In \algo, we use the textual feature by default because it transfers semantic relations between classes during fine-tuning. We also compare different textual prompting methods and report the results in \Cref{sec:textual_prompts}. The other strategies using linear probing or class mean features to initialize the classifier achieve slightly poor performance but still significantly improve the baseline. This experiment shows that a good starting point for parameter optimization can lead to a better solution.

\subsection{More Supporting Experiments} 

Due to the page limitation, we report additional results
in \Cref{sec:addtional}, including 1) comparison of different training epochs; 2) comparison of different classifiers; 3) comparison of different losses; 4) more detailed observations on model convergence; 5) \algo\ with variant backbones.

\section{Conclusion}
This paper studies long-tail learning with the foundation model. We analyze the issue of heavy fine-tuning in distorting tail-class performance and discover that even an arbitrary lightweight fine-tuning yields significant improvement. Moreover, we propose a versatile framework \algo\ to facilitate lightweight fine-tuning for long-tail learning. \algo\ notably achieves convergence in fewer than 20 training epochs without the need for any external data, and consistently outperforms numerous baseline methods across a range of long-tail datasets, including CIFAR-100-LT, ImageNet-LT, Places-LT, and iNaturalist 2018. We emphasize the ease of training and hope that our approach serves as an inspiration for further advancements in the field of long-tail learning.

\section*{Code Availability Statement}
The implementation code for this work is available at \url{https://github.com/shijxcs/LIFT}.

\section*{Acknowledgements}
This research was supported by the National Key R\&D Program of China (2022YFC3340901), the National Science Foundation of China (62176118).

\section*{Impact Statement}
This paper aims to advance the field of long-tail learning. In critical and high-stakes applications such as autonomous driving, medical image diagnosis, and industrial anomaly detection, the presence of long-tail data poses the risk of producing biased predictions. Despite the strong capacity of the foundation model, its deployment may result in non-negligible performance degradation if not utilized properly. By shedding light on this problem, we aim to inspire more research on efficient and accurate long-tail learning algorithms with foundation models.

\nocite{langley00}

\bibliography{reference}
\bibliographystyle{icml2024}

\newpage
\appendix
\onecolumn

\section{Implementation Details}
\label{sec:imp_details}

For all experiments, we use the SGD optimizer with a batch size of 128, weight decay of $5\times 10^{-4}$, and momentum of 0.9. For lightweight fine-tuning methods, the learning rate is 0.01. For full fine-tuning, we search the learning rate from \{0.02, 0.01, 0.005, 0.002, 0.001, 0.0005\} considering its weak stability. For ImageNet-LT, Places-LT, and CIFAR-100-LT, we train the model for only 10 epochs; and for iNaturalist 2018, we train 20 epochs considering that it has much more data. In \algo, we set the bottleneck dimension $r=2^{\lfloor\log_{2}{(\frac{K}{2L})}\rfloor}$ for Adapter and AdaptFormer such that it learns even fewer parameters than the classifier (please refer to \Cref{sec:peft_analysis} for detailed analysis). The scaling factor $\sigma$ of the cosine classifier is set to 25 (please refer to \Cref{table:comp_classifier} and the corresponding paragraph for the analysis). All experiments are conducted on a single NVIDIA A800 GPU. In fact, a GPU with 20GB of memory is sufficient for all reproductions.

\section{Understanding Logit-Adjusted Loss}
\label{sec:la_loss}

\paragraph{Proof.} Logit-adjusted loss \citep{menon2021longtail} (or termed Balanced-Softmax \citep{ren2020balanced}) is widely used in previous literature \citep{hong2021disentangling,zhao2022adaptive,li2022long} because of its theoretical optimality and formal simplicity. Following we give a brief proof:
\begin{align}
\label{eq:source}
\oP_s(y=j\mid\vx)
&= \cfrac{\oP_s(y=j\mid\vx)}{\sum_{k\in[K]}\oP_s(y=k\mid\vx)} \nonumber \\
&= \cfrac{\oP_t(y=j\mid\vx)\cdot\cfrac{\oP_s(y=j\mid\vx)}{\oP_t(y=j\mid\vx)}}{\sum_{k\in[K]}\oP_t(y=k\mid\vx)\cdot\cfrac{\oP_s(y=k\mid\vx)}{\oP_t(y=k\mid\vx)}} \nonumber \\
&= \cfrac{\oP_t(y=j\mid\vx)\cdot\cfrac{\oP_s(\vx\mid y=j)}{\oP_t(\vx\mid y=j)}\cdot\cfrac{\oP_s(y=j)}{\oP_t(y=j)}\cdot\cfrac{\oP_t(\vx)}{\oP_s(\vx)}}{\sum_{k\in[K]}\oP_t(y=k\mid\vx)\cdot\cfrac{\oP_s(\vx\mid y=k)}{\oP_t(\vx\mid y=k)}\cdot\cfrac{\oP_s(y=k)}{\oP_t(y=k)}\cdot\cfrac{\oP_t(\vx)}{\oP_s(\vx)}} \nonumber \\
&= \cfrac{\oP_t(y=j\mid\vx)\cdot\cfrac{\oP_s(\vx\mid y=j)}{\oP_t(\vx\mid y=j)}\cdot\cfrac{\oP_s(y=j)}{\oP_t(y=j)}}{\sum_{k\in[K]}\oP_t(y=k\mid\vx)\cdot\cfrac{\oP_s(\vx\mid y=k)}{\oP_t(\vx\mid y=k)}\cdot\cfrac{\oP_s(y=k)}{\oP_t(y=k)}}
\end{align}
where $\oP_s$ is the probability distribution in the source domain (\ie the training dataset) and $\oP_t$ is the probability distribution in the target domain (\ie the test dataset). For long-tail learning, $\oP_s(y)$ appears a long-tail distribution and $\oP_t(y)$ is a uniform distribution, \ie $\oP_t(y=k)=\cfrac{1}{K}$. Moreover, by assuming that $\oP_s(\vx\mid y)=\oP_t(\vx\mid y)$, we have
\begin{equation}
\label{eq:target_to_source}
\oP_s(y=j\mid\vx)
= \cfrac{\oP_t(y=j\mid\vx)\cdot\oP_s(y=j)}{\sum_{k\in[K]}\oP_t(y=k\mid\vx)\cdot\oP_s(y=k)}
\end{equation}
In deep classification models, $\oP_t(y\mid\vx)$ is estimated by the Softmax of the output logits $\vz$, which is
\begin{equation}
\label{eq:softmax_def}
\oP_t(y=j\mid\vx)=\cfrac{\exp(z_j)}{\sum_{k\in[K]}\exp(z_k)}
\end{equation}
In order to get the optimal probability model in the target domain, we substitute $\oP_t(y\mid\vx)$ in \Cref{eq:target_to_source} with \Cref{eq:softmax_def}. In this way, we optimize the predicted probability in the source domain as
\begin{align}
\label{eq:source_softmax}
\oP_s(y=j\mid\vx)&=\cfrac{\cfrac{\exp(z_j)}{\sum_{k\in[K]}\exp(z_k)}\cdot\oP_s(y=j)}{\sum_{k\in[K]}\cfrac{\exp(z_k)}{\sum_{k'\in[K]}\exp(z_k')}\cdot\oP_s(y=k)} \nonumber \\
&=\cfrac{\exp(z_j)\cdot\oP_s(y=j)}{\sum_{k\in[K]}\exp(z_k)\cdot\oP_s(y=k)} \nonumber \\
&=\cfrac{\exp\left(z_j+\log\oP_s(y=j)\right)}{\sum_{k\in[K]}\exp\left(z_k+\log\oP_s(y=k)\right)}
\end{align}
Therefore, the logit-adjusted loss is defined as
\begin{equation}
\label{eq:la_loss_define}
\gL_{\text{LA}}(\vx,y=j)=-\log\oP_s(y=j\mid\vx)=-\log\frac{\exp({z_j}+\log \oP_s(y=j))}{\sum_{k\in[K]}\exp({z_k}+\log \oP_s(y=k))}
\end{equation}

In practice, $\oP_s(y)$ can be estimated by calculating the class frequency in the training dataset. Compared to other elaborately designed losses, the LA loss does not require any hyperparameters and has fewer assumptions about the data distribution, making it more generalizable across different backbone models. As shown in \Cref{table:comp_loss}, some other elaborately designed losses such as LDAM and LADE, perform unsatisfactorily when applied to the CLIP model.

It is worth mentioning that, there is also a post-hoc logit-adjustment method, where $\oP_s(y=j\mid\vx)$ is directly calculated by the Softmax of the logits $\vz$. In this case, the posterior probability in the target domain can be estimated by
\begin{equation}
\label{eq:post_hoc_la}
\oP_t(y=j\mid\vx)=\cfrac{\exp\left(z_j-\log\oP_s(y=j)\right)}{\sum_{k\in[K]}\exp\left(z_k-\log\oP_s(y=k)\right)}
\end{equation}
The proof is similar to the above.

\paragraph{Impact of Class-Conditional Distribution.}
Logit-adjusted loss assumes that the class-conditional distribution is consistent between the source domain (the training data) and the target domain (the test data), \ie $\oP_s(\vx\mid y)=\oP_t(\vx\mid y)$. However, our empirical findings in \Cref{fig:feature-full} reveal that full fine-tuning breaks this assumption. By optimizing all of the model parameters, the inter-class similarities and the intra-class distances are significantly reduced. However, the limited data in tail classes makes it hard to generalize in the unknown test scenarios, and the intra-class distances of tail classes in test data remain at a high level. In this case, using $\oP_s$ to estimate the unknown data $\vx$ in $\oP_t$ will lead to an underestimated $\oP_t(\vx\mid y)$.

By revisiting the proof of logit-adjusted loss from \Cref{eq:source,eq:target_to_source,eq:softmax_def,eq:source_softmax,eq:la_loss_define} and removing the consistency assumption for class-conditional distribution, we have
\begin{align}
\gL_{\text{LA}}(\vx,y=j)=-\log\oP_s(y=j\mid\vx)
&=-\log\cfrac{\oP_t(y=j\mid\vx)\cdot{\oP_s(y=j)\cdot\zeta_{s-t}(j)}}{\sum_{k\in[K]}\oP_t(y=k\mid\vx)\cdot{\oP_s(y=k)}\cdot\zeta_{s-t}(k)} \nonumber \\
&=-\log\cfrac{\exp\left(z_j+\log\oP_s(y=j)+\log\zeta_{s-t}(j)\right)}{\sum_{k\in[K]}\exp\left(z_k+\log\oP_s(y=k)+\log\zeta_{s-t}(k)\right)}
\end{align}
where $\zeta_{s-t}(k)$ denotes $\cfrac{\oP_s(\vx\mid y=k)}{\oP_t(\vx\mid y=k)}$. For samples of class $j$, the underestimate of $\oP_t(\vx\mid y=j)$ results in the underestimate of $\gL_{\text{LA}}(\vx,y=j)$, thus inevitably leading to the optimization biased towards other classes and the performance decline of class $j$.
Similarly, for post-hoc logit adjustment in \Cref{eq:post_hoc_la}, the estimated posterior probability should be
\begin{equation}
\oP_t(y=j\mid\vx)=\cfrac{\exp\left(z_j-\log\oP_s(y=j)-\log\zeta_{s-t}(j)\right)}{\sum_{k\in[K]}\exp\left(z_k-\log\oP_s(y=k)-\log\zeta_{s-t}(k)\right)}
\end{equation}
and the underestimate of $\oP_t(\vx\mid y=j)$ will result in the underestimate of posterior probability $\oP_t(y=j\mid\vx)$.

Since the class-conditional distribution in the target domain is hard to acquire, it is impractical to estimate the precise posterior probability given a distorted representation. Nevertheless, given that the foundation model has a strong representation learning capability, we can improve its separability as well as preserve its class-conditional distribution. Our empirical studies in \Cref{fig:lightweight} demonstrate that fine-tuning a small portion of the parameters can achieve such effects.

\section{Quantifying the Overfitting Issue of Full Fine-Tuning}
\label{sec:overfitting}

To give a more thorough comparison between full fine-tuning, classifier fine-tuning, and \algo, we compute their training and test accuracy, as well as the accuracy gap, and report the results in \Cref{table:train_test_gap}. We highlight the overfitting results with red colors. 
The results show that full fine-tuning tends to cause overfitting, as the gaps between training and test accuracy are obviously higher than the other two methods, particularly on the tail classes.

\begin{table}[!h]
\caption{Training and test accuracy of different fine-tuning methods on ImageNet-LT, Places-LT, and iNaturalist 2018. Note that for the class-imbalanced training data, we calculate their mean class accuracy. The {\red red} number denotes that the gap between training and test accuracy is larger than \algo\ by at least 1\%, which indicates overfitting.
}
\label{table:train_test_gap}
\setlength{\tabcolsep}{0.9ex} 
\begin{subtable}{\linewidth}
\caption{ImageNet-LT.}
\centering
\begin{small}
\begin{tabular}{l cccccccccccc}
\toprule
\multirow{2.5}{*}{\bf Methods} & \multicolumn{3}{c}{\bf Overall}  & \multicolumn{3}{c}{\bf Head} & \multicolumn{3}{c}{\bf Medium}  & \multicolumn{3}{c}{\bf Tail} \\
\cmidrule(lr){2-4}\cmidrule(lr){5-7}\cmidrule(lr){8-10}\cmidrule(lr){11-13}
 & train & test & $\Delta$  & train & test & $\Delta$  & train & test & $\Delta$  & train & test & $\Delta$ \\
\midrule
Full fine-tuning (\textit{best lr}) & 91.6 & 72.9 &\red 18.7 & 92.3 & 80.8 &\red 11.5 & 88.5 & 72.4 &\red 16.1 & 72.8 & 52.1 &\red 20.7 \\
Full fine-tuning (\textit{lr equal to \algo}) & 91.6 & 61.7 & \red 29.9 & 91.8 & 70.3 &\red 21.5 & 91.3 & 60.0 &\red 21.3 & 86.0 & 43.4 &\red 42.6 \\
Classifier (\textit{best lr}) & 86.1 & 73.5 &\red 12.6 & 83.2 & 76.6 & 6.6 & 86.9 & 72.8 &\red 14.1 & 91.7 & 67.2 &\red 24.5 \\
Classifier (\textit{lr equal to \algo}) & 82.8 & 73.1 & 9.7 & 82.6 & 77.0 & 5.6 & 83.8 & 73.1 & 10.7 & 79.3 & 61.6 & 17.7 \\
\midrule
\algo\ (Ours) & 87.1 & 77.0 & 10.1 & 86.8 & 80.2 & 6.6 & 87.9 & 76.1 & 11.8 & 88.9 & 71.5 & 17.4 \\
\bottomrule
\end{tabular}
\end{small}
\end{subtable}
\begin{subtable}{\linewidth}
\belowcaptionskip=0.1in
\caption{Places-LT.}
\centering
\begin{small}
\begin{tabular}{l cccccccccccc}
\toprule
\multirow{2.5}{*}{\bf Methods} & \multicolumn{3}{c}{\bf Overall}  & \multicolumn{3}{c}{\bf Head} & \multicolumn{3}{c}{\bf Medium}  & \multicolumn{3}{c}{\bf Tail} \\
\cmidrule(lr){2-4}\cmidrule(lr){5-7}\cmidrule(lr){8-10}\cmidrule(lr){11-13}
 & train & test & $\Delta$  & train & test & $\Delta$  & train & test & $\Delta$  & train & test & $\Delta$ \\
\midrule
Full fine-tuning (\textit{best lr}) & 74.7 & 46.4 &\red 28.3 & 74.8 & 51.0 &\red 23.8 & 78.1 & 47.6\red &\red 30.5 & 66.8 & 35.3 &\red 31.5 \\
Full fine-tuning (\textit{lr equal to \algo}) & 81.9 & 41.8 &\red 40.1 & 77.0 & 46.3 &\red 30.7 & 85.7 & 43.1 &\red 42.6 & 82.0 & 30.5 &\red 51.5 \\
Classifier (\textit{best lr}) & 64.8 & 48.5 &\red 16.3 & 55.0 & 48.6 & 6.4 & 66.3 & 49.1 &\red 17.2 & 79.1 & 46.8 &\red 32.3 \\
Classifier (\textit{lr equal to \algo}) & 59.6 & 48.3 & 11.3 & 54.5 & 49.3 & 5.2 & 62.4 & 49.9 & 12.5 & 62.6 & 42.5 & 20.1 \\
\midrule
\algo\ (Ours) & 64.2 & 51.5 & 12.7 & 58.3 & 51.3 & 7.0 & 65.9 & 52.2 & 13.7 & 71.1 & 50.5 & 20.6 \\
\bottomrule
\end{tabular}
\end{small}
\end{subtable}
\begin{subtable}{\linewidth}
\belowcaptionskip=0.1in
\caption{iNaturalist 2018.}
\centering
\begin{small}
\begin{tabular}{l cccccccccccc}
\toprule
\multirow{2.5}{*}{\bf Methods} & \multicolumn{3}{c}{\bf Overall}  & \multicolumn{3}{c}{\bf Head} & \multicolumn{3}{c}{\bf Medium}  & \multicolumn{3}{c}{\bf Tail} \\
\cmidrule(lr){2-4}\cmidrule(lr){5-7}\cmidrule(lr){8-10}\cmidrule(lr){11-13}
 & train & test & $\Delta$  & train & test & $\Delta$  & train & test & $\Delta$  & train & test & $\Delta$ \\
\midrule
Full fine-tuning (\textit{best lr}) & 95.9 & 72.7 &\red 23.2 & 91.8 & 70.3 &\red 21.5 & 96.1 & 73.1 &\red 23.0 & 96.8 & 72.7 &\red 24.1 \\
Full fine-tuning (\textit{lr equal to \algo}) & 95.8 & 70.1 &\red 25.7 & 85.1 & 63.7 &\red 21.4 & 95.8 & 70.0 &\red 25.8 & 98.7 & 71.8 &\red 26.9 \\
Classifier (\textit{best lr}) & 76.8 & 57.9 & 18.9 & 53.7 & 46.7 & 7.0 & 74.7 & 56.8 & 17.9 & 85.4 & 62.1 &\red 23.3 \\
Classifier (\textit{lr equal to \algo}) & 76.8 & 57.9 & 18.9 & 53.7 & 46.7 & 7.0 & 74.7 & 56.8 & 17.9 & 85.4 & 62.1 &\red 23.3 \\
\midrule
\algo\ (Ours) & 97.3 & 79.1 & 18.2 & 88.0 & 72.4 & 15.6 & 97.4 & 79.0 & 18.4 & 99.4 & 81.1 & 18.3 \\
\bottomrule
\end{tabular}
\end{small}
\end{subtable}
\end{table}

\section{Detailed Analysis of Structured Lightweight Fine-Tuning}
\label{sec:peft_analysis}

\paragraph{Preliminary to the Transformer Architecture.} A Transformer model consists of an embedding layer and multiple Transformer blocks. Formally, it first divides an input image $\vx$ into $m$ patches $\{\vx^{\text{p}}_i\}_{i=1}^m$. These patches are then embeded into sequences of $d$-dimensional vectors $\mE^0=\mathrm{Embed}([\vx^{\text{p}}_1;\cdots;\vx^{\text{p}}_m])\in\sR^{m\times d}$. The input embeddings are subsequently passed through $L$ Transformer blocks $\{\Phi^l\}_{l=1}^L$ within the model:
\begin{equation}
\label{eq:block}
\begin{split}
\mX^l = \Phi^l(\mX^{l-1}).\;
\text{Specifically}, \left \{
\begin{array}{ll}
\hat{\mX}^l=\mathrm{MSA}(\mathrm{LN}(\mX^{l-1}))+\mX^{l-1}\\
\mX^l=\mathrm{FFN}(\mathrm{LN}(\hat{\mX}^l))+\hat{\mX}^l\\
\end{array}
\right.
\end{split}
\end{equation}
\begin{equation}
\label{eq:msa}
\mathrm{MSA}^{(l)}(\mX)=\mathrm{Concat}_{h=1}^{H}\left(\mathrm{Softmax}\left(\frac{\mX\mW^{l,h}_Q(\mX\mW^{l,h}_K)^\top}{\sqrt{d}}\right)\mX\mW^{l,h}_V\right)\mW^{l}_O
\end{equation}
\begin{equation}
\label{eq:ffn}
\mathrm{FFN}^{(l)}(\mX)=\mathrm{ReLU}(\mX\mW^l_1)\mW^l_2
\end{equation}
Here, $\mathrm{MSA}$ denotes the multi-head self-attention and $H$ is the number of heads. $\mathrm{FFN}$ indicates the feed-forward network, and $\mathrm{LN}$ denotes layer normalization \citep{ba2016layer}. $\mW^{l,h}_Q,\mW^{l,h}_K,\mW^{l,h}_V\in\sR^{d\times \frac{d}{H}}$, $\mW^{l}_O\in\sR^{d\times d}$, $\mW^l_1\in\sR^{d\times 4d}$ and $\mW^l_2\in\sR^{4d\times d}$ are learnable projection weights. We hide the bias terms for simplification. $\mX^0$ ($\mX^l$ with $l=0$) is normally set as $\mE^0$, and will be added with an extra learnable token $\vc^0$ when performing classification tasks, \ie $\mX^0=[\vc^0;\mE^0]$. The feature is extracted from the same location of the last-layer sequence, which is $\vf=\mathrm{LN}(\vc^L)$.

The components of the Vision Transformer (ViT) are detailed in \Cref{table:clip_quantity}.
The total number of parameters in ViT can be calculated as follows:  $12Ld^2+(13L+m+d_0+6)d$, where $m$ denotes the number of separated image patches, $d_0$ represents the dimension of each image patch, and $d$ represents the dimension of the embedding features. For example, in the case of ViT-B/16, where $m=\frac{224}{16}\times\frac{224}{16}=196$, $d_0=16\times 16\times 3=768$, $d=768$, $L=12$, the parameter quantity amounts to 85,799,424 ($\approx$ 85.80M).
\begin{table}[!h]
\caption{Model architecture and parameter quantity for CLIP-ViT.}
\label{table:clip_quantity}
\setlength{\tabcolsep}{2ex} 
\centering
\begin{small}
\begin{tabular}{c|c|c|c|c}
\toprule
\bf Layers &\bf Components & \bf Variables &\bf Size &\bf \#Params. \\
\midrule
\multirow{3}{*}[-1.33ex]{Embedding} & $\mathrm{Projection}$ & - & $d_0\times d$ & \multirow{3}{*}[-1.33ex]{$(m+d_0+2)d$} \\
\cmidrule{2-4}
 & $\mathrm{Class~Token}$ & $\vc^0$ & $d$ & \\
\cmidrule{2-4}
 & $\mathrm{Positional}$ & - & $(m+1)\times d$ & \\
\midrule
 & $\mathrm{LN}$ & $\boldsymbol{\gamma},\boldsymbol{\beta}$ & $d, d$ & $2d$\\
\midrule
\multirow{8}{*}[-4.66ex]{\makecell{Block-$1$ \\ ($l=1$)}} & $\mathrm{LN}$ & $\boldsymbol{\gamma},\boldsymbol{\beta}$ & $d, d$ & \multirow{8}{*}[-4.66ex]{$12d^2+13d$}\\
\cmidrule{2-4}
 & \multirow{4}{*}[-2ex]{$\mathrm{MSA}$} & $\{\mW^{l,h}_Q,\vb^{l,h}_Q\}_{h=1}^{H}$ & $\{d\times\frac{d}{H}, \frac{d}{H}\}\times H$ \\ [1ex]
 & & $\{\mW^{l,h}_K,\vb^{l,h}_K\}_{h=1}^{H}$ & $\{d\times\frac{d}{H}, \frac{d}{H}\}\times H$ \\ [1ex]
 & & $\{\mW^{l,h}_V,\vb^{l,h}_V\}_{h=1}^{H}$ & $\{d\times\frac{d}{H}, \frac{d}{H}\}\times H$ \\ [1ex]
 & & $\mW^{l}_O,\vb^{l}_O$ & $d\times d, d$ \\
\cmidrule{2-4}
 & $\mathrm{LN}$ & $\boldsymbol{\gamma},\boldsymbol{\beta}$ & $d, d$ \\
\cmidrule{2-4}
 & \multirow{2}{*}[-0.66ex]{$\mathrm{FFN}$} & $\mW^{l}_1,\vb^{l}_1$ & $d\times 4d, 4d$ \\ [1ex]
 & & $\mW^{l}_2,\vb^{l}_2$ & $4d\times d, d$ \\ 
\midrule
$\vdots$ & $\vdots$ & $\vdots$ & $\vdots$ & $\vdots$ \\
\midrule
Block-$L$ & $\cdots$ & $\cdots$ & $\cdots$ & $12d^2+13d$ \\
\midrule
 & $\mathrm{LN}$ & $\boldsymbol{\gamma},\boldsymbol{\beta}$ & $d, d$ & $2d$\\
\bottomrule
\end{tabular}
\end{small}
\end{table}

\paragraph{Representative Structured Lightweight Fine-Tuning Methods.} In this paper, we explore lightweight fine-tuning by learning a few learnable parameters for the adaptation. This approach prevents the decline of generalization ability benefiting from the foundation model. As only a small set of task-specific parameters is introduced, the model not only mitigates overfitting but also exhibits rapid convergence. Concretely, we propose a unified framework \algo. 
This framework is versatile and inclusive, allowing for the incorporation of a range of structured modules, including but not limited to
\begin{itemize}[itemsep=0in,topsep=0.1in]
    \item \textit{Bias-terms Fine-tuning} (\textit{BitFit}) \citep{zaken2022bitfit} aims to fine-tune only the bias parts of the model. Formally, given a projection function $\mX\mW+\vb$, it freezes $\mW$ and optimizes $\vb$.
    \item \textit{Visual Prompt Tuning} (\textit{VPT}) \citep{jia2022visual} prepends learnable prompts $\mP^l\in\sR^{p\times d}$ at each layer to extend $\mX^l=[\vc^l;\mE^l]$ to $[\vc^l;\mP^l;\mE^l]$. It has two variations: 1) \textit{VPT-Shallow}, which only prepends prompts at the first layer; 2) \textit{VPT-Deep}, which prepends prompts at all layers.
    \item \textit{Adapter} \citep{houlsby2019parameter} proposes to optimize a bottleneck module. The definition is $\mathrm{Adapter}(\mX)=\mathrm{ReLU}(\mathrm{LN}(\mX)\mW_{\text{down}})\mW_{\text{up}}$, where $\mW_{\text{down}}\in\sR^{d\times r}$ and $\mW_{\text{up}}\in\sR^{r\times d}\ (r\ll d)$. In practical, it can be appended to the $\mathrm{FFN}$ layer to reconstruct $\mathrm{FFN}(\cdot)$ to $\mathrm{Adapter}(\mathrm{FFN}(\cdot))$.
    \item \textit{Low-Rank Adapter (LoRA)} \citep{hu2022lora} is applied to the weights in $\mathrm{MSA}$ module. Specifically, it optimizes $\mW_{\text{down}}$ and $\mW_{\text{up}}$ to update $\mW$ (\eg $\mW_Q$ or $\mW_V$) to $\mW+\mW_{\text{down}}\mW_{\text{up}}$.
    \item \textit{AdaptFormer} \citep{chen2022adaptformer} changes the sequential \textit{Adapter} to a parallel one. Formally, it computes $s\cdot\mathrm{Adapter}(\hat{\mX}^l)$ and adds it to $\mX^l$ in \Cref{eq:block}. Here, $s$ can be a manually set or learnable scaling parameter\protect\footnotemark[3].
\end{itemize}
\footnotetext[3]{\footnotesize\adaptformerurl}

We present the parameter quantities of the structured lightweight fine-tuning modules in \Cref{table:peft_quantity}. For all modules, the parameter quantities are at the polynomial level of $d$. Notably, $p$ for VPT and $r$ for Adapter are much smaller than $d$. In comparison to the entire Transformer block, a lightweight module is significantly low-complexity ($\gO(d)$ vs. $\gO(d^2)$).

Moreover, the parameter quantity for a classifier is approximately $Kd$, where $K$ is the number of classes. In \algo, we set the bottleneck dimension $r=2^{\lfloor\log_{2}{(\frac{K}{2L})}\rfloor}\leq\frac{K}{2L}$ for the AdaptFormer module, so that the total parameter quantity is $L\cdot 2rd\leq Kd$ (ignoring constant terms). As a result, it learns even fewer parameters than the classifier.
\begin{table}[!h]
\caption{Parameter quantities for structured lightweight fine-tuning modules in a Transformer block.}
\label{table:peft_quantity}
\setlength{\tabcolsep}{2ex} 
\centering
\begin{small}
\begin{tabular}{c|c|c|c|c}
\toprule
\bf Modules &\bf Components & \bf Variables &\bf Size &\bf \#Params. \\
\midrule
\multirow{8}{*}[-4.66ex]{BitFit} & $\mathrm{LN\text{-}bias}$ & $\boldsymbol{\beta}$ & $d$ & \multirow{8}{*}[-4.66ex]{$11d$}\\
\cmidrule{2-4}
 &\multirow{4}{*}[-2ex]{$\mathrm{MSA\text{-}bias}$} & $\{\vb^{l,h}_Q\}_{h=1}^{H}$ & $\{\frac{d}{H}\}\times H$ & \\ [1ex]
 & & $\{\vb^{l,h}_K\}_{h=1}^{H}$ & $\{\frac{d}{H}\}\times H$ \\ [1ex]
 & & $\{\vb^{l,h}_V\}_{h=1}^{H}$ & $\{\frac{d}{H}\}\times H$ \\ [1ex]
 & & $\vb^{l}_O$ & $d$ \\
\cmidrule{2-4}
 & $\mathrm{LN\text{-}bias}$ & $\boldsymbol{\beta}$ & $d$ & \\
\cmidrule{2-4}
 & \multirow{2}{*}[-0.66ex]{$\mathrm{FFN\text{-}bias}$} & $\vb^{l}_1$ & $4d$ \\ [1ex]
 & & $\vb^{l}_2$ & $d$ \\ 
\midrule
VPT & $\mathrm{Prompts}$ & $\mP^l$ & $p\times d$ & $pd$\\
\midrule
\multirow{3}{*}[-1.33ex]{Adapter} & $\mathrm{LN}$ & $\boldsymbol{\gamma},\boldsymbol{\beta}$ & $d, d$ & \multirow{3}{*}[-1.33ex]{$(2r+3)d+r$} \\
\cmidrule{2-4}
 & \multirow{2}{*}[-0.66ex]{$\mathrm{Projection}$} & $\mW_{\text{down}}, \vb_{\text{down}}$ & $d\times r, r$ & \\ [1ex]
 & & $\mW_{\text{up}}, \vb_{\text{up}}$ & $r\times d, d$ &\\
\midrule
\multirow{2}{*}[-0.66ex]{LoRA} & \multirow{2}{*}[-0.66ex]{$\mathrm{Projection}$} & $\mW_{\text{down}}, \mW_{\text{up}}$ (for $\mW_Q$) & $d\times r, r\times d$ & \multirow{2}{*}[-0.66ex]{$4rd$} \\ [1ex]
 & & $\mW_{\text{down}}, \mW_{\text{up}}$ (for $\mW_V$) & $d\times r, r\times d$ & \\
\midrule
\multirow{4}{*}[-2ex]{AdaptFormer} & $\mathrm{LN}$ & $\boldsymbol{\gamma},\boldsymbol{\beta}$ & $d, d$ & \multirow{4}{*}[-2ex]{$(2r+3)d+r+1$} \\
\cmidrule{2-4}
 & \multirow{2}{*}[-0.66ex]{$\mathrm{Projection}$} & $\mW_{\text{down}}, \vb_{\text{down}}$ & $d\times r, r$ & \\ [1ex]
 & & $\mW_{\text{up}}, \vb_{\text{up}}$ & $r\times d, d$ & \\
\cmidrule{2-4}
 & $\mathrm{Scaling}$ & $s$ & $1$ & \\
\bottomrule
\end{tabular}
\end{small}
\end{table}

\begin{figure*}[!h]
    \centering
    \includegraphics[width=0.32\linewidth]{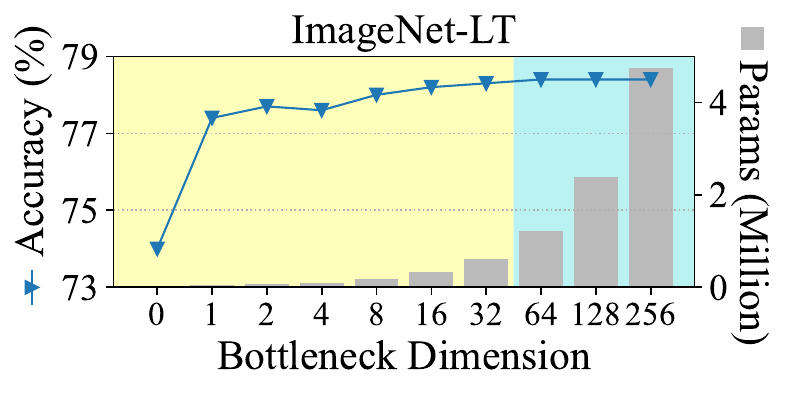}
    \hfill
    \includegraphics[width=0.32\linewidth]{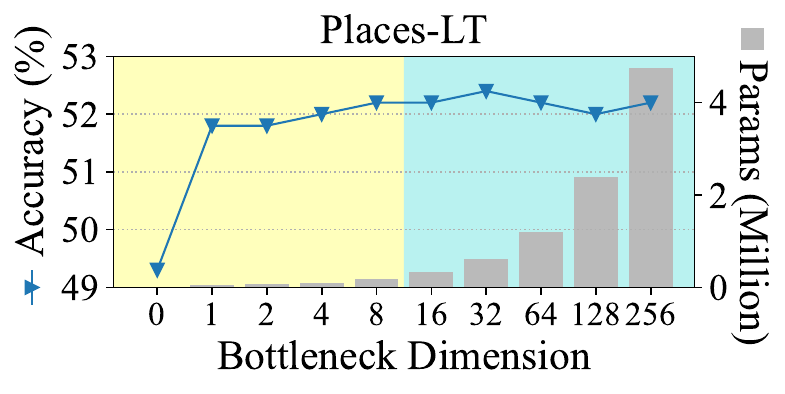}
    \hfill
    \includegraphics[width=0.32\linewidth]{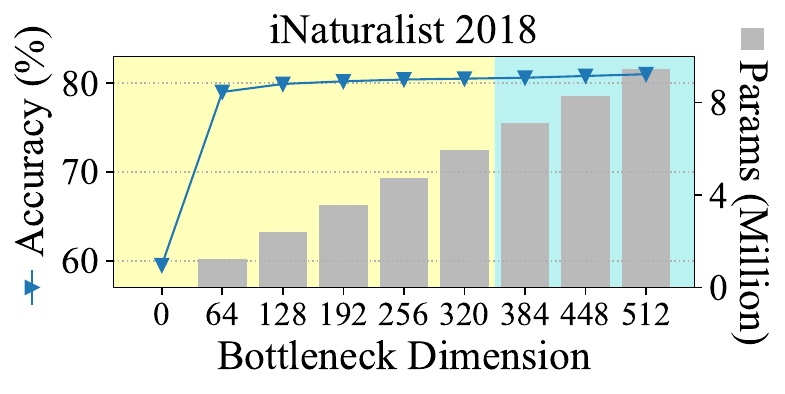}
    \caption{Comparison of different learnable parameters by changing the bottleneck dimension $r$. In the yellow area, the incorporated module has fewer learnable parameters than the classifier. The blue area is just the opposite.}
    \label{fig:bottle_dim}
\end{figure*}

\paragraph{Impact of the Quantity of Learnable Parameters.} In \algo, we can define the amounts of learnable parameters. In \Cref{fig:bottle_dim}, we study how much the parameters impact performance by controlling the bottleneck dimension $r$. Overall, we find that the performance is robust to the change of dimensions and it achieves the best results when employing comparable learnable parameters to the classifier.

\paragraph{Learned Representations of Varying Fine-Tuning Methods.}

\begin{figure}[!h]
    \centering
    \begin{subfigure}{0.32\linewidth}
        \includegraphics[width=0.32\linewidth]{figures/heatmap_imagenet_lt_zs.pdf}
        \includegraphics[width=0.32\linewidth]{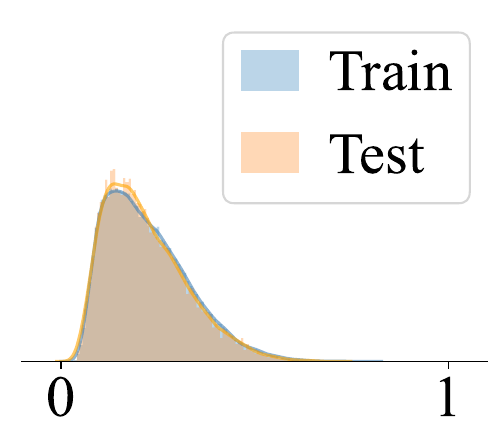}
        \includegraphics[width=0.32\linewidth]{figures/class_scatter_imagenet_lt_zs_tail.pdf}
        \belowcaptionskip=0.05in
        \caption{CLIP.}
    \end{subfigure}
    \hfill
    \begin{subfigure}{0.32\linewidth}
        \includegraphics[width=0.32\linewidth]{figures/heatmap_imagenet_lt_full.pdf}
        \includegraphics[width=0.32\linewidth]{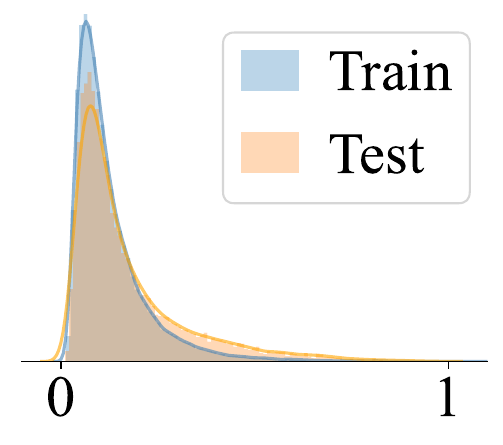}
        \includegraphics[width=0.32\linewidth]{figures/class_scatter_imagenet_lt_full_tail.pdf}
        \belowcaptionskip=0.05in
        \caption{Full fine-tuning.}
    \end{subfigure}
    \hfill
    \begin{subfigure}{0.32\linewidth}
        \includegraphics[width=0.32\linewidth]{figures/heatmap_imagenet_lt_sparse.pdf}
        \includegraphics[width=0.32\linewidth]{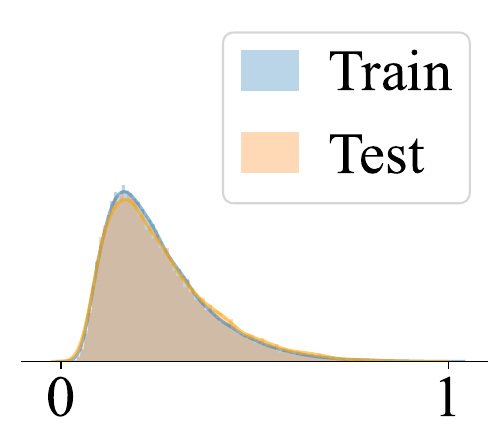}
        \includegraphics[width=0.32\linewidth]{figures/class_scatter_imagenet_lt_sparse_tail.pdf}
        \belowcaptionskip=0.05in
        \caption{Arbitrary lightweight fine-tuning.}
    \end{subfigure}
    \hfill
    \begin{subfigure}{0.32\linewidth}
        \includegraphics[width=0.32\linewidth]{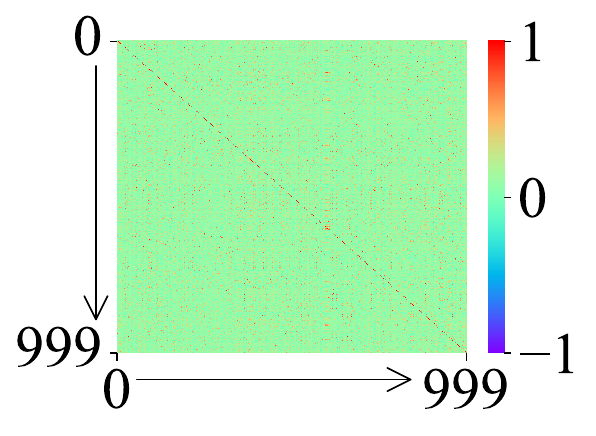}
        \includegraphics[width=0.32\linewidth]{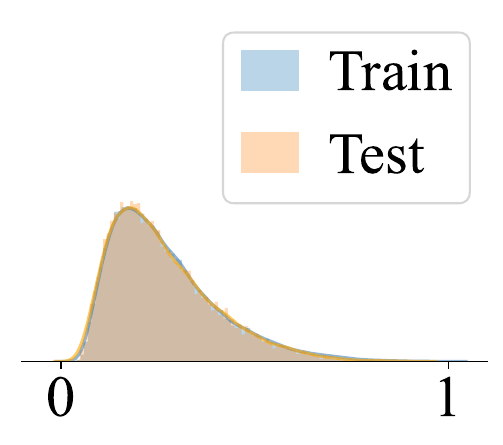}
        \includegraphics[width=0.32\linewidth]{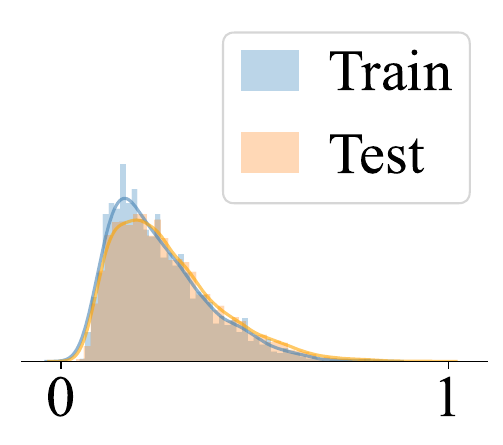}
        \belowcaptionskip=0.05in
        \caption{BitFit.}
    \end{subfigure}
    \hfill
    \begin{subfigure}{0.32\linewidth}
        \includegraphics[width=0.32\linewidth]{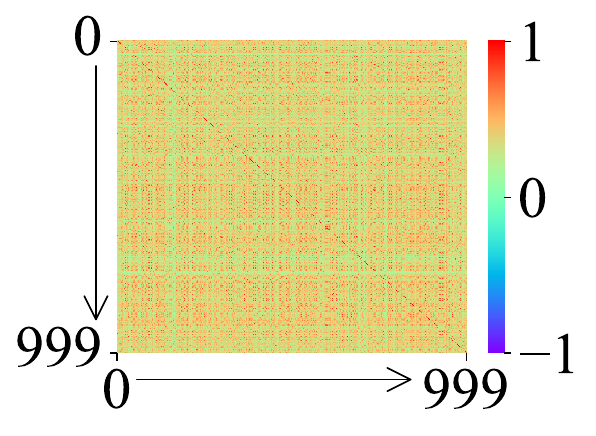}
        \includegraphics[width=0.32\linewidth]{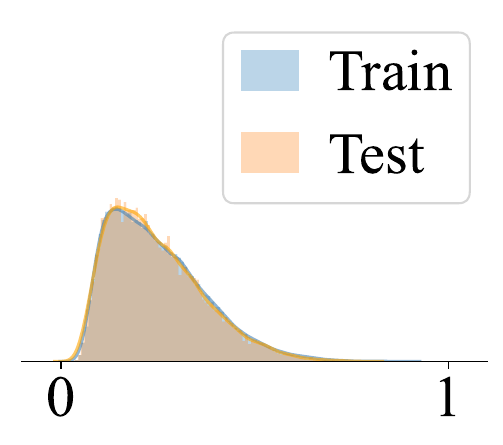}
        \includegraphics[width=0.32\linewidth]{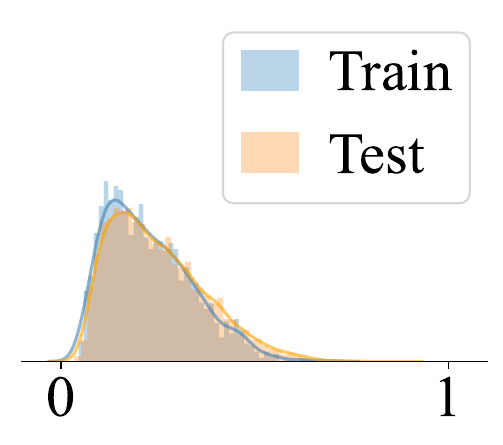}
        \belowcaptionskip=0.05in
        \caption{VPT-shallow.}
    \end{subfigure}
    \hfill
    \begin{subfigure}{0.32\linewidth}
        \includegraphics[width=0.32\linewidth]{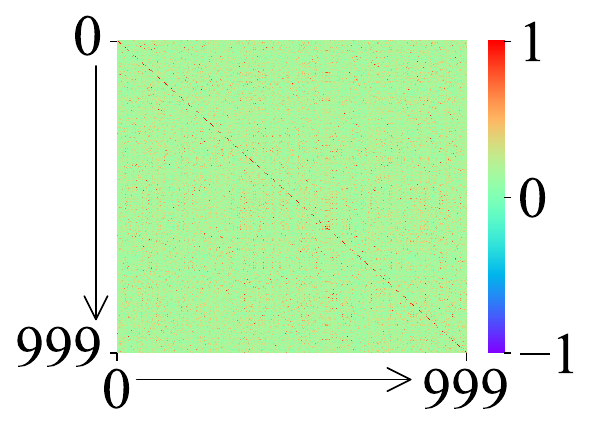}
        \includegraphics[width=0.32\linewidth]{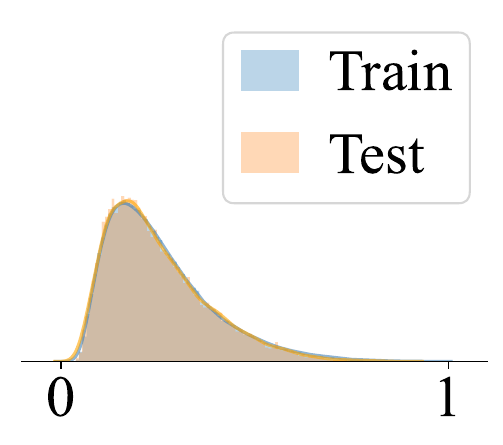}
        \includegraphics[width=0.32\linewidth]{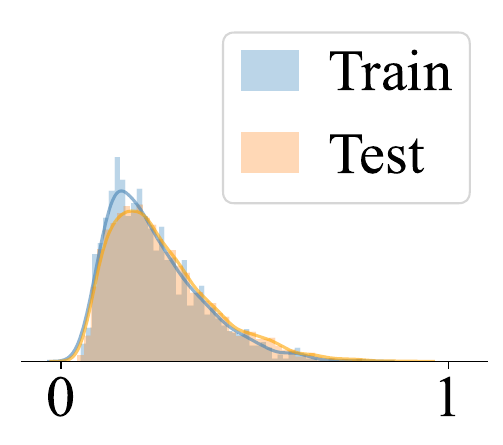}
        \belowcaptionskip=0.05in
        \caption{VPT-deep.}
    \end{subfigure}
    \hfill
    \begin{subfigure}{0.32\linewidth}
        \includegraphics[width=0.32\linewidth]{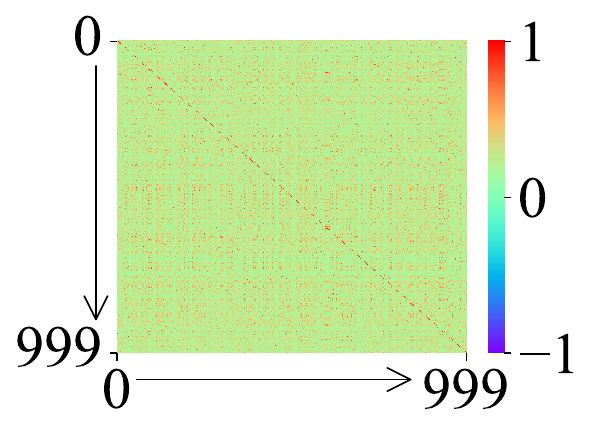}
        \includegraphics[width=0.32\linewidth]{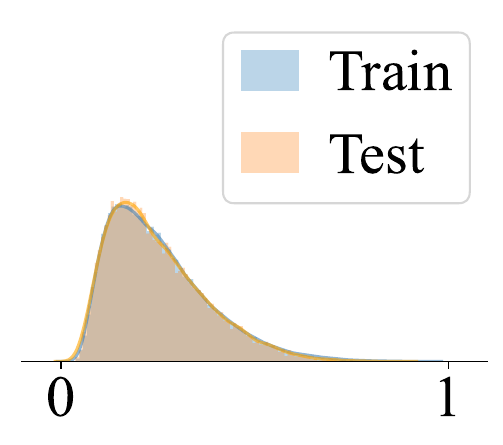}
        \includegraphics[width=0.32\linewidth]{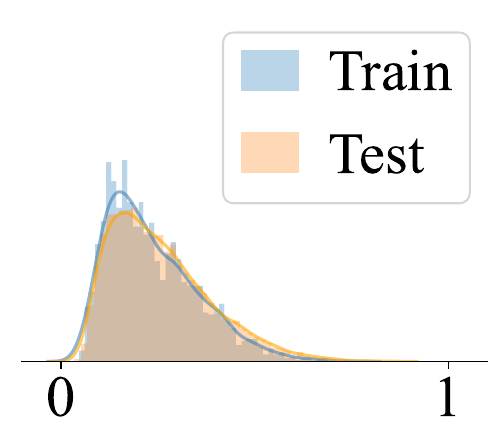}
        \caption{LoRA.}
    \end{subfigure}
    \hfill
    \begin{subfigure}{0.32\linewidth}
        \includegraphics[width=0.32\linewidth]{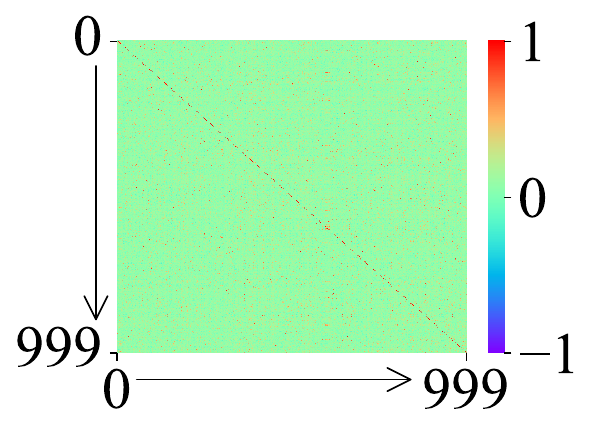}
        \includegraphics[width=0.32\linewidth]{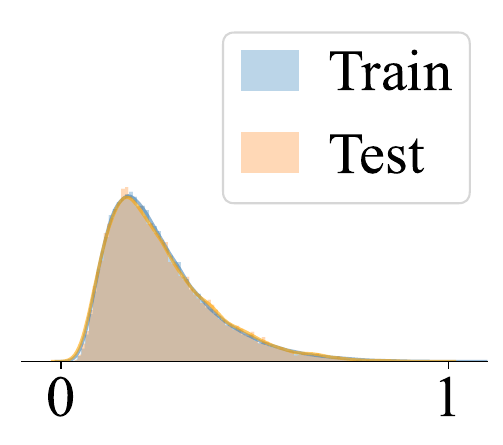}
        \includegraphics[width=0.32\linewidth]{figures/class_scatter_imagenet_lt_adapter_head.pdf}
        \caption{Adapter.}
    \end{subfigure}
    \hfill
    \begin{subfigure}{0.32\linewidth}
        \includegraphics[width=0.32\linewidth]{figures/heatmap_imagenet_lt_peft.pdf}
        \includegraphics[width=0.32\linewidth]{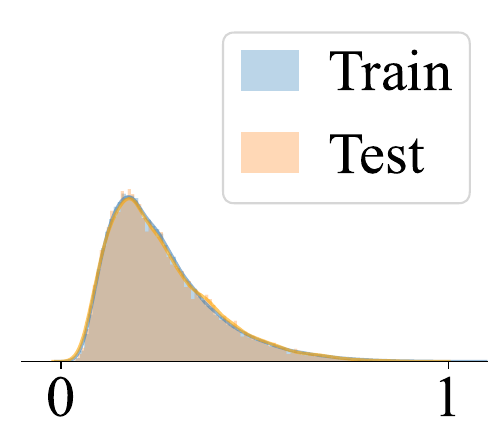}
        \includegraphics[width=0.32\linewidth]{figures/class_scatter_imagenet_lt_peft_tail.pdf}
        \caption{AdaptFormer.}
    \end{subfigure}
    \caption{Visualization of the inter-class feature similarities (the heatmaps) and intra-class distance distributions from head classes (the left histograms) and tail classes (the right histograms) on ImageNet-LT.}
    \label{fig:heatmap-and-scatter-imagenetlt}
\end{figure}

\begin{figure}[!h]
    \centering
    \begin{subfigure}{0.32\linewidth}
        \includegraphics[width=0.32\linewidth]{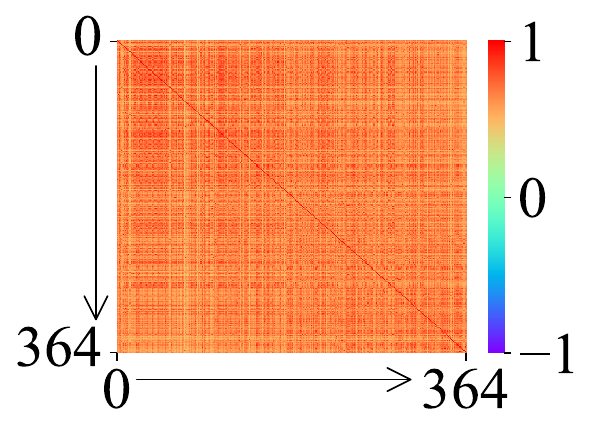}
        \includegraphics[width=0.32\linewidth]{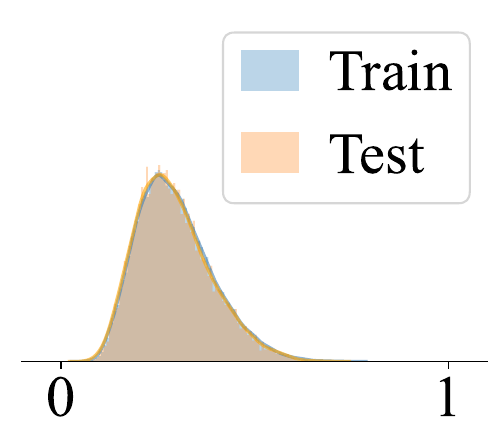}
        \includegraphics[width=0.32\linewidth]{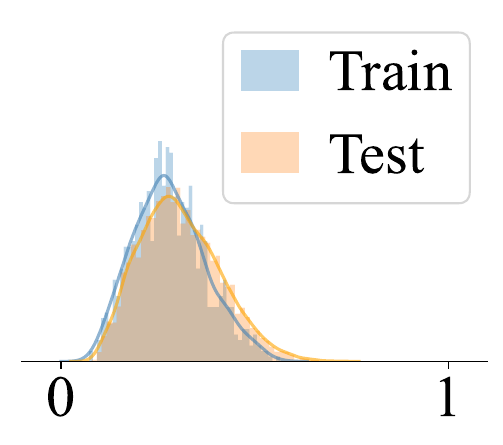}
        \belowcaptionskip=0.05in
        \caption{CLIP.}
    \end{subfigure}
    \hfill
    \begin{subfigure}{0.32\linewidth}
        \includegraphics[width=0.32\linewidth]{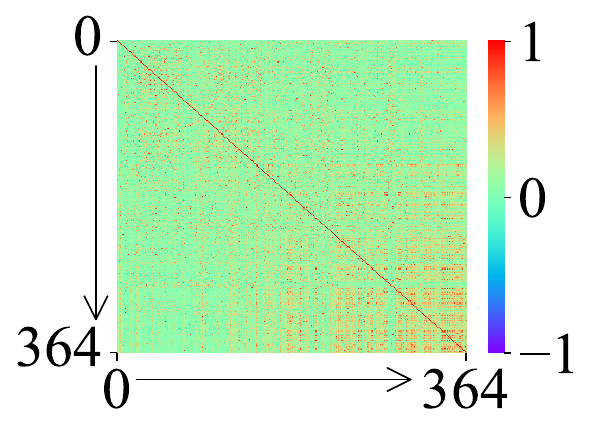}
        \includegraphics[width=0.32\linewidth]{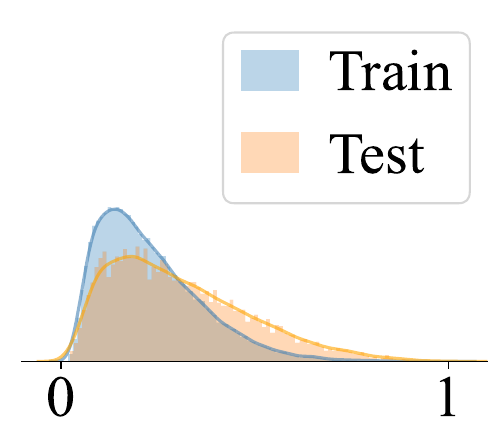}
        \includegraphics[width=0.32\linewidth]{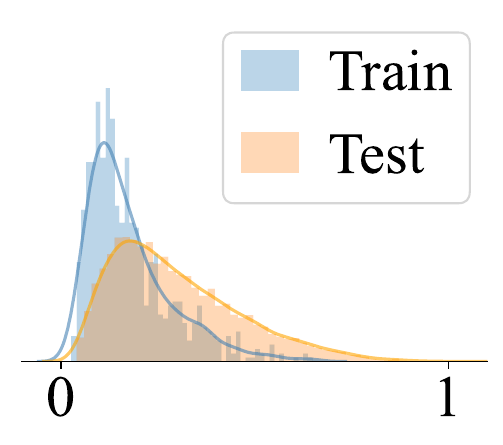}
        \belowcaptionskip=0.05in
        \caption{Full fine-tuning.}
    \end{subfigure}
    \hfill
    \begin{subfigure}{0.32\linewidth}
        \includegraphics[width=0.32\linewidth]{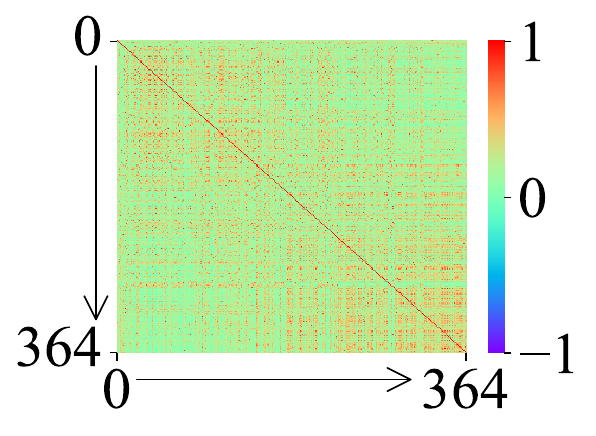}
        \includegraphics[width=0.32\linewidth]{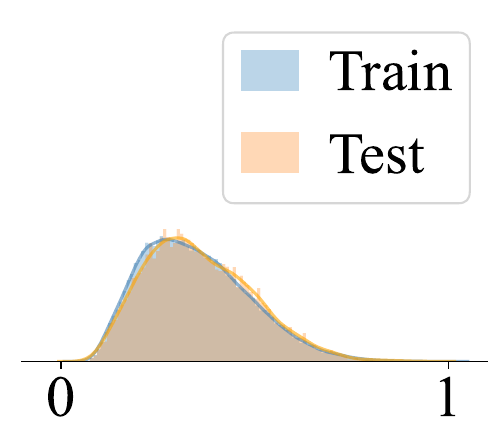}
        \includegraphics[width=0.32\linewidth]{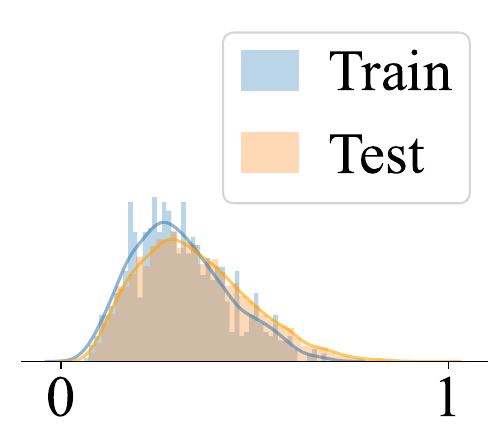}
        \belowcaptionskip=0.05in
        \caption{Arbitrary lightweight fine-tuning.}
    \end{subfigure}
    \hfill
    \begin{subfigure}{0.32\linewidth}
        \includegraphics[width=0.32\linewidth]{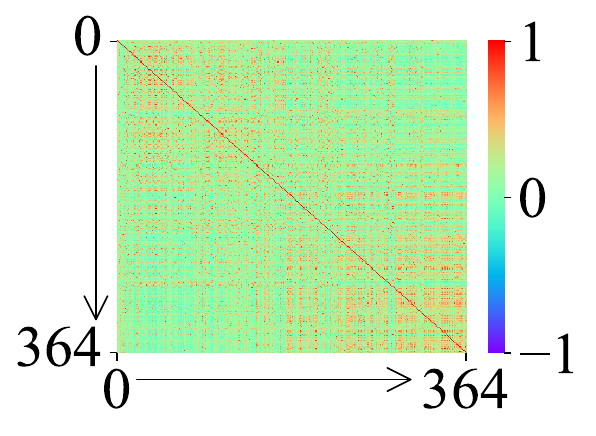}
        \includegraphics[width=0.32\linewidth]{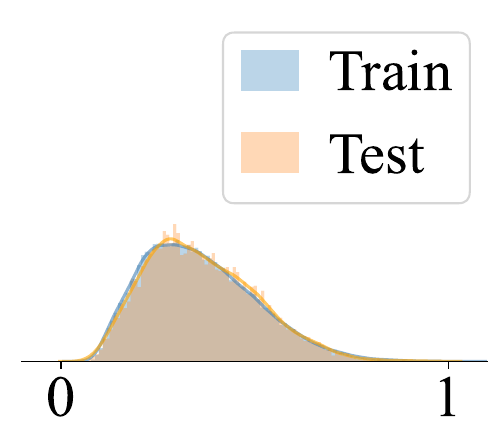}
        \includegraphics[width=0.32\linewidth]{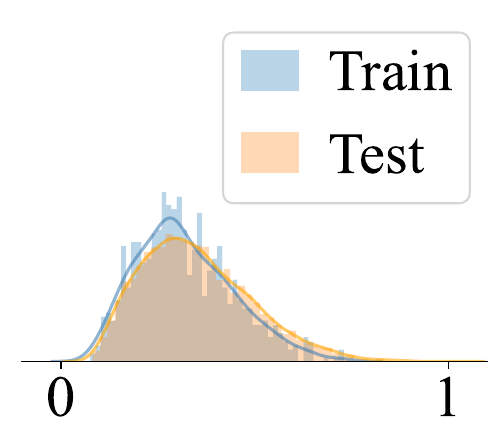}
        \belowcaptionskip=0.05in
        \caption{BitFit.}
    \end{subfigure}
    \hfill
    \begin{subfigure}{0.32\linewidth}
        \includegraphics[width=0.32\linewidth]{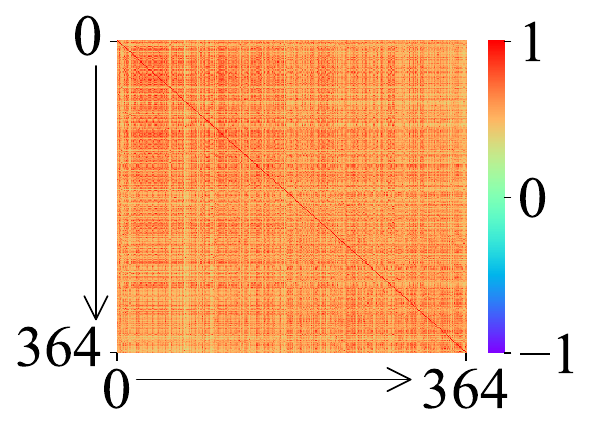}
        \includegraphics[width=0.32\linewidth]{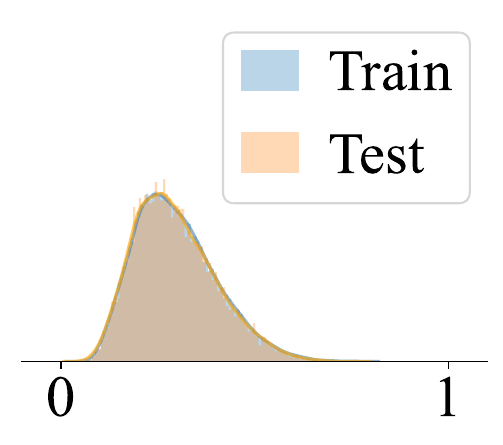}
        \includegraphics[width=0.32\linewidth]{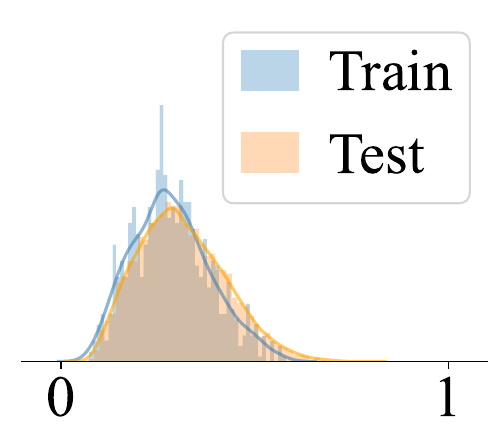}
        \belowcaptionskip=0.05in
        \caption{VPT-shallow.}
    \end{subfigure}
    \hfill
    \begin{subfigure}{0.32\linewidth}
        \includegraphics[width=0.32\linewidth]{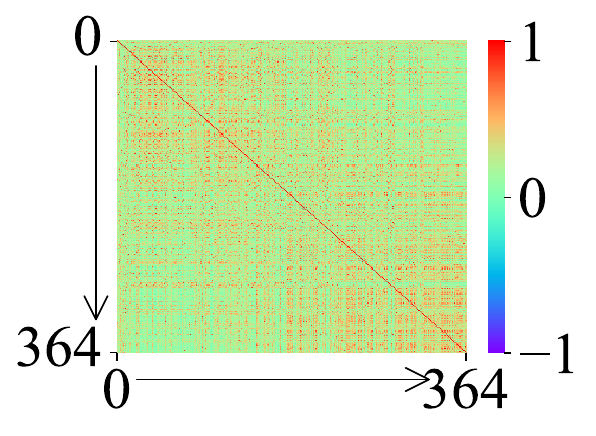}
        \includegraphics[width=0.32\linewidth]{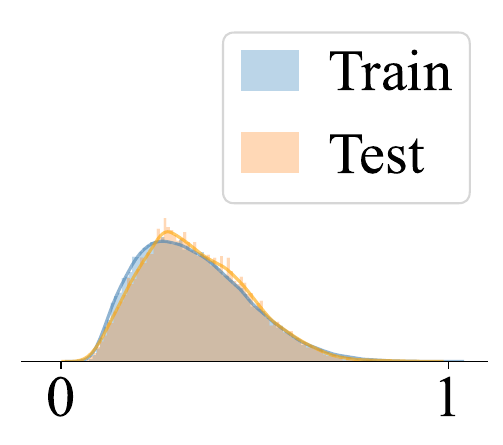}
        \includegraphics[width=0.32\linewidth]{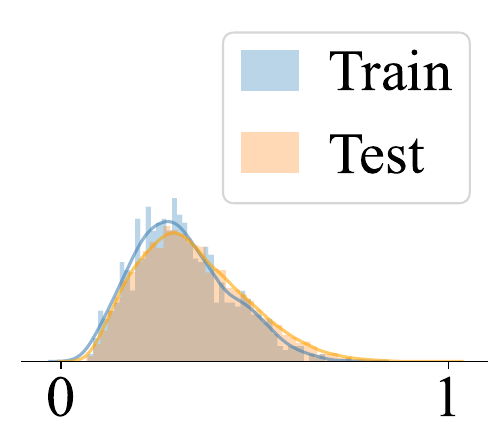}
        \belowcaptionskip=0.05in
        \caption{VPT-deep.}
    \end{subfigure}
    \hfill
    \begin{subfigure}{0.32\linewidth}
        \includegraphics[width=0.32\linewidth]{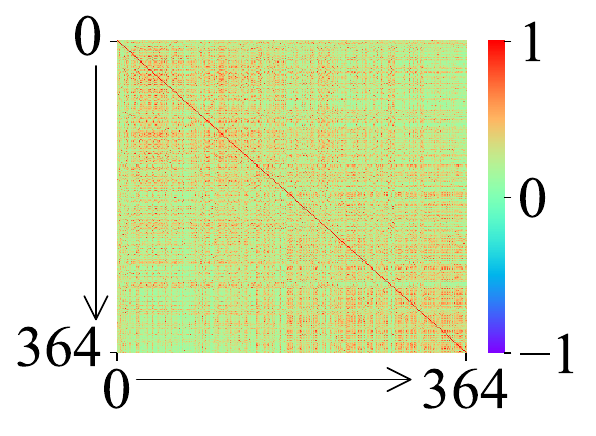}
        \includegraphics[width=0.32\linewidth]{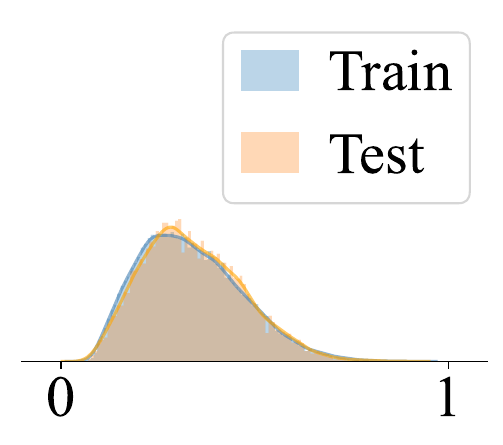}
        \includegraphics[width=0.32\linewidth]{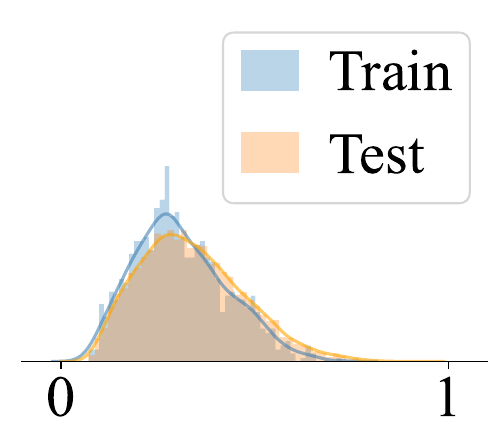}
        \caption{LoRA.}
    \end{subfigure}
    \hfill
    \begin{subfigure}{0.32\linewidth}
        \includegraphics[width=0.32\linewidth]{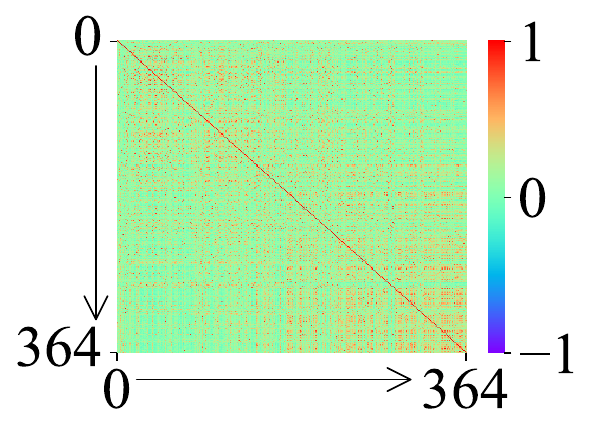}
        \includegraphics[width=0.32\linewidth]{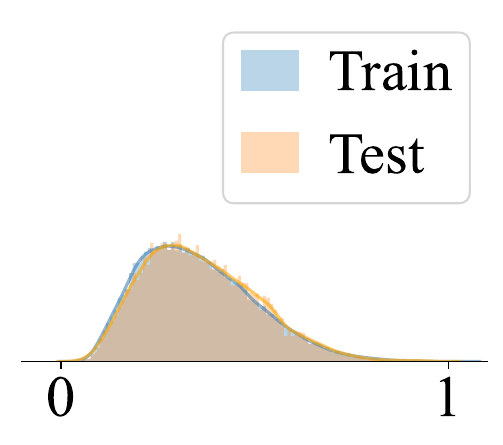}
        \includegraphics[width=0.32\linewidth]{figures/class_scatter_places_lt_adapter_head.pdf}
        \caption{Adapter.}
    \end{subfigure}
    \hfill
    \begin{subfigure}{0.32\linewidth}
        \includegraphics[width=0.32\linewidth]{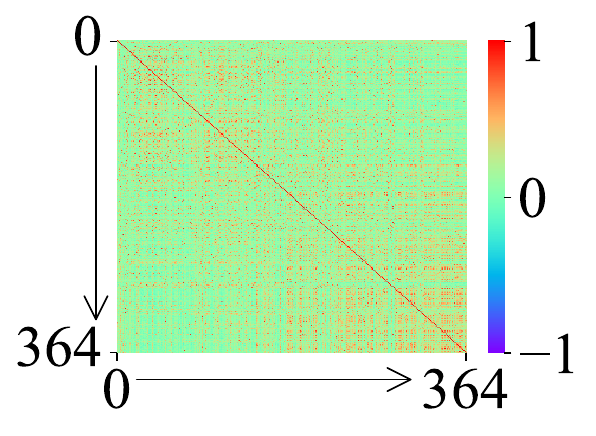}
        \includegraphics[width=0.32\linewidth]{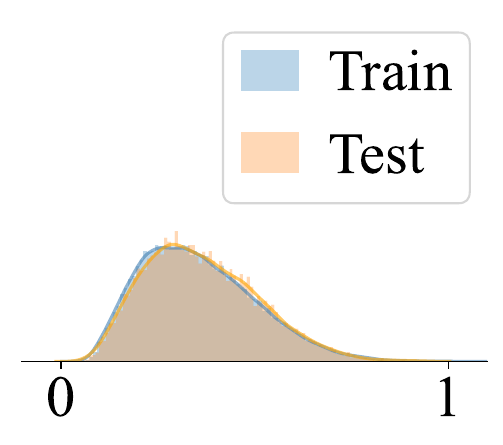}
        \includegraphics[width=0.32\linewidth]{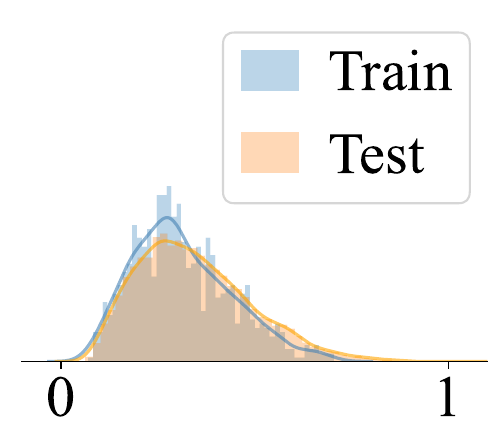}
        \caption{AdaptFormer.}
    \end{subfigure}
    \caption{Visualization of the inter-class feature similarities (the heatmaps) and intra-class distance distributions from head classes (the left histograms) and tail classes (the right histograms) on Places-LT.}
    \label{fig:heatmap-and-scatter-placeslt}
\end{figure}

In \Cref{fig:heatmap-and-scatter-imagenetlt,fig:heatmap-and-scatter-placeslt}, we illustrate the inter-class feature separabilities and intra-class distance distributions, based on the representation learned by (a) original CLIP, (b) full fine-tuning, (c) arbitrary lightweight fine-tuning, and (d-i) structured lightweight fine-tuning methods.

Compared to the original CLIP, all of these lightweight fine-tuning methods contribute to more distinctive representations. Among these methods, VPT-shallow may yield relatively weaker effects. This may be attributed to its design since VPT-shallow only prepends learnable prompts at the first layer. Beyond this, the other methods enable the feature separability even comparable to full fine-tuning. Notably, Adapter and AdaptFormer are more effective in enhancing separability, which is aligned with their superior performance in \Cref{table:comp_peft_module}.

Compared to full fine-tuning, all of these lightweight fine-tuning methods yield undistorted intra-class distributions. For both head and tail classes, the features of training and test data almost overlap under the same distribution, which is similar to the original CLIP. This brings about their stable performance improvements, especially on tail classes, as shown in \Cref{table:comp_peft_module}.

\paragraph{Effects of the Structured Lightweight Fine-Tuning Module on Each Layer.}

\begin{figure}[!h]
    \centering
    \includegraphics[width=0.32\linewidth]{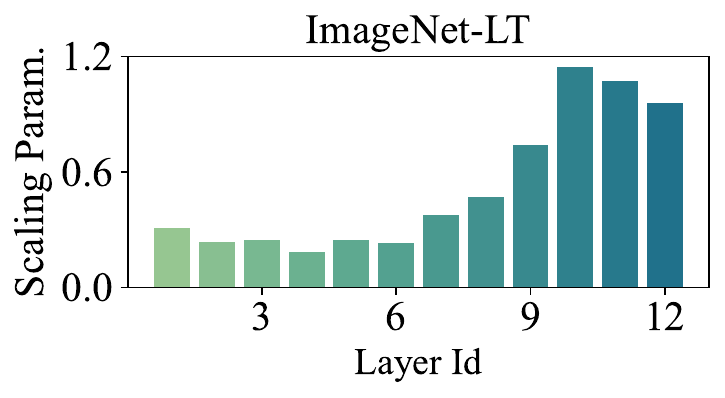}
    \hfill
    \includegraphics[width=0.32\linewidth]{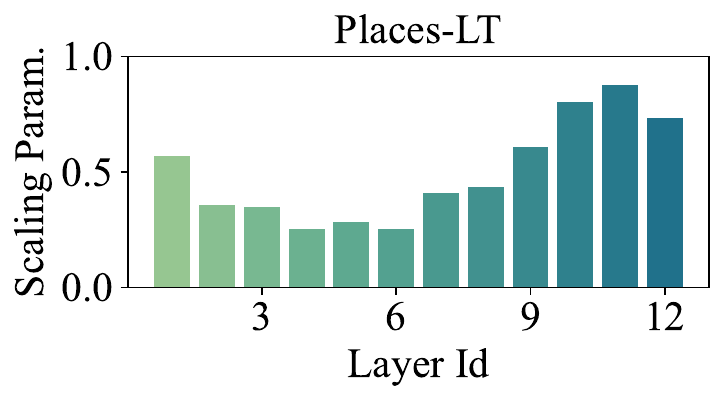}
    \hfill
    \includegraphics[width=0.32\linewidth]{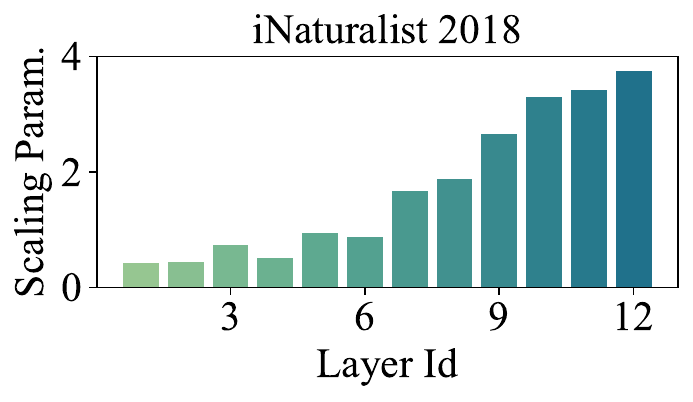}
    \caption{Learned scaling parameters of the AdaptFormer modules in different layers. AdaptFormer adaptively learns suitable scaling parameters for each layer, and the last several layers tend to have larger scaling parameters.}
    \label{fig:adaptformer_scale}
\end{figure}

In each layer, the output of the AdaptFormer module is multiplied by a learnable scaling parameter $s$ before being added to the corresponding block. Therefore, we can compare the values of $s$ to analyze the effects of the module for different layers. In \Cref{fig:adaptformer_scale}, we visualize the value of $s$ from each layer. It is inspiring that AdaptFormer can adaptively learn suitable scaling parameters for different layers. Moreover, the values of the last layers tend to be larger, which indicates that the adaptation of the last several layers is more significant for downstream classification tasks.

\paragraph{Lightweight Fine-Tuning vs. Partial Fine-Tuning.}

\begin{figure}[!h]
    \centering
    \includegraphics[width=0.275\linewidth]{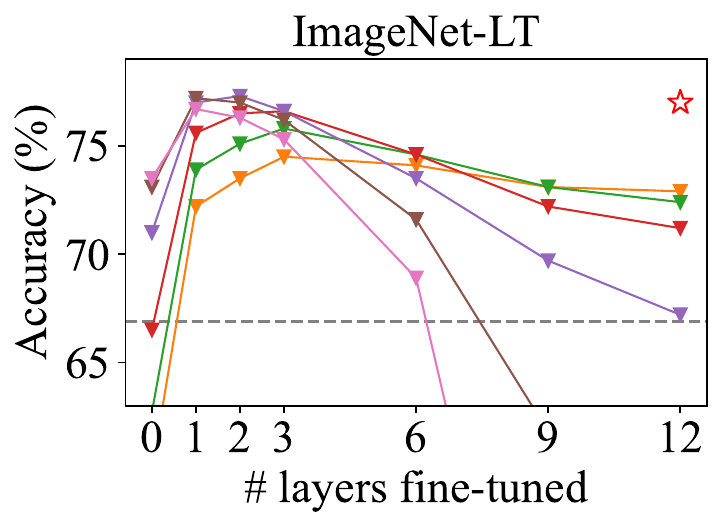}
    \includegraphics[width=0.275\linewidth]{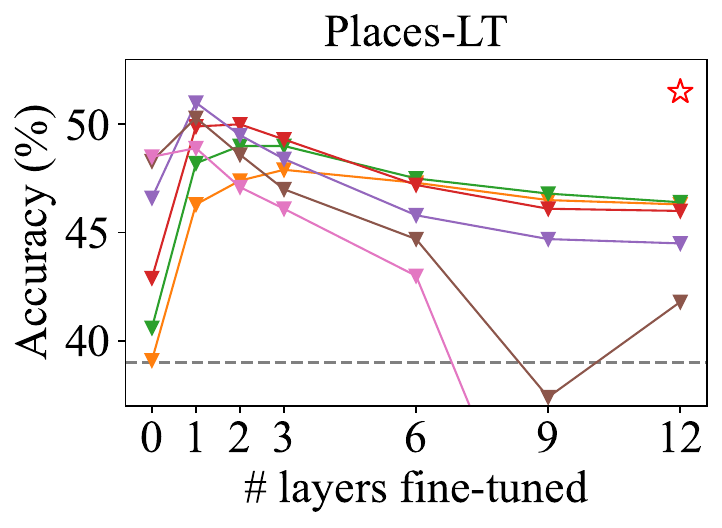}
    \includegraphics[width=0.422\linewidth]{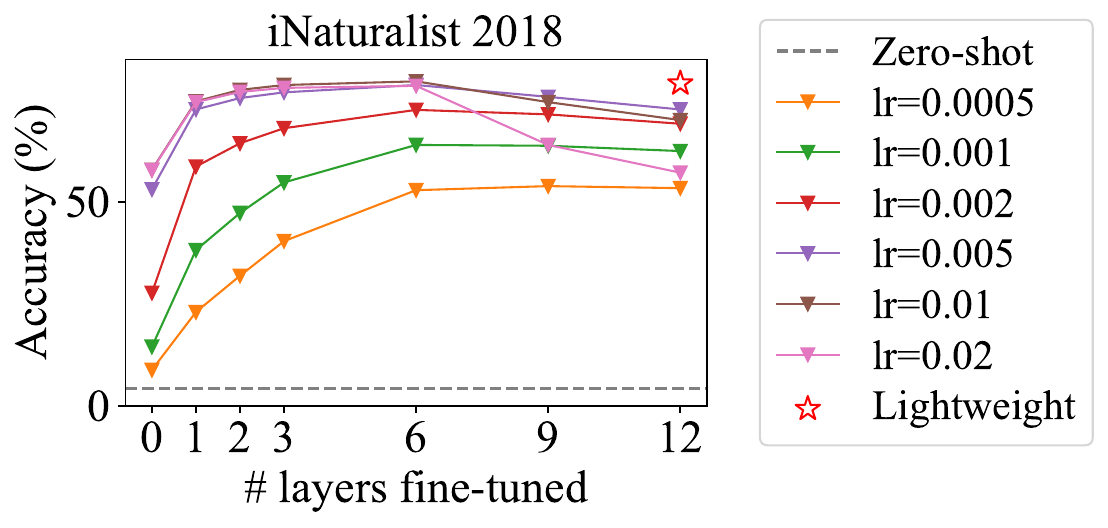}
    \caption{Partial fine-tuning the last $k$ layers. Both methods use cosine classifier and semantic-aware initialization for fair comparison. Similar to full fine-tuning, we search learning rate from \{0.02, 0.01, 0.005, 0.002, 0.001, 0.0005\} for partial fine-tuning. For \algo, we fix the learning rate to 0.01. Partial fine-tuning needs to elaborately choose the proper learning rate and the fine-tuned layers for the best performance, while \algo\ consistently performs optimally.}
    \label{fig:partial}
\end{figure}

Partial fine-tuning \citep{he2022masked} is an intuitive way to reduce the learnable parameters and avoid overfitting. Specifically, it fine-tunes the last $k$ layers of Transformer blocks while freezing the others. In \Cref{fig:partial}, we compare partial fine-tuning and lightweight fine-tuning (\algo) on ImageNet-LT, Places-LT, and iNaturalist 2018. Similar to full fine-tuning, partial fine-tuning is also sensitive to learning rate. When $k$ is small (\eg $0, 1, 2$), a higher learning rate is better. However, when $k$ is large (\eg $9, 12$), the high learning rate leads to a severe deterioration in the accuracy, whereby a smaller learning rate is more appropriate. Moreover, even if we have searched for the optimal learning rate, it is non-trivial to choose the number of fine-tuned layers $k$ for different datasets, since the best $k$ is $2$ for ImageNet-LT, $1$ for Places-LT, and $6$ for iNaturalist 2018. In contrast, lightweight fine-tuning consistently performs well with fixed hyperparameters.

\paragraph{Time Cost Analysis.} We also record the time cost of each structured lightweight fine-tuning method and report the results in \Cref{table:time_cost}. The results suggest that the time costs for different methods are highly close. 

\begin{table}[!h]
\caption{Training time per epoch when using different structured lightweight fine-tuning methods.}
\label{table:time_cost}
\setlength{\tabcolsep}{1.2ex} 
\centering
\begin{small}
\begin{tabular}{L{8ex}  L{18ex} | C{15ex} | C{15ex}}
\toprule
\multicolumn{2}{l|}{\bf Methods} &\bf ImageNet-LT &\bf Places-LT \\ 
\midrule
\multirow{7}{*}{\algo\ w/} & BitFit & 2 m 32 s & 1 m 23 s \\
& VPT-shallow & 2 m 31 s & 1 m 22 s \\
& VPT-deep & 2 m 40 s & 1 m 27 s \\
& Adapter & 2 m 37 s & 1 m 25 s \\
& LoRA & 2 m 33 s & 1 m 23 s \\
& AdaptFormer & 2 m 40 s & 1 m 28 s \\
\bottomrule
\end{tabular}
\end{small}
\end{table}

\section{Explanation of Test-Time Ensembling}
\label{sec:tte}

\begin{figure}[!h]
    \centering
    \includegraphics[width=0.32\linewidth]{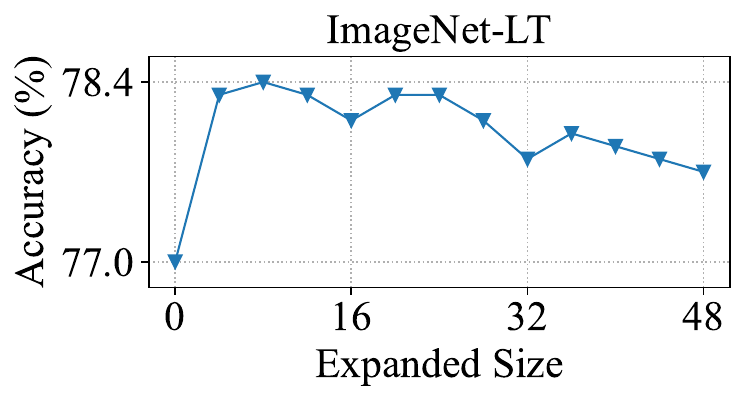}
    \hfill
    \includegraphics[width=0.32\linewidth]{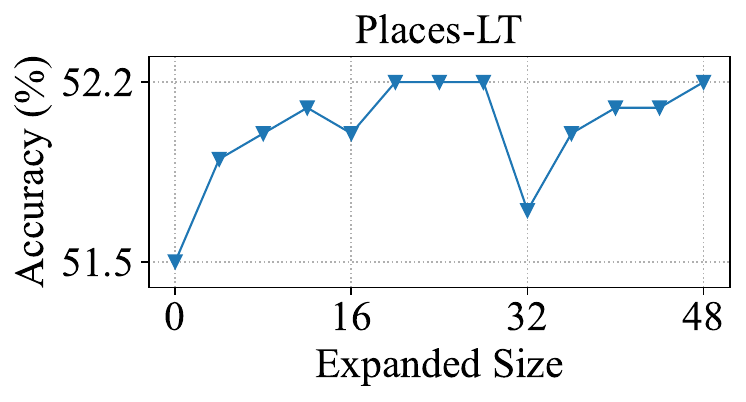}
    \hfill
    \includegraphics[width=0.32\linewidth]{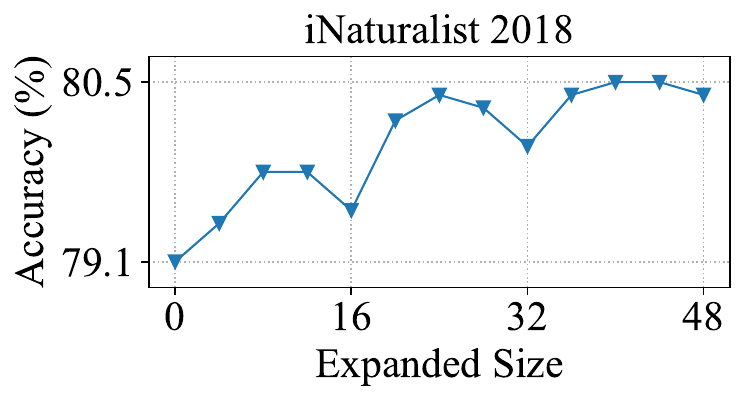}
    \caption{Performance of TTE with different expanded size $e$. Setting $e=0$ indicates not applying TTE. Setting $e$ to a multiple of the patch size $16$ yields suboptimal performance. Generally, $e=24$ is suitable for enhancing the generalization.}
    \label{fig:expand_size}
\end{figure}

We present the detailed procedures of \textit{test-time ensembling} (TTE) in \Cref{alg:tte}, using ViT-B/16 ($224\times 224$ resolution) as the backbone model. The highlighted lines denote the additional steps introduced by TTE. Conventionally, an image is first resized and center-cropped, and then split into patches before being fed into the Transformer model. However, this approach inevitably leads to the segmentation of important patterns across different patches, thus impeding the generalization. By employing diverse croppings, patterns that might be segmented in one cropping will be preserved in another. It is crucial to emphasize that the expanded size $e$ should not be a multiple of the patch size $16$; otherwise, the five cropped images will share a significant portion of the same patches, rendering the expected diversity unattainable. In \algo, we default to set $e=24$. Furthermore, we conduct a comparison of different expanded sizes and report the results in \Cref{fig:expand_size}.
Aside from TTE, we explore other augmentation techniques such as TTE + Flipping or Random Augmentations \citep{he2016deep} multiple times. The results in \Cref{table:augmentation_method} demonstrate that TTE is more effective than other augmentation methods.

\begin{algorithm}[!h]
\renewcommand{\algorithmicrequire}{\textbf{Input:}}
\renewcommand{\algorithmicensure}{\textbf{Output:}}
\caption{\textsc{Test-Time Ensembling}}
\label{alg:tte}
\small
\setstretch{1.2}
\begin{algorithmic}[1]
\REQUIRE 
Image $\vx$, expanded size $e$, input resolution ($224$).
\STATE Resize $\vx$ to $\vx'$ sized $(224+e)\times(224+e)$.
\STATE Crop the center $224\times 224$ portion of $\vx'$, denoted by $\vx^\text{c}$.
\STATE Split $\vx^\text{c}$ evenly into $m$ patches $[\vx^{\text{p}}_1;\cdots;\vx^{\text{p}}_m]$ (each $\vx^{\text{p}}_i$ is sized $16\times 16$, and $m=\frac{224}{16}\times\frac{224}{16}=196$).
\STATE Calculate the feature $\vf^\text{c}$ and then the logits $\vz^\text{c}$.
\STATE \textcolor{mplgreen}{Crop the top left $224\times 224$ portion of $\vx'$, repeat procedure 3-5 and obtain the logits $\vz^\text{tl}$.}
\STATE \textcolor{mplgreen}{Crop the top right $224\times 224$ portion of $\vx'$, repeat procedure 3-5 and obtain the logits $\vz^\text{tr}$.}
\STATE \textcolor{mplgreen}{Crop the bottom left $224\times 224$ portion of $\vx'$, repeat procedure 3-5 and obtain the logits $\vz^\text{bl}$.}
\STATE \textcolor{mplgreen}{Crop the bottom right $224\times 224$ of portion $\vx'$, repeat procedure 3-5 and obtain the logits $\vz^\text{br}$.}
\ENSURE Predicted logits $\vz=\operatorname{Average}(\vz^\text{c}+\vz^\text{tl}+\vz^\text{tr}+\vz^\text{bl}+\vz^\text{br})$.
\end{algorithmic}
\end{algorithm}

\begin{table}[!h]
\caption{Comparison of different augmentation methods on ImageNet-LT. }
\label{table:augmentation_method}
\setlength{\tabcolsep}{1.2ex} 
\centering
\begin{small}
\begin{tabular}{l | c | cccc }
\toprule
\bf Augmentation methods & \bf Augmentation times &\bf Overall &\bf Head &\bf Medium &\bf Tail \\
\midrule
TTE (Ours) & 5 & \bf 78.3 &\bf 81.3 &\bf 77.4 &\bf 73.4 \\
TTE + Flipping & 10 &\bf 78.3 &\bf 81.3 & 77.3 & 73.3 \\
Random Augmentation & 5 & 76.7 & 79.7 & 75.5 & 71.9 \\
Random Augmentation & 10 & 77.3 & 80.3 & 76.3 & 72.2 \\
Random Augmentation & 15 & 77.7 & 80.8 & 76.6 & 72.3 \\
Random Augmentation & 20 & 77.8 & 80.9 & 76.7 & 72.7 \\
\bottomrule
\end{tabular}
\end{small}
\end{table}

\section{Textual Prompts for Semantic-Aware Initialization}
\label{sec:textual_prompts}

In \algo, we use ``\texttt{a photo of a [CLASS].}'' as the template to generate textual prompts and then compute their features to initialize the classifier weights. One may be concerned with the impact of the used prompts. We conduct experiments to compare different prompting methods, including 1) the original class name (``\texttt{[CLASS]}'') and 2) prompt ensembling \citep{radford2021clip} which applies different templates to class names. The results in \Cref{table:prompting_method} show that these prompts have similar performances and that using ``\texttt{a photo of a [CLASS].}'' is adequate for generalization.

Moreover, we posit that CLIP has seen sufficient language corpus, considering its pre-training on web-scale datasets. However, it is noteworthy that CLIP may fail to recognize specific class names. This probably stems from its limited vocabulary size (CLIP contains approximately 49K vocabulary) or encountering uncommon or novel concepts. In this case, semantic-aware initialization may regress to random initialization. In response to this, we explore an alternative approach by incorporating class descriptions (which can be crafted manually or generated using large language models). In practice, we follow \citet{menon2023visual} to generate the descriptions for each class, then combine these descriptions and calculate the textual feature for initialization. The results are reported in the bottom line of \Cref{table:prompting_method}, which shows that the use of class descriptions can also enhance the performance compared to random initialization.

\begin{table}[!h]
\caption{Comparison of different prompting methods on ImageNet-LT. }
\label{table:prompting_method}
\setlength{\tabcolsep}{1.2ex} 
\centering
\begin{small}
\begin{tabular}{L{40ex}| C{6ex} C{6ex} C{6ex} C{6ex}}
\toprule
\bf Prompting methods &\bf Overall &\bf Head &\bf Medium &\bf Tail \\
\midrule
None prompt (random initialization) & 76.1 & 80.8 & 75.9 & 63.2 \\
``\texttt{[CLASS]}'' & 78.2 &\bf 81.4 & 77.3 & 72.3 \\
``\texttt{a photo of a [CLASS].}'' &\bf 78.3 & 81.3 &\bf 77.4 &\bf 73.4 \\
Prompt ensembling &\bf 78.3 & 81.3 &\bf 77.4 & 73.3 \\
Class descriptions (w/o class names) & 77.4 & 81.3 & 76.9 & 68.2 \\
\bottomrule
\end{tabular}
\end{small}
\end{table}

\section{Additional Experiments}
\label{sec:addtional}

\paragraph{Comparison of Different Training Epochs.}

\begin{figure}[!h]
    \centering
    \begin{subfigure}{0.48\linewidth}
        \includegraphics[width=0.49\linewidth]{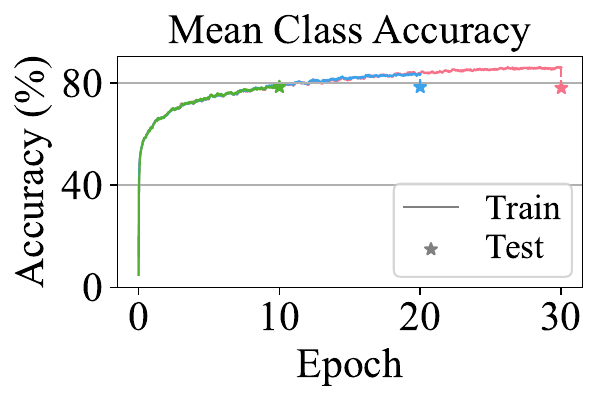}
        \includegraphics[width=0.49\linewidth]{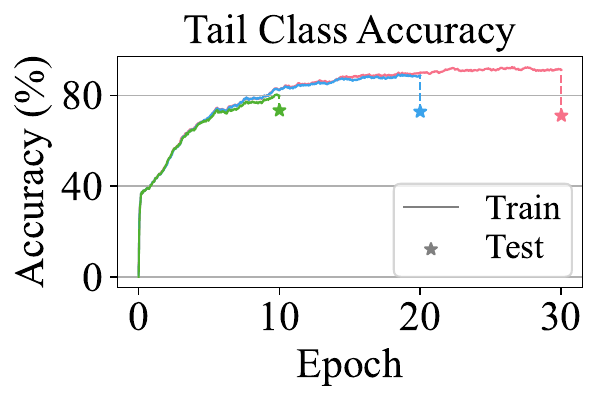}
        \caption{ImageNet-LT.}
    \end{subfigure}
    \begin{subfigure}{0.48\linewidth}
        \includegraphics[width=0.49\linewidth]{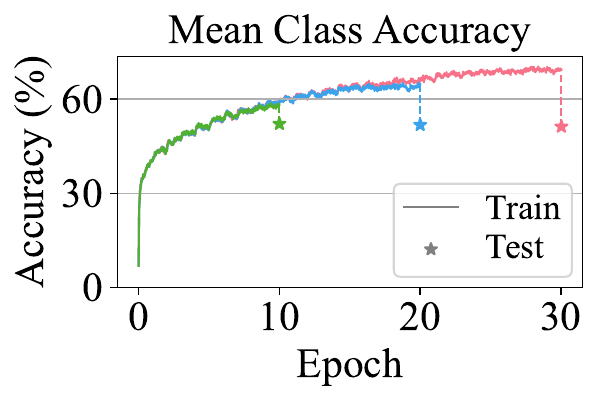}
        \includegraphics[width=0.49\linewidth]{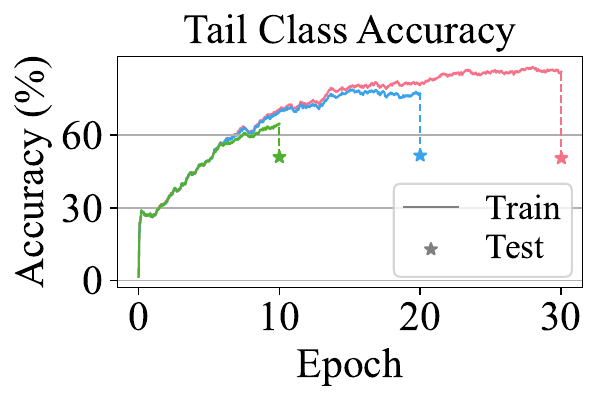}
        \caption{Places-LT.}
    \end{subfigure}
    \caption{Convergence curve of mean class and tail class accuracy.}
    \label{fig:convergence_epoch}
\end{figure}

In \algo, we train 10 epochs on ImageNet-LT and Places-LT, and 20 epochs on iNaturalist 2018 considering its large data scale. In \Cref{table:comp_epochs}, we report the results of training different epochs. On ImageNet-LT and Places-LT, increasing the training epochs does not yield significant improvements. Generally, 10-20 epochs are appropriate for most cases. We also visualize the training and test accuracy as a function of epochs in \Cref{fig:convergence_epoch}. When trained for more epochs ($>$20), \algo\ converges with higher training accuracy. Nonetheless, the test accuracy shows no corresponding enhancement. On iNaturalist 2018, when training for 5 epochs, \algo\ achieves an overall accuracy of 67.3\% (w/o TTE) / 68.6\% (w/ TTE), which surpasses Decoder \citep{wang2023exploring} by more than 8\% (please refer to \Cref{table:comp_inat18} for comparison). Moreover, by training more epochs (\eg 30 epochs), \algo\ achieves an additional performance improvement by 1\%. However, this will increase the computational overhead, so we abort this approach in \algo.

\begin{table}[!h]
\caption{Results of \algo\ (with and without TTE) on iNaturalist 2018 by training different epochs.}
\label{table:comp_epochs}
\setlength{\tabcolsep}{1.2ex} 
\centering
\begin{small}
\begin{tabular}{l|c| C{6ex} C{4ex} C{4ex} C{4ex} | C{6ex} C{4ex} C{4ex} C{4ex} | C{6ex} C{4ex} C{4ex} C{4ex}}
\toprule
\multirow{2}{*}{\bf Methods} & \multirow{2}{*}{\bf \#Epochs} & \multicolumn{4}{c|}{\bf ImageNet-LT} & \multicolumn{4}{c|}{\bf Places-LT} & \multicolumn{4}{c}{\bf iNaturalist 2018} \\
& &\bf Overall &\bf Head &\bf Med. &\bf Tail &\bf Overall &\bf Head &\bf Med. &\bf Tail &\bf Overall &\bf Head &\bf Med. &\bf Tail  \\
\midrule
\multirow{4}{*}{\algo\ (Ours)} & 5 & 76.2 & 79.9 & 75.8 & 67.2 & 50.7 & 51.2 & 52.0 & 46.6 & 67.3 & 70.4 & 71.0 & 61.8 \\
 & 10 & 77.0 & 80.2 & 76.1 & 71.5 & 51.5 & 51.3 & 52.2 & 50.5 & 76.1 & 71.3 & 75.9 & 77.5 \\
 & 20 & 77.1 & 80.7 & 75.9 & 71.4 & 51.1 & 51.0 & 51.3 & 51.0 & 79.1 & 72.4 & 79.0 & 81.1 \\
 & 30 & 76.9 & 81.1 & 75.6 & 69.4 & 50.3 & 50.9 & 50.1 & 49.5 & 80.1 & 73.8 & 80.0 & 81.9 \\
\midrule
\multirow{4}{*}{\algo\ w/ TTE (Ours)} & 5 & 77.5 & 80.9 & 77.2 & 69.0 & 51.3 &\bf 51.7 & 53.0 & 46.8 & 68.6 & 70.5 & 72.3 & 63.5 \\
 & 10 &\bf 78.3 & 81.3 &\bf 77.4 &\bf 73.4 &\bf 52.2 &\bf 51.7 &\bf 53.1 & 50.9 & 77.3 & 71.9 & 77.1 & 78.9 \\
 & 20 &\bf 78.3 & 81.8 & 77.1 & 72.8 & 51.8 & 51.3 & 52.3 &\bf 51.6 & 80.4 & 74.0 & 80.3 & 82.2 \\
 & 30 & 78.0 &\bf 82.2 & 76.6 & 71.1 & 51.3 & 51.4 & 51.5 & 50.5 &\bf 81.3 &\bf 75.1 &\bf 81.2 &\bf 83.0 \\
\bottomrule
\end{tabular}
\end{small}
\end{table}

\begin{figure}[!h]
    \centering
    \includegraphics[width=0.32\linewidth]{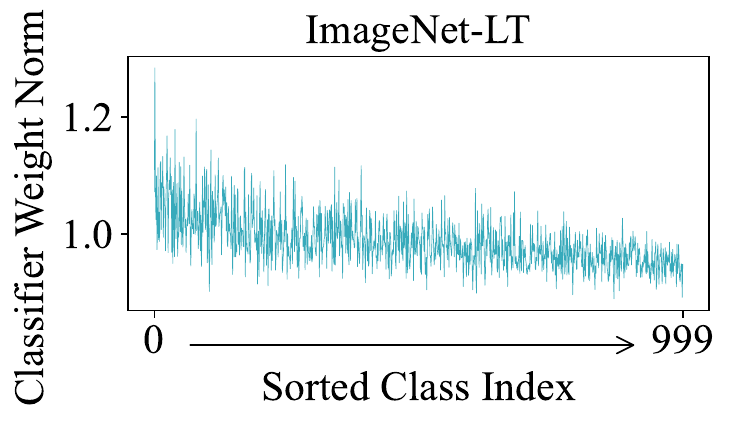}
    \hfill
    \includegraphics[width=0.32\linewidth]{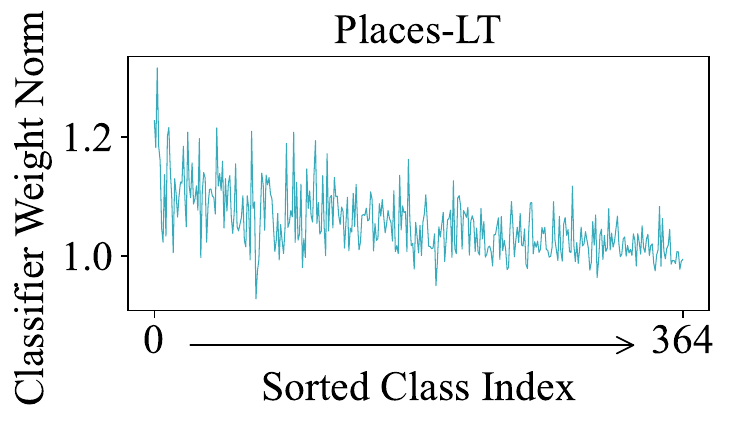}
    \hfill
    \includegraphics[width=0.32\linewidth]{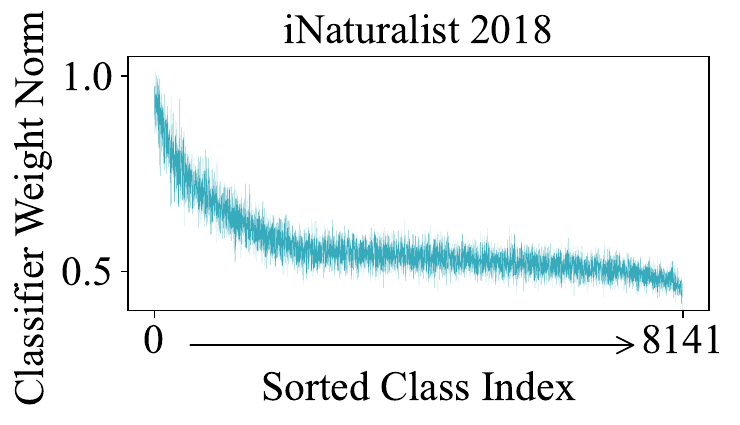}
    \caption{Weight norms of the learned linear classifier on three long-tail datasets. Classes are sorted by their frequency in the training dataset. On iNaturalist 2018, the weight norms are much more imbalanced, leading to a suboptimal performance of the linear classifier.}
    \label{fig:weight_norm}
\end{figure}

\paragraph{Comparison of Different Classifiers.}
In \algo, we default to optimize the cosine classifier with a scaling factor $\sigma=25$. We propose to use the cosine classifier to overcome the biased weight norms. We further assess the linear classifier $z_k=\vw_k^{\top}\vf+b$, the L2-normalized classifier $z_k=\frac{\vw_k^{\top}}{\Vert \vw_k\Vert_2}\vf$, as well as the cosine classifier with $\sigma\in\{15, 20, 25, 30, 35\}$, and report the results in \Cref{table:comp_classifier}. 
The results show that the linear classifier performs well on ImageNet-LT and Places-LT, but unsatisfactorily on the more challenged iNaturalist 2018 dataset. This can be inferred from the classifier weight norms shown in \Cref{fig:weight_norm}, where the weight norms of iNaturalist 2018 are much more skewed.
By removing the impact of weight norms, the L2-normalized classifier achieves higher performance, especially on the tail classes. When adopting the cosine classifier, setting $\sigma$ to $25$ or $30$ leads to the best performance. Without loss of generality, we default to set $\sigma=25$.

\begin{table}[!h]
\caption{Performance of \algo\ with different classifiers.}
\label{table:comp_classifier}
\setlength{\tabcolsep}{1.2ex} 
\centering
\begin{small}
\begin{tabular}{l|c| C{6ex} C{4ex} C{4ex} C{4ex} | C{6ex} C{4ex} C{4ex} C{4ex} | C{6ex} C{4ex} C{4ex} C{4ex}}
\toprule
\multicolumn{2}{l|}{\multirow{2}{*}{\bf Classifiers}} & \multicolumn{4}{c|}{\bf ImageNet-LT} & \multicolumn{4}{c|}{\bf Places-LT} & \multicolumn{4}{c}{\bf iNaturalist 2018} \\ 
\multicolumn{2}{l|}{} &\bf Overall &\bf Head &\bf Med. &\bf Tail &\bf Overall &\bf Head &\bf Med. &\bf Tail &\bf Overall &\bf Head &\bf Med. &\bf Tail  \\
\midrule
\multicolumn{2}{l|}{Linear} & 78.2 & 81.2 & 77.2 & 72.8 &\bf 52.3 & 51.7 & 52.8 & 52.0 & 75.7 &\bf 75.8 & 77.4 & 73.6 \\
\multicolumn{2}{l|}{L2-normalized} & 78.4 & 81.2 & 77.2 &\bf 74.7 & 52.2 & 51.3 & 52.7 &\bf 52.4 & 80.0 & 74.1 & 79.9 & 81.8 \\
\multirow{5}{*}{Cosine} & $\sigma=15$ & 75.3 & 81.1 & 76.1 & 55.9 & 49.4 &\bf 52.8 & 52.8 & 35.3 & 76.5 & 73.4 & 76.4 & 77.4 \\
 & $\sigma=20$ & 77.5 & 81.1 & 77.1 & 69.1 & 51.6 & 52.2 &\bf 53.3 & 46.8 & 79.6 & 73.9 & 79.3 & 81.6 \\
 & $\sigma=25$ & 78.3 & 81.3 &\bf 77.4 & 73.4 & 52.2 & 51.7 & 53.1 & 50.9 &\bf 80.4 & 74.0 & 80.3 &\bf 82.2 \\
 & $\sigma=30$ &\bf 78.5 &\bf 81.5 & 77.3 & 74.2 & 52.1 & 51.5 & 52.6 & 51.8 & 80.3 & 73.8 &\bf 80.4 & 81.9 \\
 & $\sigma=35$ & 78.4 &\bf 81.5 & 77.1 & 73.9 & 51.7 & 51.3 & 52.1 & 51.7 & 79.8 & 74.1 & 79.9 & 81.3 \\
\bottomrule
\end{tabular}
\end{small}
\end{table}

\paragraph{Comparison of Different Losses.}
In \Cref{table:comp_loss}, we compare the performance of \algo\ with different losses, including cross-entropy (CE) loss, focal loss \citep{lin2017focal}, label-distribution-aware margin (LDAM) loss \citep{cao2019learning}, class-balanced (CB) loss \citep{cui2019class}, generalized re-weighting (GRW) loss \citep{zhang2021distribution}, label distribution disentangling (LADE) loss \citep{hong2021disentangling}. The results are shown in \Cref{table:comp_loss}, wherein the LA loss achieves the highest performance among all of the cases. In contrast, the other losses such as LDAM and LADE can not achieve satisfactory performance in all cases. Moreover, we give a theoretical proof of the LA loss and analyze the impact of the class-conditional distribution in \Cref{sec:la_loss},.

\begin{table}[!h]
\caption{Performance of \algo\ with different losses.}
\label{table:comp_loss}
\setlength{\tabcolsep}{1.2ex} 
\centering
\begin{small}
\begin{tabular}{l| C{6ex} C{4ex} C{4ex} C{4ex} | C{6ex} C{4ex} C{4ex} C{4ex} | C{6ex} C{4ex} C{4ex} C{4ex}}
\toprule
\multirow{2}{*}{\bf Losses} & \multicolumn{4}{c|}{\bf ImageNet-LT} & \multicolumn{4}{c|}{\bf Places-LT}  & \multicolumn{4}{c}{\bf iNaturalist 2018} \\ 
&\bf Overall &\bf Head &\bf Med. &\bf Tail &\bf Overall &\bf Head &\bf Med. &\bf Tail &\bf Overall &\bf Head &\bf Med. &\bf Tail \\
\midrule
CE & 71.8 & 86.1 & 68.7 & 42.1 & 42.1 & 56.7 & 38.0 & 24.5 & 74.8 & 82.1 & 75.4 & 72.0 \\
Focal \citep{lin2017focal} & 72.1 & 85.5 & 69.1 & 44.7 & 42.7 & 56.1 & 38.8 & 26.8 & 73.0 & 81.1 & 74.0 & 69.6 \\
LDAM \citep{cao2019learning} & 69.6 &\bf 86.4 & 66.5 & 33.4 & 40.4 &\bf 56.8 & 36.0 & 20.1 & 75.9 &\bf 84.3 & 77.0 & 72.4 \\
CB \citep{cui2019class} & 76.9 & 82.3 & 76.3 & 63.5 & 50.0 & 52.6 & 51.5 & 41.9 & 78.6 & 71.6 & 79.0 & 79.8 \\
GRW \citep{zhang2021distribution} & 76.9 & 82.3 & 76.3 & 63.7 & 50.1 & 52.4 & 51.7 & 42.0 & 78.6 & 71.9 & 79.1 & 79.8 \\
LADE \citep{hong2021disentangling} & 78.0 & 81.2 & 76.7 &\bf 73.4 & 51.2 & 51.3 & 51.7 & 49.6 &\bf 80.4 & 73.8 & 80.0 &\bf 82.5 \\
LA \citep{menon2021longtail} &\bf 78.3 & 81.3 &\bf 77.4 &\bf 73.4 &\bf 52.2 & 51.7 &\bf 53.1 &\bf 50.9 &\bf 80.4 & 74.0 &\bf 80.3 & 82.2 \\
\bottomrule
\end{tabular}
\end{small}
\end{table}

\paragraph{More Detailed Observations on Model Convergence.}

\begin{figure}[!h]
    \centering
    \includegraphics[width=0.195\linewidth]{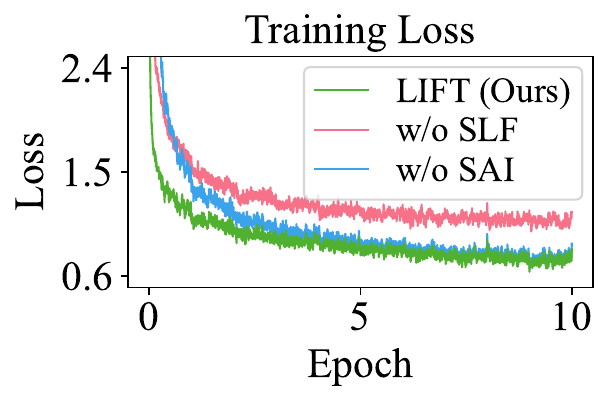}
    \includegraphics[width=0.195\linewidth]{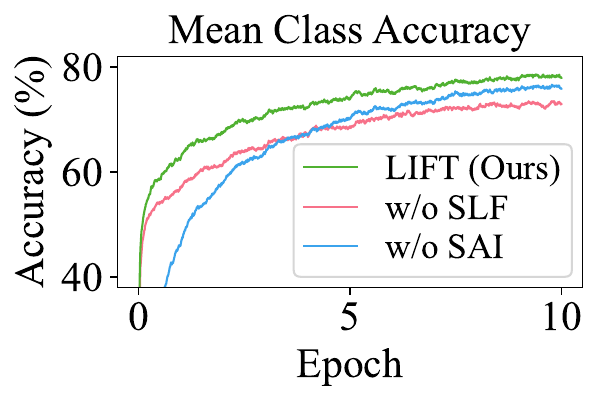}
    \includegraphics[width=0.195\linewidth]{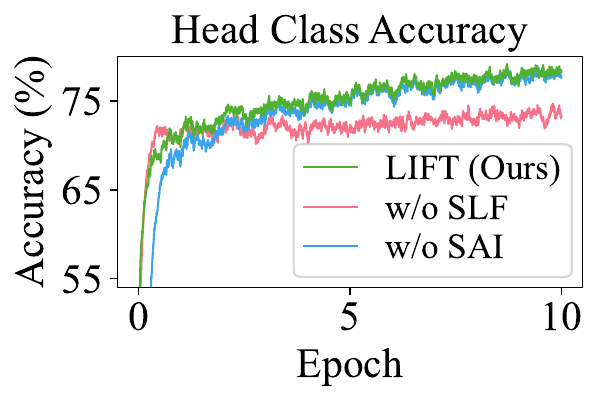}
    \includegraphics[width=0.195\linewidth]{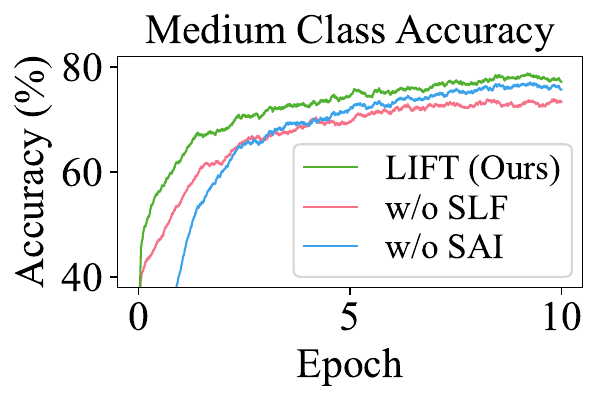}
    \includegraphics[width=0.195\linewidth]{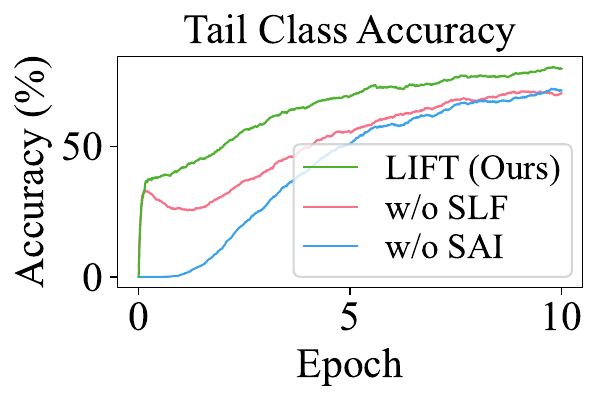}
    \caption{Convergence curves of training loss and accuracy on ImageNet-LT.}    
    \label{fig:convergence_imagenet_lt}
\end{figure}

\begin{figure}[!h]
    \centering
    \includegraphics[width=0.195\linewidth]{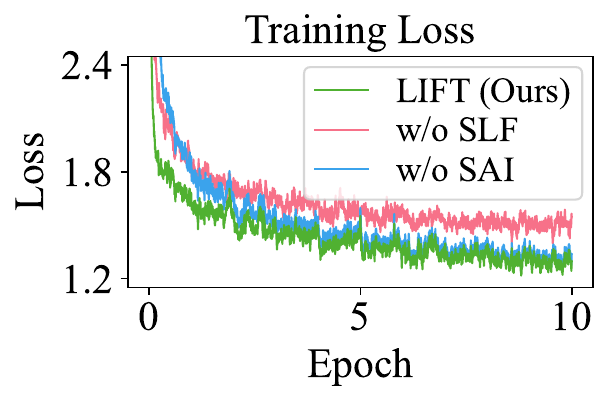}
    \includegraphics[width=0.195\linewidth]{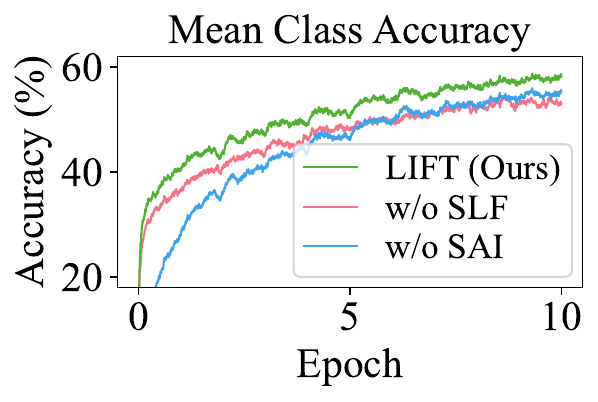}
    \includegraphics[width=0.195\linewidth]{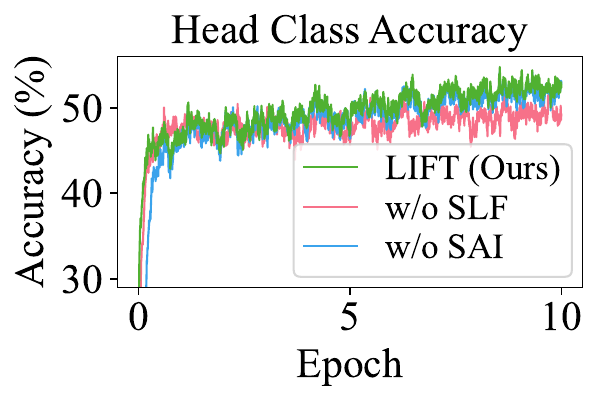}
    \includegraphics[width=0.195\linewidth]{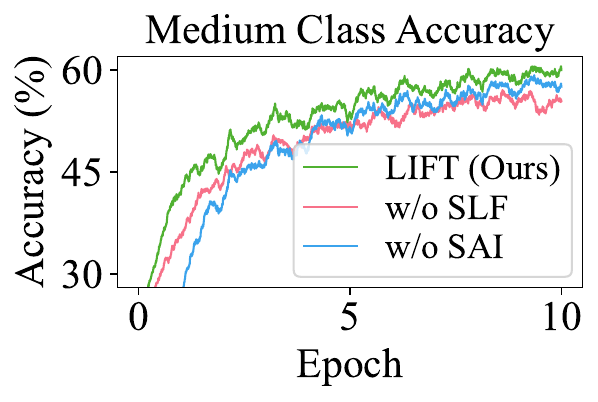}
    \includegraphics[width=0.195\linewidth]{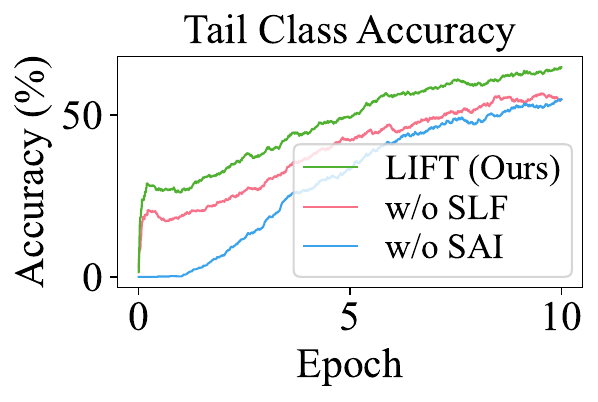}
    \caption{Convergence curves of training loss and accuracy on Places-LT.}
    \label{fig:convergence_places_lt}
\end{figure}

In \Cref{fig:convergence_imagenet_lt,fig:convergence_places_lt}, we illustrate the convergence curve on training loss and accuracy. We report the mean class accuracy, as well as the head, medium, and tail class accuracy. The results show that \algo\ converges rapidly with 10 training epochs. Without the structured lightweight fine-tuning (SLF) module, the training loss and accuracy converge suboptimally on all classes. Without semantic-aware initialization (SAI), the head-class accuracy is slightly affected, while the tail-class accuracy decreases by a large margin.

\paragraph{\algo\ with Variant Backbones.}

In addition to ViT-B/16, we also assess \algo\ based on the larger ViT-L/14. The results in \Cref{table:vitl_imagenetlt,table:vitl_placeslt,table:vitl_inat18,table:vitl_inat18_336} show that \algo\ surpasses the state-of-the-art method Decoder \citep{wang2023exploring} by a considerable margin, showcasing improvements of 3.1\% on ImageNet-LT, 5.0\% on Places-LT and 12.1\% on iNaturalist 2018. Additionally, the model with higher resolution (336$\times$336 pixels) yields improved performance. Furthermore, the incorporation of TTE consistently enhances the generalization. These results underscore the adaptability of \algo\ across variant backbones.

\begin{table}[!h]
\caption{Results on ImageNet-LT with ViT-L/14 as the backbone.}
\label{table:vitl_imagenetlt}
\setlength{\tabcolsep}{1.2ex} 
\centering
\begin{small}
\begin{tabular}{l|c|c|c|cccc}
\toprule
\bf Methods &\bf Backbone &\makecell{\bf Learnable \\ \bf Params.} &\bf \#Epochs &\bf Overall &\bf Head &\bf Medium &\bf Tail \\
\midrule
Zero-shot CLIP & ViT-L/14 & - & - & 73.6 & 74.6 & 73.1 & 72.3 \\
Decoder \citep{wang2023exploring} & ViT-L/14 & 39.79M & $\sim$18 & 79.3 & - & - & - \\
\algo\ (Ours) & ViT-L/14 &\bf 0.86M &\bf 10 & 82.4 & 84.8 & 81.7 & 78.0 \\
\algo\ w/ TTE (Ours) & ViT-L/14 &\bf 0.86M &\bf 10 &\bf 82.9 &\bf 85.3 &\bf 82.2 & \bf 78.6 \\
\bottomrule
\end{tabular}
\end{small}
\vskip -0.05in
\end{table}

\begin{table}[!h]
\caption{Results on Places-LT with ViT-L/14 as the backbone.}
\label{table:vitl_placeslt}
\setlength{\tabcolsep}{1.2ex} 
\centering
\begin{small}
\begin{tabular}{l|c|c|c|cccc}
\toprule
\bf Methods &\bf Backbone &\makecell{\bf Learnable \\ \bf Params.} &\bf \#Epochs &\bf Overall &\bf Head &\bf Medium &\bf Tail \\
\midrule
Zero-shot CLIP & ViT-L/14 & - & - & 39.9 & 38.1 & 39.2 & 45.1 \\
Decoder \citep{wang2023exploring} & ViT-L/14 & 39.79M & $\sim$34 & 48.4 & - & - & - \\
\algo\ (Ours) & ViT-L/14 &\bf 0.27M &\bf 10 & 53.4 & 52.6 &\bf 54.1 & 53.3 \\
\algo\ w/ TTE (Ours) & ViT-L/14 &\bf 0.27M &\bf 10 &\bf 53.7 &\bf 53.1 &\bf 54.1 & \bf 53.8 \\
\bottomrule
\end{tabular}
\end{small}
\vskip -0.05in
\end{table}

\begin{table}[!h]
\caption{Results on iNaturalist 2018 with ViT-L/14 as the backbone.}
\label{table:vitl_inat18}
\setlength{\tabcolsep}{1.2ex} 
\centering
\begin{small}
\begin{tabular}{l|c|c|c|cccc}
\toprule
\bf Methods &\bf Backbone &\makecell{\bf Learnable \\ \bf Params.} &\bf \#Epochs &\bf Overall &\bf Head &\bf Medium &\bf Tail \\
\midrule
Zero-shot CLIP & ViT-L/14 & - & - & 5.8 & 11.1 & 5.4 & 4.9 \\
Decoder \citep{wang2023exploring} & ViT-L/14 & 39.79M &\bf $\sim$5 & 72.3 & 65.5 & 73.2 & 73.0 \\
\algo\ (Ours) & ViT-L/14 &\bf 6.37M & 20 & 84.4 & 79.3 & 84.6 & 85.5 \\
\algo\ w/ TTE (Ours) & ViT-L/14 &\bf 6.37M & 20 &\bf 85.2 &\bf 80.2 &\bf 85.1 & \bf 86.6 \\
\bottomrule
\end{tabular}
\end{small}
\vskip -0.05in
\end{table}

\begin{table}[!h]
\caption{Results on iNaturalist 2018 with ViT-L/14 (336$\times$336 pixels) as the backbone.}
\label{table:vitl_inat18_336}
\setlength{\tabcolsep}{1.2ex} 
\centering
\begin{small}
\begin{tabular}{l|c|c|c|cccc}
\toprule
\bf Methods &\bf Backbone &\makecell{\bf Learnable \\ \bf Params.} &\bf \#Epochs &\bf Overall &\bf Head &\bf Medium &\bf Tail \\
\midrule
Zero-shot CLIP & ViT-L/14@336px & - & - & 6.2 & 11.5 & 5.8 & 5.3 \\
\algo\ (Ours) & ViT-L/14@336px &\bf 6.37M & 20 & 87.0 & 83.2 & 87.0 & 87.9 \\
\algo\ w/ TTE (Ours) & ViT-L/14@336px &\bf 6.37M & 20 &\bf 87.4 &\bf 83.6 &\bf 87.4 & \bf 88.3 \\
\bottomrule
\end{tabular}
\end{small}
\end{table}

\algo\ employs ViT as its backbone, and there may be concerns regarding the adoption of the widely used ResNet \citep{he2016deep}. However, due to the absence of a dedicated structured lightweight fine-tuning method tailored for ResNet, it is challenging to integrate our method with ResNet. Despite this limitation, we have explored some straightforward strategies, including 1) incorporating a scaling and shifting (SSF) \citep{lian2022scaling} module after the backbone, and 2) fine-tuning solely the bias terms of ResNet. The results are presented in \Cref{table:resnet_imagenetlt,table:resnet_placeslt}. In comparison to zero-shot CLIP and previous methods reported in \Cref{table:comp_imagenetlt,table:comp_placeslt}, our method achieves significantly superior performance with lower computational costs.

\begin{table}[!h]
\caption{Results on ImageNet-LT with ResNet-50 as the backbone. All methods use TTE for fair comparison.}
\label{table:resnet_imagenetlt}
\setlength{\tabcolsep}{1.2ex} 
\centering
\begin{small}
\begin{tabular}{l|c|c|c|cccc}
\toprule
\bf Methods &\bf Backbone &\makecell{\bf Learnable \\ \bf Params.} &\bf \#Epochs &\bf Overall &\bf Head &\bf Medium &\bf Tail \\
\midrule
Zero-shot CLIP & ResNet-50 & - & - & 57.6 & 58.6 & 56.9 & 56.9 \\
\algo\ w/ SSF & ResNet-50 & 0.002M & 10 & 66.9 & 72.0 & 66.5 & 54.1 \\
\algo\ w/ bias tuning & ResNet-50 & 0.034M & 10 & 67.8 & 72.0 & 67.3 & 57.6 \\
\algo\ w/ bias tuning \& SSF & ResNet-50 & 0.036M & 10 &\bf 68.3 &\bf 72.5 &\bf 67.8 & \bf 58.2 \\
\bottomrule
\end{tabular}
\end{small}
\vskip -0.05in
\end{table}

\begin{table}[!h]
\caption{Results on Places-LT with ResNet-50 as the backbone. All methods use TTE for fair comparison.}
\label{table:resnet_placeslt}
\setlength{\tabcolsep}{1.2ex} 
\centering
\begin{small}
\begin{tabular}{l|c|c|c|cccc}
\toprule
\bf Methods &\bf Backbone &\makecell{\bf Learnable \\ \bf Params.} &\bf \#Epochs &\bf Overall &\bf Head &\bf Medium &\bf Tail \\
\midrule
Zero-shot CLIP & ResNet-50 & - & - & 35.2 & 33.1 & 34.6 & 40.4 \\
\algo\ w/ SSF & ResNet-50 & 0.002M & 10 & 46.7 & 47.5 & 48.7 & 40.7 \\
\algo\ w/ bias tuning & ResNet-50 & 0.034M & 10 & 47.9 &\bf 48.2 & 49.8 & 42.9 \\
\algo\ w/ bias tuning \& SSF & ResNet-50 & 0.036M & 10 &\bf 48.1 & 48.1 &\bf 50.0 & \bf 43.9 \\
\bottomrule
\end{tabular}
\end{small}
\end{table}

Apart from the vision-language model CLIP, we have also validated LIFT using the ImageNet-21K pre-training model, which is a vision-only model. We employ the class mean features to initialize the classifier due to the lack of a corresponding text encoder. The results are provided in \Cref{table:comp_imagenetlt_in21k,table:comp_placeslt_in21k,table:comp_inat18_in21k}. It is worth noting that the superior performance on ImageNet-LT may attributed to potential data leakage from ImageNet-21K. The performance on Places-LT is lower than CLIP pre-training, while still outperforming most state-of-the-art methods in \Cref{table:comp_placeslt}. The performance on iNaturalist 2018 exceeds that of CLIP pre-training and outperforms all state-of-the-art methods in \Cref{table:comp_inat18}.

\begin{table}[!h]
\caption{Results of \algo\ on ImageNet-LT with different pre-training models. All methods use TTE for fair comparison.}
\label{table:comp_imagenetlt_in21k}
\setlength{\tabcolsep}{1.2ex} 
\centering
\begin{small}
\begin{tabular}{l|c|c|c|cccc}
\toprule
 &\bf Backbone &\makecell{\bf Learnable \\ \bf Params.} &\bf \#Epochs &\bf Overall &\bf Head &\bf Medium &\bf Tail \\
\midrule
CLIP Pre-training & ViT-B/16 & 0.62M & 10 & 78.3 & 81.3 & 77.4 & 73.4 \\
ImageNet-21K Pre-training & ViT-B/16 & 0.62M & 10 &\bf 84.2 &\bf 86.0 &\bf 83.6 &\bf 80.6 \\
\bottomrule
\end{tabular}
\end{small}
\vskip -0.05in
\end{table}

\begin{table}[!h]
\caption{Results of \algo\ on Places-LT with different pre-training models. All methods use TTE for fair comparison.}
\label{table:comp_placeslt_in21k}
\setlength{\tabcolsep}{1.2ex} 
\centering
\begin{small}
\begin{tabular}{l|c|c|c|cccc}
\toprule
 &\bf Backbone &\makecell{\bf Learnable \\ \bf Params.} &\bf \#Epochs &\bf Overall &\bf Head &\bf Medium &\bf Tail \\
\midrule
CLIP Pre-training & ViT-B/16 & 0.18M & 10 &\bf 52.2 &\bf 51.7 &\bf 53.1 &\bf 50.9 \\
ImageNet-21K Pre-training & ViT-B/16 & 0.18M & 10 & 49.2 & 48.9 & 50.2 & 47.4 \\
\bottomrule
\end{tabular}
\end{small}
\vskip -0.05in
\end{table}

\begin{table}[!h]
\caption{Results of \algo\ on iNaturalist 2018 with different pre-training models. All methods use TTE for fair comparison.}
\label{table:comp_inat18_in21k}
\setlength{\tabcolsep}{1.2ex} 
\renewcommand{\arraystretch}{0.95}
\centering
\begin{small}
\begin{tabular}{l|c|c|c|cccc}
\toprule
 &\bf Backbone &\makecell{\bf Learnable \\ \bf Params.} &\bf \#Epochs &\bf Overall &\bf Head &\bf Medium &\bf Tail \\
\midrule
CLIP Pre-training & ViT-B/16 & 4.75M & 20 & 80.4 & 74.0 & 80.3 & 82.2 \\
ImageNet-21K Pre-training & ViT-B/16 & 4.75M & 20 &\bf 81.9 &\bf 74.9 &\bf 82.3 &\bf 83.3 \\
\bottomrule
\end{tabular}
\end{small}
\end{table}

\section{Related Work}

\paragraph{Long-Tail Learning via Deep Learning.} Conventional methods train convolutional neural network models like ResNet and ResNeXt on long-tail datasets. Concerning the class imbalance, there are three main directions to improve the performance: 
1) data manipulation \citep{zhou2020bbn,kang2020decoupling,yang2020rethinking,he2021distilling,park2022majority,ahn2023cuda,shi2023re,gao2023enhancing}, 2) representation learning \citep{liu2019large,kang2021exploring,wang2021contrastive,cui2021parametric,samuel2021distributional,zhu2022balanced,yang2022inducing,peifeng2023feature,ma2023curvature}, and 3) model output adjustment \citep{cao2019learning,ren2020balanced,menon2021longtail,hong2021disentangling,zhang2021distribution,wei2022robust,han2023wrapped,shi2024residual}. Data manipulation typically includes designing re-sampling strategies, and data augmentations. Many works improve the performance by adopting two-stage training where the first stage learns representations and the second stage learns the classifier \citep{zhong2021improving,wei2023towards,nam2023decoupled}. The adjustment of the model's outputs can be done during training by optimizing unbiased loss functions or after training.
In contrast to the aforementioned works, this paper presents an end-to-end training framework that combines the advantages of foundation models and multiple existing techniques. We conduct in-depth research on how to properly utilize the foundation models and enable the unbiased loss function to achieve optimal effects.

\paragraph{Long-Tail Learning via Foundation Model.} Fine-tuning foundation models such as CLIP \citep{radford2021clip} and ViT \citep{dosovitskiy2021an} has attracted widespread attention \citep{steiner2021train,zhou2022learning,zhou2022conditional,yu2023visual,jia2024lamda,zhou2024decoop}, and has emerged as an effective strategy to address class imbalance due to the strong representation learning capabilities \citep{ma2021simple,long2022retrieval,tian2022vl,iscen2023improving,dong2023lpt,xia2023lmpt,he2023uniformly,song2023long,wang2023exploring,li2024rectify}. However, it is important to note that these methods often require prolonged training time and, in some cases, rely on external training data to facilitate the learning process. In contrast, our proposed approach exhibits the remarkable ability to achieve convergence in fewer than 20 epochs and does not need external data. Furthermore, our method is general, allowing for seamless integration with various lightweight fine-tuning approaches.

\section{Limitations}

\paragraph{Limitations of Arbitrary Lightweight Fine-Tuning.}
Despite the remarkable performance achieved through arbitrary lightweight fine-tuning, this approach exhibits a slightly slower training process (3'10'' on ImageNet-LT and 1'43'' on Places-LT per epoch). When compared to structured lightweight fine-tuning methods in \Cref{table:time_cost}, its time cost is higher (approximately 1.2$\times$). This is because the GPU has challenges in accelerating computation with unstructured parameters. Nonetheless, the proposed arbitrary lightweight fine-tuning is capable of achieving rapid convergence within 20 epochs, which is much more efficient compared to prior works. We will focus on enhancing the efficiency in our future work. In this paper, we propose arbitrary lightweight fine-tuning primarily to demonstrate the effectiveness of lightweight fine-tuning, as even arbitrarily selected parameters without guidance can still achieve superior performance.

\paragraph{Limitations of Semantic-Aware Initialization.}
In this paper, we propose semantic-aware initialization to leverage the semantic knowledge from CLIP and enhance the initialization. However, when deploying vision-only foundation models, integrating semantic knowledge becomes a challenging task. In such scenarios, we have identified that utilizing class mean features proves to be a viable choice. As delineated in \Cref{table:comp_clf_init}, this approach leads to remarkable performance improvements, surpassing most of the existing methods in \Cref{table:comp_imagenetlt,table:comp_placeslt}. Additionally, \Cref{table:comp_cifar100lt,table:comp_imagenetlt_in21k,table:comp_placeslt_in21k,table:comp_inat18_in21k} also demonstrate the effectiveness of class mean features when adopted to ImageNet-21K pre-training vision models. It remains an intriguing challenge how to initialize the classifier for visual-only foundation models with long-tail data. Nonetheless, our findings demonstrate that opting for class mean features is an effective approach.


\end{document}

%% file: math_commands.tex
\usepackage{amsmath,amsfonts,bm}

\def\eqref#1{equation~\ref{#1}}

\def\1{\bm{1}}

\def\vb{{\bm{b}}}
\def\vc{{\bm{c}}}

\def\vf{{\bm{f}}}

\def\vw{{\bm{w}}}
\def\vx{{\bm{x}}}
\def\vy{{\bm{y}}}
\def\vz{{\bm{z}}}

\def\mE{{\bm{E}}}

\def\mM{{\bm{M}}}

\def\mP{{\bm{P}}}

\def\mW{{\bm{W}}}
\def\mX{{\bm{X}}}

\DeclareMathAlphabet{\mathsfit}{\encodingdefault}{\sfdefault}{m}{sl}
\SetMathAlphabet{\mathsfit}{bold}{\encodingdefault}{\sfdefault}{bx}{n}

\def\gL{{\mathcal{L}}}

\def\gO{{\mathcal{O}}}

\def\sR{{\mathbb{R}}}










%% file: paper.bbl
\begin{thebibliography}{72}
\providecommand{\natexlab}[1]{#1}
\providecommand{\url}[1]{\texttt{#1}}
\expandafter\ifx\csname urlstyle\endcsname\relax
  \providecommand{\doi}[1]{doi: #1}\else
  \providecommand{\doi}{doi: \begingroup \urlstyle{rm}\Url}\fi

\bibitem[Ahn et~al.(2023)Ahn, Ko, and Yun]{ahn2023cuda}
Ahn, S., Ko, J., and Yun, S.-Y.
\newblock {CUDA}: Curriculum of data augmentation for long-tailed recognition.
\newblock In \emph{International Conference on Learning Representations}, 2023.

\bibitem[Ba et~al.(2016)Ba, Kiros, and Hinton]{ba2016layer}
Ba, J.~L., Kiros, J.~R., and Hinton, G.~E.
\newblock Layer normalization.
\newblock \emph{arXiv preprint arXiv:1607.06450}, 2016.

\bibitem[Cao et~al.(2019)Cao, Wei, Gaidon, Arechiga, and Ma]{cao2019learning}
Cao, K., Wei, C., Gaidon, A., Arechiga, N., and Ma, T.
\newblock Learning imbalanced datasets with label-distribution-aware margin loss.
\newblock In \emph{Advances in Neural Information Processing Systems}, volume~32, pp.\  1565--1576, 2019.

\bibitem[Chen et~al.(2022)Chen, GE, Tong, Wang, Song, Wang, and Luo]{chen2022adaptformer}
Chen, S., GE, C., Tong, Z., Wang, J., Song, Y., Wang, J., and Luo, P.
\newblock Adaptformer: Adapting vision transformers for scalable visual recognition.
\newblock In \emph{Advances in Neural Information Processing Systems}, volume~35, pp.\  16664--16678, 2022.

\bibitem[Cui et~al.(2021)Cui, Zhong, Liu, Yu, and Jia]{cui2021parametric}
Cui, J., Zhong, Z., Liu, S., Yu, B., and Jia, J.
\newblock Parametric contrastive learning.
\newblock In \emph{Proceedings of the IEEE/CVF International Conference on Computer Vision}, pp.\  715--724, 2021.

\bibitem[Cui et~al.(2019)Cui, Jia, Lin, Song, and Belongie]{cui2019class}
Cui, Y., Jia, M., Lin, T.-Y., Song, Y., and Belongie, S.
\newblock Class-balanced loss based on effective number of samples.
\newblock In \emph{Proceedings of the IEEE/CVF Conference on Computer Vision and Pattern Recognition}, pp.\  9268--9277, 2019.

\bibitem[Devlin et~al.(2019)Devlin, Chang, Lee, and Toutanova]{kenton2019bert}
Devlin, J., Chang, M., Lee, K., and Toutanova, K.
\newblock {BERT}: Pre-training of deep bidirectional transformers for language understanding.
\newblock In \emph{Proceedings of NAACL-HLT}, pp.\  4171--4186, 2019.

\bibitem[Dong et~al.(2023)Dong, Zhou, Yan, and Zuo]{dong2023lpt}
Dong, B., Zhou, P., Yan, S., and Zuo, W.
\newblock {LPT}: Long-tailed prompt tuning for image classification.
\newblock In \emph{International Conference on Learning Representations}, 2023.

\bibitem[Dosovitskiy et~al.(2021)Dosovitskiy, Beyer, Kolesnikov, Weissenborn, Zhai, Unterthiner, Dehghani, Minderer, Heigold, Gelly, Uszkoreit, and Houlsby]{dosovitskiy2021an}
Dosovitskiy, A., Beyer, L., Kolesnikov, A., Weissenborn, D., Zhai, X., Unterthiner, T., Dehghani, M., Minderer, M., Heigold, G., Gelly, S., Uszkoreit, J., and Houlsby, N.
\newblock An image is worth 16x16 words: Transformers for image recognition at scale.
\newblock In \emph{International Conference on Learning Representations}, 2021.

\bibitem[Gao et~al.(2023)Gao, Zhao, Li, and Guo]{gao2023enhancing}
Gao, J., Zhao, H., Li, Z., and Guo, D.
\newblock Enhancing minority classes by mixing: An adaptative optimal transport approach for long-tailed classification.
\newblock In \emph{Advances in Neural Information Processing Systems}, volume~36, pp.\  60329--60348, 2023.

\bibitem[Han(2023)]{han2023wrapped}
Han, B.
\newblock Wrapped cauchy distributed angular softmax for long-tailed visual recognition.
\newblock In \emph{Proceedings of the 40th International Conference on Machine Learning}, pp.\  12368--12388, 2023.

\bibitem[He et~al.(2016)He, Zhang, Ren, and Sun]{he2016deep}
He, K., Zhang, X., Ren, S., and Sun, J.
\newblock Deep residual learning for image recognition.
\newblock In \emph{Proceedings of the IEEE Conference on Computer Vision and Pattern Recognition}, pp.\  770--778, 2016.

\bibitem[He et~al.(2022)He, Chen, Xie, Li, Doll\'ar, and Girshick]{he2022masked}
He, K., Chen, X., Xie, S., Li, Y., Doll\'ar, P., and Girshick, R.
\newblock Masked autoencoders are scalable vision learners.
\newblock In \emph{Proceedings of the IEEE/CVF Conference on Computer Vision and Pattern Recognition}, pp.\  16000--16009, 2022.

\bibitem[He et~al.(2023)He, Fu, Ding, Cao, and Wang]{he2023uniformly}
He, X., Fu, S., Ding, X., Cao, Y., and Wang, H.
\newblock Uniformly distributed category prototype-guided vision-language framework for long-tail recognition.
\newblock In \emph{Proceedings of the 31st ACM International Conference on Multimedia}, pp.\  5027--5037, 2023.

\bibitem[He et~al.(2021)He, Wu, and Wei]{he2021distilling}
He, Y.-Y., Wu, J., and Wei, X.-S.
\newblock Distilling virtual examples for long-tailed recognition.
\newblock In \emph{Proceedings of the IEEE/CVF International Conference on Computer Vision}, pp.\  235--244, 2021.

\bibitem[Hong et~al.(2021)Hong, Han, Choi, Seo, Kim, and Chang]{hong2021disentangling}
Hong, Y., Han, S., Choi, K., Seo, S., Kim, B., and Chang, B.
\newblock Disentangling label distribution for long-tailed visual recognition.
\newblock In \emph{Proceedings of the IEEE/CVF Conference on Computer Vision and Pattern Recognition}, pp.\  6626--6636, 2021.

\bibitem[Houlsby et~al.(2019)Houlsby, Giurgiu, Jastrzebski, Morrone, De~Laroussilhe, Gesmundo, Attariyan, and Gelly]{houlsby2019parameter}
Houlsby, N., Giurgiu, A., Jastrzebski, S., Morrone, B., De~Laroussilhe, Q., Gesmundo, A., Attariyan, M., and Gelly, S.
\newblock Parameter-efficient transfer learning for {NLP}.
\newblock In \emph{Proceedings of the 36th International Conference on Machine Learning}, pp.\  2790--2799, 2019.

\bibitem[Hu et~al.(2022)Hu, yelong shen, Wallis, Allen-Zhu, Li, Wang, Wang, and Chen]{hu2022lora}
Hu, E.~J., yelong shen, Wallis, P., Allen-Zhu, Z., Li, Y., Wang, S., Wang, L., and Chen, W.
\newblock Lo{RA}: Low-rank adaptation of large language models.
\newblock In \emph{International Conference on Learning Representations}, 2022.

\bibitem[Iscen et~al.(2023)Iscen, Fathi, and Schmid]{iscen2023improving}
Iscen, A., Fathi, A., and Schmid, C.
\newblock Improving image recognition by retrieving from web-scale image-text data.
\newblock In \emph{Proceedings of the IEEE/CVF Conference on Computer Vision and Pattern Recognition}, pp.\  19295--19304, 2023.

\bibitem[Jia et~al.(2024)Jia, Guo, Zhou, and Li]{jia2024lamda}
Jia, L.-H., Guo, L.-Z., Zhou, Z., and Li, Y.-F.
\newblock {LAMDA-SSL}: A comprehensive semi-supervised learning toolkit.
\newblock \emph{SCIENCE CHINA Information Sciences}, 67\penalty0 (1):\penalty0 117101, 2024.

\bibitem[Jia et~al.(2022)Jia, Tang, Chen, Cardie, Belongie, Hariharan, and Lim]{jia2022visual}
Jia, M., Tang, L., Chen, B.-C., Cardie, C., Belongie, S., Hariharan, B., and Lim, S.-N.
\newblock Visual prompt tuning.
\newblock In \emph{Proceedings of the 17th European Conference on Computer Vision}, pp.\  709--727, 2022.

\bibitem[Kang et~al.(2020)Kang, Xie, Rohrbach, Yan, Gordo, Feng, and Kalantidis]{kang2020decoupling}
Kang, B., Xie, S., Rohrbach, M., Yan, Z., Gordo, A., Feng, J., and Kalantidis, Y.
\newblock Decoupling representation and classifier for long-tailed recognition.
\newblock In \emph{International Conference on Learning Representations}, 2020.

\bibitem[Kang et~al.(2021)Kang, Li, Xie, Yuan, and Feng]{kang2021exploring}
Kang, B., Li, Y., Xie, S., Yuan, Z., and Feng, J.
\newblock Exploring balanced feature spaces for representation learning.
\newblock In \emph{International Conference on Learning Representations}, 2021.

\bibitem[Kumar et~al.(2022)Kumar, Raghunathan, Jones, Ma, and Liang]{kumar2022finetuning}
Kumar, A., Raghunathan, A., Jones, R.~M., Ma, T., and Liang, P.
\newblock Fine-tuning can distort pretrained features and underperform out-of-distribution.
\newblock In \emph{International Conference on Learning Representations}, 2022.

\bibitem[Li et~al.(2024)Li, Yao, Tan, Gong, Lu, and Luo]{li2024rectify}
Li, B., Yao, Y., Tan, J., Gong, R., Lu, J., and Luo, Y.
\newblock Rectify representation bias in vision-language models for long-tailed recognition.
\newblock \emph{Neural Networks}, 172:\penalty0 106134, 2024.

\bibitem[Li et~al.(2022{\natexlab{a}})Li, Tan, Wan, Lei, and Guo]{li2022nested}
Li, J., Tan, Z., Wan, J., Lei, Z., and Guo, G.
\newblock Nested collaborative learning for long-tailed visual recognition.
\newblock In \emph{Proceedings of the IEEE/CVF Conference on Computer Vision and Pattern Recognition}, pp.\  6949--6958, 2022{\natexlab{a}}.

\bibitem[Li et~al.(2022{\natexlab{b}})Li, Cheung, and Lu]{li2022long}
Li, M., Cheung, Y.-m., and Lu, Y.
\newblock Long-tailed visual recognition via gaussian clouded logit adjustment.
\newblock In \emph{Proceedings of the IEEE/CVF Conference on Computer Vision and Pattern Recognition}, pp.\  6929--6938, 2022{\natexlab{b}}.

\bibitem[Lian et~al.(2022)Lian, Zhou, Feng, and Wang]{lian2022scaling}
Lian, D., Zhou, D., Feng, J., and Wang, X.
\newblock Scaling \&amp; shifting your features: A new baseline for efficient model tuning.
\newblock In \emph{Advances in Neural Information Processing Systems}, volume~35, pp.\  109--123, 2022.

\bibitem[Lin et~al.(2017)Lin, Goyal, Girshick, He, and Doll{\'a}r]{lin2017focal}
Lin, T.-Y., Goyal, P., Girshick, R., He, K., and Doll{\'a}r, P.
\newblock Focal loss for dense object detection.
\newblock In \emph{Proceedings of the IEEE International Conference on Computer Vision}, pp.\  2980--2988, 2017.

\bibitem[Liu et~al.(2019)Liu, Miao, Zhan, Wang, Gong, and Yu]{liu2019large}
Liu, Z., Miao, Z., Zhan, X., Wang, J., Gong, B., and Yu, S.~X.
\newblock Large-scale long-tailed recognition in an open world.
\newblock In \emph{Proceedings of the IEEE/CVF Conference on Computer Vision and Pattern Recognition}, pp.\  2537--2546, 2019.

\bibitem[Long et~al.(2022)Long, Yin, Ajanthan, Nguyen, Purkait, Garg, Blair, Shen, and van~den Hengel]{long2022retrieval}
Long, A., Yin, W., Ajanthan, T., Nguyen, V., Purkait, P., Garg, R., Blair, A., Shen, C., and van~den Hengel, A.
\newblock Retrieval augmented classification for long-tail visual recognition.
\newblock In \emph{Proceedings of the IEEE/CVF Conference on Computer Vision and Pattern Recognition}, pp.\  6959--6969, 2022.

\bibitem[Ma et~al.(2021)Ma, Geng, Wang, Shao, Lu, Li, Gao, and Qiao]{ma2021simple}
Ma, T., Geng, S., Wang, M., Shao, J., Lu, J., Li, H., Gao, P., and Qiao, Y.
\newblock A simple long-tailed recognition baseline via vision-language model.
\newblock \emph{arXiv preprint arXiv:2111.14745}, 2021.

\bibitem[Ma et~al.(2023)Ma, Jiao, Liu, Yang, Liu, and Li]{ma2023curvature}
Ma, Y., Jiao, L., Liu, F., Yang, S., Liu, X., and Li, L.
\newblock Curvature-balanced feature manifold learning for long-tailed classification.
\newblock In \emph{Proceedings of the IEEE/CVF Conference on Computer Vision and Pattern Recognition}, pp.\  15824--15835, 2023.

\bibitem[Menon et~al.(2021)Menon, Jayasumana, Rawat, Jain, Veit, and Kumar]{menon2021longtail}
Menon, A.~K., Jayasumana, S., Rawat, A.~S., Jain, H., Veit, A., and Kumar, S.
\newblock Long-tail learning via logit adjustment.
\newblock In \emph{International Conference on Learning Representations}, 2021.

\bibitem[Menon \& Vondrick(2023)Menon and Vondrick]{menon2023visual}
Menon, S. and Vondrick, C.
\newblock Visual classification via description from large language models.
\newblock In \emph{International Conference on Learning Representations}, 2023.

\bibitem[Nam et~al.(2023)Nam, Jang, and Lee]{nam2023decoupled}
Nam, G., Jang, S., and Lee, J.
\newblock Decoupled training for long-tailed classification with stochastic representations.
\newblock In \emph{International Conference on Learning Representations}, 2023.

\bibitem[Park et~al.(2022)Park, Hong, Heo, Yun, and Choi]{park2022majority}
Park, S., Hong, Y., Heo, B., Yun, S., and Choi, J.~Y.
\newblock The majority can help the minority: Context-rich minority oversampling for long-tailed classification.
\newblock In \emph{Proceedings of the IEEE/CVF Conference on Computer Vision and Pattern Recognition}, pp.\  6887--6896, 2022.

\bibitem[Peifeng et~al.(2023)Peifeng, Xu, Wen, Yang, Shao, and Huang]{peifeng2023feature}
Peifeng, G., Xu, Q., Wen, P., Yang, Z., Shao, H., and Huang, Q.
\newblock Feature directions matter: Long-tailed learning via rotated balanced representation.
\newblock In \emph{Proceedings of the 40th International Conference on Machine Learning}, pp.\  27542--27563, 2023.

\bibitem[Radford et~al.(2021)Radford, Kim, Hallacy, Ramesh, Goh, Agarwal, Sastry, Askell, Mishkin, Clark, et~al.]{radford2021clip}
Radford, A., Kim, J.~W., Hallacy, C., Ramesh, A., Goh, G., Agarwal, S., Sastry, G., Askell, A., Mishkin, P., Clark, J., et~al.
\newblock Learning transferable visual models from natural language supervision.
\newblock In \emph{Proceedings of the 38th International Conference on Machine Learning}, pp.\  8748--8763, 2021.

\bibitem[Ren et~al.(2020)Ren, Yu, sheng, Ma, Zhao, Yi, and Li]{ren2020balanced}
Ren, J., Yu, C., sheng, s., Ma, X., Zhao, H., Yi, S., and Li, h.
\newblock Balanced meta-softmax for long-tailed visual recognition.
\newblock In \emph{Advances in Neural Information Processing Systems}, volume~33, pp.\  4175--4186, 2020.

\bibitem[Samuel \& Chechik(2021)Samuel and Chechik]{samuel2021distributional}
Samuel, D. and Chechik, G.
\newblock Distributional robustness loss for long-tail learning.
\newblock In \emph{Proceedings of the IEEE/CVF International Conference on Computer Vision}, pp.\  9495--9504, 2021.

\bibitem[Shi et~al.(2023)Shi, Wei, Xiang, and Li]{shi2023re}
Shi, J.-X., Wei, T., Xiang, Y., and Li, Y.-F.
\newblock How re-sampling helps for long-tail learning?
\newblock In \emph{Advances in Neural Information Processing Systems}, volume~36, pp.\  75669--75687, 2023.

\bibitem[Shi et~al.(2024)Shi, Wei, and Li]{shi2024residual}
Shi, J.-X., Wei, T., and Li, Y.-F.
\newblock Residual diverse ensemble for long-tailed multi-label text classification.
\newblock \emph{SCIENCE CHINA Information Sciences}, 2024.

\bibitem[Song et~al.(2023)Song, Li, and Wang]{song2023long}
Song, Y., Li, M., and Wang, B.
\newblock Long-tailed visual recognition via improved cross-window self-attention and trivialaugment.
\newblock \emph{IEEE Access}, 11:\penalty0 49601--49610, 2023.

\bibitem[Steiner et~al.(2021)Steiner, Kolesnikov, Zhai, Wightman, Uszkoreit, and Beyer]{steiner2021train}
Steiner, A., Kolesnikov, A., Zhai, X., Wightman, R., Uszkoreit, J., and Beyer, L.
\newblock How to train your vit? data, augmentation, and regularization in vision transformers.
\newblock \emph{arXiv preprint arXiv:2106.10270}, 2021.

\bibitem[Suh \& Seo(2023)Suh and Seo]{suh2023long}
Suh, M.-K. and Seo, S.-W.
\newblock Long-tailed recognition by mutual information maximization between latent features and ground-truth labels.
\newblock In \emph{Proceedings of the 40th International Conference on Machine Learning}, volume 202, pp.\  32770--32782, 2023.

\bibitem[Sun et~al.(2020)Sun, Wang, Liu, Miller, Efros, and Hardt]{sun2020test}
Sun, Y., Wang, X., Liu, Z., Miller, J., Efros, A., and Hardt, M.
\newblock Test-time training with self-supervision for generalization under distribution shifts.
\newblock In \emph{Proceedings of the 37th International Conference on Machine Learning}, pp.\  9229--9248, 2020.

\bibitem[Tian et~al.(2022)Tian, Wang, Zhu, Dai, and Qiao]{tian2022vl}
Tian, C., Wang, W., Zhu, X., Dai, J., and Qiao, Y.
\newblock {VL-LTR:} learning class-wise visual-linguistic representation for long-tailed visual recognition.
\newblock In \emph{Proceedings of the 17th European Conference on Computer Vision}, pp.\  73--91, 2022.

\bibitem[Van~Horn et~al.(2018)Van~Horn, Mac~Aodha, Song, Cui, Sun, Shepard, Adam, Perona, and Belongie]{van2018inaturalist}
Van~Horn, G., Mac~Aodha, O., Song, Y., Cui, Y., Sun, C., Shepard, A., Adam, H., Perona, P., and Belongie, S.
\newblock The inaturalist species classification and detection dataset.
\newblock In \emph{Proceedings of the IEEE Conference on Computer Vision and Pattern Recognition}, pp.\  8769--8778, 2018.

\bibitem[Vaswani et~al.(2017)Vaswani, Shazeer, Parmar, Uszkoreit, Jones, Gomez, Kaiser, and Polosukhin]{vaswani2017attention}
Vaswani, A., Shazeer, N., Parmar, N., Uszkoreit, J., Jones, L., Gomez, A.~N., Kaiser, {\L}., and Polosukhin, I.
\newblock Attention is all you need.
\newblock In \emph{Advances in Neural Information Processing Systems}, volume~30, pp.\  5998--6008, 2017.

\bibitem[Wang et~al.(2021{\natexlab{a}})Wang, Shelhamer, Liu, Olshausen, and Darrell]{wang2021tent}
Wang, D., Shelhamer, E., Liu, S., Olshausen, B., and Darrell, T.
\newblock Tent: Fully test-time adaptation by entropy minimization.
\newblock In \emph{International Conference on Learning Representations}, 2021{\natexlab{a}}.

\bibitem[Wang et~al.(2021{\natexlab{b}})Wang, Han, Wei, Zhang, and Wang]{wang2021contrastive}
Wang, P., Han, K., Wei, X.-S., Zhang, L., and Wang, L.
\newblock Contrastive learning based hybrid networks for long-tailed image classification.
\newblock In \emph{Proceedings of the IEEE/CVF Conference on Computer Vision and Pattern Recognition}, pp.\  943--952, 2021{\natexlab{b}}.

\bibitem[Wang et~al.(2021{\natexlab{c}})Wang, Lian, Miao, Liu, and Yu]{wang2021longtailed}
Wang, X., Lian, L., Miao, Z., Liu, Z., and Yu, S.
\newblock Long-tailed recognition by routing diverse distribution-aware experts.
\newblock In \emph{International Conference on Learning Representations}, 2021{\natexlab{c}}.

\bibitem[Wang et~al.(2024)Wang, Yu, Wang, Heng, Chen, Ye, Xie, Xie, and Zhang]{wang2023exploring}
Wang, Y., Yu, Z., Wang, J., Heng, Q., Chen, H., Ye, W., Xie, R., Xie, X., and Zhang, S.
\newblock Exploring vision-language models for imbalanced learning.
\newblock \emph{International Journal of Computer Vision}, 132:\penalty0 224--237, 2024.

\bibitem[Wei \& Gan(2023)Wei and Gan]{wei2023towards}
Wei, T. and Gan, K.
\newblock Towards realistic long-tailed semi-supervised learning: Consistency is all you need.
\newblock In \emph{Proceedings of the IEEE/CVF Conference on Computer Vision and Pattern Recognition}, pp.\  3469--3478, 2023.

\bibitem[Wei et~al.(2021)Wei, Tu, Li, and Yang]{wei2021towards}
Wei, T., Tu, W.-W., Li, Y.-F., and Yang, G.-P.
\newblock Towards robust prediction on tail labels.
\newblock In \emph{Proceedings of the 27th ACM SIGKDD Conference on Knowledge Discovery \& Data Mining}, pp.\  1812--1820, 2021.

\bibitem[Wei et~al.(2022)Wei, Wang, Tu, and Li]{wei2022robust}
Wei, T., Wang, H., Tu, W.-W., and Li, Y.-F.
\newblock Robust model selection for positive and unlabeled learning with constraints.
\newblock \emph{Science China Information Sciences}, 65\penalty0 (11):\penalty0 212101, 2022.

\bibitem[Xia et~al.(2023)Xia, Xu, Ju, Hu, Chen, and Ge]{xia2023lmpt}
Xia, P., Xu, D., Ju, L., Hu, M., Chen, J., and Ge, Z.
\newblock Lmpt: Prompt tuning with class-specific embedding loss for long-tailed multi-label visual recognition.
\newblock \emph{arXiv preprint arXiv:2305.04536}, 2023.

\bibitem[Xu et~al.(2023)Xu, Liu, Yang, Chai, and Yuan]{xu2023learning}
Xu, Z., Liu, R., Yang, S., Chai, Z., and Yuan, C.
\newblock Learning imbalanced data with vision transformers.
\newblock In \emph{Proceedings of the IEEE/CVF Conference on Computer Vision and Pattern Recognition}, pp.\  15793--15803, 2023.

\bibitem[Yang \& Xu(2020)Yang and Xu]{yang2020rethinking}
Yang, Y. and Xu, Z.
\newblock Rethinking the value of labels for improving class-imbalanced learning.
\newblock In \emph{Advances in Neural Information Processing Systems}, volume~33, pp.\  19290--19301, 2020.

\bibitem[Yang et~al.(2022)Yang, Chen, Li, Xie, Lin, and Tao]{yang2022inducing}
Yang, Y., Chen, S., Li, X., Xie, L., Lin, Z., and Tao, D.
\newblock Inducing neural collapse in imbalanced learning: Do we really need a learnable classifier at the end of deep neural network?
\newblock In \emph{Advances in Neural Information Processing Systems}, volume~35, pp.\  37991--38002, 2022.

\bibitem[Yu et~al.(2023)Yu, Chang, Wang, Liu, Wang, Wang, Lin, Xie, Li, Lin, et~al.]{yu2023visual}
Yu, B.~X., Chang, J., Wang, H., Liu, L., Wang, S., Wang, Z., Lin, J., Xie, L., Li, H., Lin, Z., et~al.
\newblock Visual tuning.
\newblock \emph{arXiv preprint arXiv:2305.06061}, 2023.

\bibitem[Zaken et~al.(2022)Zaken, Goldberg, and Ravfogel]{zaken2022bitfit}
Zaken, E.~B., Goldberg, Y., and Ravfogel, S.
\newblock Bitfit: Simple parameter-efficient fine-tuning for transformer-based masked language-models.
\newblock In \emph{Proceedings of the 60th Annual Meeting of the Association for Computational Linguistics (Volume 2: Short Papers)}, pp.\  1--9, 2022.

\bibitem[Zhang et~al.(2021)Zhang, Li, Yan, He, and Sun]{zhang2021distribution}
Zhang, S., Li, Z., Yan, S., He, X., and Sun, J.
\newblock Distribution alignment: A unified framework for long-tail visual recognition.
\newblock In \emph{Proceedings of the IEEE/CVF Conference on Computer Vision and Pattern Recognition}, pp.\  2361--2370, 2021.

\bibitem[Zhao et~al.(2022)Zhao, Chen, Tan, Huang, and Zhu]{zhao2022adaptive}
Zhao, Y., Chen, W., Tan, X., Huang, K., and Zhu, J.
\newblock Adaptive logit adjustment loss for long-tailed visual recognition.
\newblock In \emph{Proceedings of the AAAI Conference on Artificial Intelligence}, volume~36, pp.\  3472--3480, 2022.

\bibitem[Zhong et~al.(2021)Zhong, Cui, Liu, and Jia]{zhong2021improving}
Zhong, Z., Cui, J., Liu, S., and Jia, J.
\newblock Improving calibration for long-tailed recognition.
\newblock In \emph{Proceedings of the IEEE/CVF Conference on Computer Vision and Pattern Recognition}, pp.\  16489--16498, 2021.

\bibitem[Zhou et~al.(2020)Zhou, Cui, Wei, and Chen]{zhou2020bbn}
Zhou, B., Cui, Q., Wei, X.-S., and Chen, Z.-M.
\newblock {BBN}: Bilateral-branch network with cumulative learning for long-tailed visual recognition.
\newblock In \emph{Proceedings of the IEEE/CVF Conference on Computer Vision and Pattern Recognition}, pp.\  9719--9728, 2020.

\bibitem[Zhou et~al.(2022{\natexlab{a}})Zhou, Yang, Loy, and Liu]{zhou2022conditional}
Zhou, K., Yang, J., Loy, C.~C., and Liu, Z.
\newblock Conditional prompt learning for vision-language models.
\newblock In \emph{Proceedings of the IEEE/CVF Conference on Computer Vision and Pattern Recognition}, pp.\  16816--16825, 2022{\natexlab{a}}.

\bibitem[Zhou et~al.(2022{\natexlab{b}})Zhou, Yang, Loy, and Liu]{zhou2022learning}
Zhou, K., Yang, J., Loy, C.~C., and Liu, Z.
\newblock Learning to prompt for vision-language models.
\newblock \emph{International Journal of Computer Vision}, 130\penalty0 (9):\penalty0 2337--2348, 2022{\natexlab{b}}.

\bibitem[Zhou et~al.(2023)Zhou, Guo, Jia, Zhang, and Li]{zhou2023ods}
Zhou, Z., Guo, L.-Z., Jia, L.-H., Zhang, D., and Li, Y.-F.
\newblock {ODS}: Test-time adaptation in the presence of open-world data shift.
\newblock In \emph{Proceedings of the 40th International Conference on Machine Learning}, pp.\  42574--42588, 2023.

\bibitem[Zhou et~al.(2024)Zhou, Yang, Shi, Guo, and Li]{zhou2024decoop}
Zhou, Z., Yang, M., Shi, J.-X., Guo, L.-Z., and Li, Y.-F.
\newblock Decoop: Robust prompt tuning with out-of-distribution detection.
\newblock In \emph{Proceedings of the 41st International Conference on Machine Learning}, 2024.

\bibitem[Zhu et~al.(2022)Zhu, Wang, Chen, Chen, and Jiang]{zhu2022balanced}
Zhu, J., Wang, Z., Chen, J., Chen, Y.-P.~P., and Jiang, Y.-G.
\newblock Balanced contrastive learning for long-tailed visual recognition.
\newblock In \emph{Proceedings of the IEEE/CVF Conference on Computer Vision and Pattern Recognition}, pp.\  6908--6917, 2022.

\end{thebibliography}
